\documentclass[10pt,twocolumn,letterpaper]{article}

\usepackage{wacv}              

\usepackage{graphicx}
\usepackage{amsmath}
\usepackage{amssymb}
\usepackage{booktabs}

\usepackage{adjustbox}
\usepackage{multirow}
\usepackage[accsupp]{axessibility}  
\usepackage{mathtools}
\usepackage{algpseudocode}
\usepackage{makecell}
\usepackage{algorithm}
\usepackage[linesnumbered,ruled,lined,boxed,algo2e]{algorithm2e}
\usepackage{longtable}
\usepackage[accsupp]{axessibility}

\usepackage[pagebackref,breaklinks,colorlinks]{hyperref}

\usepackage[capitalize]{cleveref}
\crefname{section}{Sec.}{Secs.}
\Crefname{section}{Section}{Sections}
\Crefname{table}{Table}{Tables}
\crefname{table}{Tab.}{Tabs.}

\def\wacvPaperID{1120} 
\def\confName{WACV}
\def\confYear{2024}

\begin{document}

\title{Planar Gaussian Splatting}

\author{Farhad G. Zanjani \qquad Hong Cai \qquad Hanno Ackermann \qquad Leila Mirvakhabova \qquad Fatih Porikli \\ \\
Qualcomm AI Research\thanks{Qualcomm AI Research is an initiative of Qualcomm Technologies, Inc.}\\
{\tt\small \{fzanjani, hongcai, hackerma, lmirvakh, fporikli\}@qti.qualcomm.com}
}
\maketitle
\begin{abstract}
   This paper presents Planar Gaussian Splatting (PGS), a novel neural rendering approach to learn the 3D geometry and parse the 3D planes of a scene, directly from multiple RGB images. The PGS leverages Gaussian primitives to model the scene and employ a hierarchical Gaussian mixture approach to group them. Similar Gaussians are progressively merged probabilistically in the tree-structured Gaussian mixtures to identify distinct 3D plane instances and form the overall 3D scene geometry. In order to enable the grouping, the Gaussian primitives contain additional parameters, such as plane descriptors derived by lifting 2D masks from a general 2D segmentation model and surface normals.
    Experiments show that the proposed PGS achieves state-of-the-art performance in 3D planar reconstruction without requiring either 3D plane labels or depth supervision. In contrast to existing supervised methods that have limited generalizability and struggle under domain shift, PGS maintains its performance across datasets thanks to its neural rendering and scene-specific optimization mechanism, while also being significantly faster than existing optimization-based approaches.

\end{abstract}

\section{Introduction}
\label{sec:intro}

Identifying 3D planar surfaces in indoor settings using multi-view posed monocular video is a pre-requisite for many applications, including augmented reality, virtual reality, robot navigation, and 3D interior modeling. 
Since man-made environments feature many diverse planar surfaces whose appearances can be ambiguous, this is a challenging task. By approximating scene geometry with a collection of basic planar shapes, we achieve a compact and efficient representation that facilitates interaction with the physical space.

Recent deep learning methods treat 3D planar surface understanding as supervised learning tasks, relying on annotations of either 2D planes~\cite{liu2018planenet, liu2019planercnn, yu2019single, tan2021planetr, agarwala2022planeformers} or 3D structures~\cite{xie2022planarrecon}. 
However, acquiring plane annotations in both high-quality and large-scale is an expensive endeavor. 
Furthermore, these models struggle to generalize to unseen scenes or those captured with different imaging sensors. 

Recent advancements in differentiable rendering enable 3D geometry reconstruction solely from multi-view 2D images, eliminating the need for 3D ground truth. While methods like Neural Radiance Fields (NeRF) and their successors~\cite{mildenhall2021nerf, verbin2022refnerf, deng2022depth} achieve impressive novel view synthesis (NVS) quality, it remains a challenge to extract explicit planar surfaces from their implicit representations~\cite{atzmon2019controlling}.
Specifically, volume-based approaches~\cite{wang2021neus, yariv2021volume, yu2022monosdf} rely on computationally expensive steps like ray marching and density field prediction for implicit surface modeling, followed by Marching Cubes~\cite{lorensen1987marching} for surface extraction and Sequential RANSAC~\cite{fischler1981random} for plane detection. These steps require careful tuning of numerous hyperparameters (e.g., in RANSAC), adding complexity and hindering broader application.

Comparing to implicit methods, explicit neural representations offer several advantages. They allow direct optimization of the geometry through volumetric tetrahedral mesh~\cite{gao2020learning,shen2021deep,munkberg2022extracting,chen2022mobilenerf}, triangle surface mesh~\cite{zanjani2024neural}, or point cloud~\cite{kerbl20233d} on the geometric primitives themselves. This makes it easier to constrain the reconstructed surfaces, for example, to be locally planar. However, most of existing explicit methods are primarily developed for novel view synthesis and require additional steps for planar reconstruction. Among existing explicit neural approaches, recently NMF~\cite{zanjani2024neural} has proposed direct optimization on the 3D vertex positions of a triangle mesh to jointly reconstruct the geometry and perform contrastive learning for 3D planar parsing. 

In this paper, we propose, Planar Gaussian Splatting (PGS), to represent planar surfaces with a set of Gaussian primitives, equipped with learned plane descriptors, which are jointly optimized with other Gaussian parameters, \ie, without requiring complex and error-prone post-hoc heuristics. More specifically, we propose a hierarchical, tree-structured Gaussian Mixture Model (GMM) to model the scene. This probabilistic approach permits grouping the Gaussian geometric primitives trained by Gaussian Splatting while providing an interpretable interface for parsing and optimizing for planes. In order to facilitate the grouping, we leverage 2D segmentation from foundation models like SAM~\cite{kirillov2023segment}. This allows our algorithm to optimize multi-view partial consistency between 2D segmentation pseudo labels to identify 3D planes while optimizing for the 3D geometry concurrently.

\begin{figure*}[t!]
\center
\includegraphics[width=.85\textwidth, trim={4cm 5cm 3cm 4cm}]{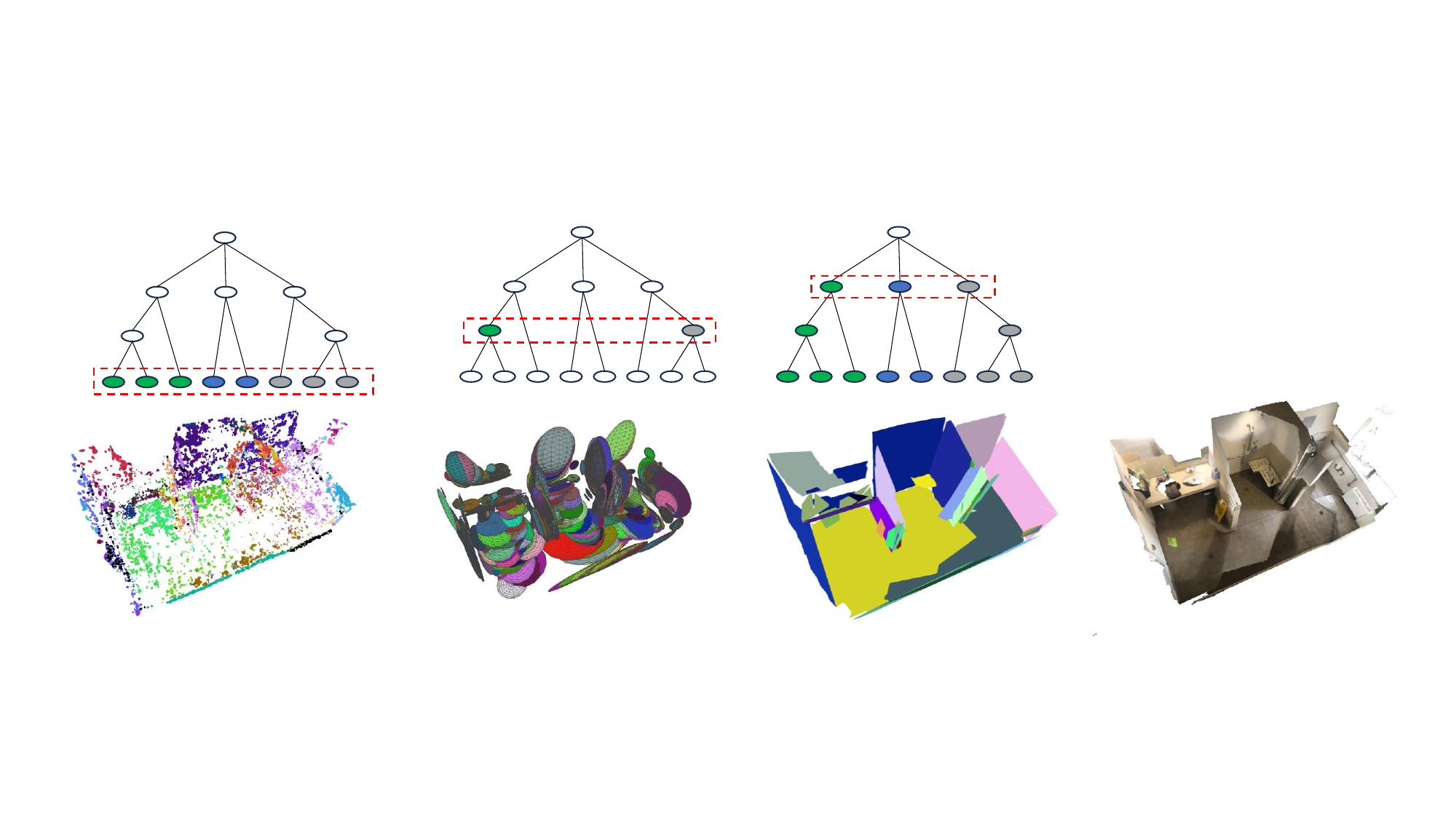}
\vspace{-2pt}
\caption{\small The proposed PGS method constructs the entire scene using a tree-structured arrangement of Gaussian nodes. At the leaf nodes, Gaussian primitives are optimized through Gaussian splatting rendering. Meanwhile, parent nodes are recursively formed by merging Gaussian nodes probabilistically. The child nodes of the root correspond to distinct 3D planes within the scene, as depicted in the accompanying figure.}
\vspace{-5pt}
\label{fig:graphic}
\end{figure*}

Our main contributions are summarized as follows:
\begin{itemize}
    \vspace{-5pt}
    \item We propose Planar Gaussian Splatting (PGS), a novel unsupervised, neural-rendering-based framework for 3D planar reconstruction of a scene from RGB images. PGS does not require any 3D labels or depth supervision (either ground-truth or predicted).
    \vspace{-5pt}
    \item We leverage 3D Gaussian primitives to model the scene, which are probabilistically grouped via hierarchical, tree-structured Gaussian mixtures to identify 3D planar instances.
    \vspace{-5pt}
    \item In order to enable the grouping, we propose discriminative 3D descriptors, learned from Segment Anything~\cite{kirillov2023segment} proposal masks. We resolve challenges such as the lack of multi-view proposal association and variable numbers of masks, by formulating and solving a linear regression problem followed by merging segments using a Region Adjacency Graph (RAG).
    \vspace{-5pt}
    \item PGS achieves state-of-the-art performance in 3D planar reconstruction, as compared to existing supervised and optimization-based methods. In particular, it can readily reconstruct 3D planes on any new test scenes, which cannot be done well by supervised learning models.
\end{itemize}

\section{Related work}
\vspace{-3pt}
\subsection{Planar Reconstruction}
\vspace{-5pt}
Planar reconstruction from a single RGB image have been investigated in several works, for instance using ConvNets~\cite{yang2018recovering, liu2018planenet}.
Those works predict both segmentation and 3D plane parameters and furthermore require a prescribed maximum number of planes in an image, which limits model applicability. 
Other works address this limitation on single image planar reconstruction~\cite{liu2019planercnn, yu2019single, qian2020learning} and can handle any number of planes.
PlaneTR \cite{tan2021planetr} leverages transformers to consider context information and geometric cues like line segmentation and ground-truth depth in a sequence-to-sequence way. Dependencies on ground-truth depth or plane annotation and single image reconstruction limit the applications of these approaches.

Alternatively, multi-view reconstruction utilizes multiple images, which contain richer geometric information. Several works share a common two-stage approach: local plane detection and plane parameter estimation~\cite{jin2021planar, liu2022planemvs}. More recently, PlanarRecon~\cite{xie2022planarrecon} proposes to detect planes from video fragments and combine them to create a comprehensive global planar reconstruction, which is supervised by 3D ground-truth planes in training. In contrast, this paper presents a multi-view 3D planar surface reconstruction method without requiring 2D or 3D plane annotations. 

\vspace{-3pt}
\subsection{Objects \& Semantics in Volumetric Rendering}
\vspace{-5pt}
In recent years, there has been significant progress in radiance field rendering, particularly in the context of semantic 3D modeling and scene decomposition. 

\textbf{Neural scene representations for scene decomposition}:
Some works~\cite{fan2022nerf, mirzaei2022laterf, sharma2023neural, tschernezki2021neuraldiff, xie2021fig, yu2021unsupervised} employ neural scene representations to decompose scenes into foreground and background components, remarkably without explicit supervision or only relying on weak signals such as text or object motion. Semantic NeRF~\cite{zhi2021place}, for instance, introduces a separate branch that predicts semantic labels. NeSF~\cite{vora2021nesf} predicts a semantic field by utilizing a density field as input to a 3D semantic segmentation model.

\textbf{Supplementing NeRFs with 2D annotations}:
Some works have explored ways to enhance NeRF models using readily available 2D annotations from datasets. Panoptic NeRF~\cite{fu2022panoptic} and Instance-NeRF~\cite{liu2023instance} incorporate 3D instance supervision, allowing for more accurate scene representation. Contrastive Lift~\cite{bhalgat2023contrastive} takes a novel approach by lifting 2D instance segmentation to 3D without relying on explicit 3D masks. It achieves this through contrastive learning. Similarly, NMF~\cite{zanjani2024neural} introduces a contrastive learning scheme for lifting 2D pixel clusters into a 3D mesh. Panoptic Lifting~\cite{siddiqui2023panoptic} addresses the challenge of lifting 2D instance segmentation by employing linear assignment techniques to ensure consistency across multi-view annotations.

\textbf{Fusing 2D analysis output into 3D space}:
The aforementioned works — Semantic NeRF~\cite{zhi2021place}, Panoptic Lifting~\cite{siddiqui2023panoptic} and  Constrastive Lift~\cite{bhalgat2023contrastive} — represent a recent research direction. They aim to seamlessly integrate the insights gained from 2D analysis into the 3D domain. SA3D~\cite{cen2024segment}, leverages the powerful vision foundation model SAM~\cite{kirillov2023segment}. SAM excels at segmenting objects in 2D images, and SA3D utilizes this capability for interactive 3D segmentation using NeRF. 

Our work shares similar concept with the above mentioned method since the proposed method lifts the 2D analysis into the 3D domain but has a different objective which is 3D planar reconstruction of the scene. A more recent and most similar work, NMF~\cite{zanjani2024neural} introduces an explicit rendering approach for 3D planar reconstruction by optimizing the vertices of triangle mesh and incorporating a contrastive learning for jointly learning the scene geometry and decomposition the mesh into planar surfaces. NMF leverages a depth prediction network for lifting the 2D analysis into 3D. Our approach doesn't require the ground truth or predicted depth network. Furthermore, similar to Constrastive Lift~\cite{bhalgat2023contrastive}, the NMF utilizes a clustering method as post-processing for grouping the features of the plane or object instances. The constant hyper-parameters of clustering introduces a sub-optimal solution across different scene, while our approach utilizes a probabilistic method for grouping the features.

These advancements bridge the gap between 2D and 3D representations, opening up exciting possibilities for richer scene understanding and modeling.

\vspace{-0pt}
\section{Planar Gaussian Splatting}
\vspace{-5pt}
In this section, we present the proposed Planar Gaussian Splatting (PGS) method. Section~\ref{sec:primer} starts with a primer on Gaussian Splatting and the standard parameterization of the Gaussian primitives. Section~\ref{sec:GMT} introduces the Gaussian Mixture Tree, a probabilistic approach for planar construction of the scene geometry in a bottom-up modeling. Section~\ref{sec:lpd} introduces a learning procedure for optimizing a latent vector per Gaussian primitives, called plane descriptor which represents 3D plane instances. Afterwards, local planar alignment as a geometric constraint applies on Gaussian positions close to surfaces is explained in Section~\ref{sec:lpa}. In Section~\ref{sec:meanshift}, we discussed how a holistic separability across descriptors is maintained using recurrent mean-shift layer.   

\vspace{-2pt}
\subsection{Primer on 3D Gaussian Splatting (3DGS)}\label{sec:primer}
\vspace{-4pt}
3DGS~\cite{kerbl20233d} models the scene as a set of multivariate Gaussians in 3D space, which is an explicit form of representation, in contrast to the implicit representation used in NeRF. Each Gaussian is characterized by a covariance matrix ${\Sigma}$ and a center (mean) point $\mathbf{\mu}$, \ie, \vspace{-5pt}
\begin{equation}\label{eq:3DGaussian}
    G(\mathbf{x}) = e^{-\frac{1}{2}(\mathbf{x}-\mathbf{\mu})^{T}{\Sigma}^{-1}(\mathbf{x}-\mathbf{\mu})}. \\[-5pt]
\end{equation}

The centers of these 3D Gaussians are initialized from a set of sparse points (\eg, randomly initialized or obtained from SfM). More specifically, each Gaussian is parameterized by the following parameters: (a) a center position $\mathbf{\mu} \in \mathbb{R}^3$, (b) a covariance matrix which has the form ${\Sigma=RSS^{T}R^{T}}$ computed from scaling $\mathbf{s}\in \mathbb{R}^3$ and rotation factors $\mathbf{r} \in \mathbb{R}^4$ (in quaternion), (c) opacity $\alpha \in \mathbb{R}$, and (d) spherical harmonics (SH) coefficient $\mathbf{c} \in \mathbb{R}^k$ that represents the color, where $k$ is the degree of the SH. 
Given a view transform $W$, the 2D covariance matrix in camera coordinates can be expressed by $\Sigma^{2D}=JW\Sigma W^{T}J^{T}$, where $J$ is the Jacobian of the affine approximation of the projection transformation. For each pixel in the image, rendering the color is performed as in~\cite{kerbl20233d} by blending the color vectors of $N$ ordered  Gaussians, which overlap at the pixel position $(u,v)$, by \vspace{-5pt}
\begin{equation}\label{eq:color_rendering}
\begin{split}
\hat{\mathbf{c}}_{uv}=&\sum_{i}^{N} \mathbf{c}_{i}\alpha_{i}\prod\limits_{j}^{i-1} (1-\alpha_{j}), \\   
\mathcal{L}_{\text{rgb}}=& \sum_{(u,v)}\left\Vert \hat{\mathbf{c}}_{uv}-\mathbf{c}_{uv} \right\Vert_{1} + \lambda \cdot \texttt{SSIM}(\hat{\mathbf{c}}_{uv}, \mathbf{c}_{uv}), \\[-5pt]
\end{split}
\end{equation}
where $\alpha_{i}$ and $\mathbf{c}_{i}$ are learnable opacity and RGB color of $i^{th}$ Gaussian, obtained by SH coefficients. The weighting coefficient $\lambda$ is set the same as in~\cite{kerbl20233d}. 
By minimizing the loss in Eq.~\ref{eq:color_rendering}, the model learns the 3D scene geometry through the sparse unstructured Gaussian primitives by optimizing their opacity parameters. In addition to optimizing the Gaussian parameters, the training process involves splitting, cloning, and culling of Gaussian primitives to express the scene geometry by maximizing the photometric likelihood between the rendered and the actual images, as proposed in~\cite{kerbl20233d}.

\vspace{-2pt}
\subsection{Gaussian Mixture Tree}
\vspace{-4pt}\label{sec:GMT}
Optimizing the parameters of 3D Gaussians leads to spatially moving the center points close to the object surfaces in the scene. In order to identify distinct 3D plane instances, a novel probabilistic approach is proposed that involves compositional modeling of the 3D scene using a Gaussian Mixture Tree (GMT).  

The whole scene is modeled in a tree structure, which is constructed recursively from the leaf nodes to the root node. The Gaussian Mixture Model (GMM) is involved to join nodes and derive intermediate parent nodes. In this way, the GMT represents the entire scene in a hierarchical way. In the GMT, the child nodes of the root represent individual 3D plane instances; see Figure~\ref{fig:graphic}. 

Each parent in the tree (except the root) specify a Gaussian distribution, $G_p(\mathbf{\mu}_p, \Sigma_p)$, in 3D space, which encompasses all the center points ($\mathbf{\mu}$) of its child Gaussian nodes. Since each node specifies a Gaussian distribution, merging two nodes is equivalent to merging their distributions. More specifically, for two Gaussian nodes, the merging is performed as follows. \vspace{-5pt}
\begin{equation}\label{eq:merge}
\begin{split}
    \mathbf{\Sigma}_p =& \mathbf{\Sigma}_j \cdot (\mathbf{\Sigma}_i+\mathbf{\Sigma}_j)^{-1} \cdot \mathbf{\Sigma}_i, \\
    \mathbf{\mu}_p =& \mathbf{\Sigma}_j \cdot (\mathbf{\Sigma}_i+\mathbf{\Sigma}_j)^{-1} \cdot \mathbf{\mu}_i + \mathbf{\Sigma}_i \cdot (\mathbf{\Sigma}_i+\mathbf{\Sigma}_j)^{-1} \cdot \mathbf{\mu}_j, \\[-5pt]
\end{split}
\end{equation}
where $(\mathbf{\mu}_p, \mathbf{\Sigma}_p)$ specifies the Gaussian parameters of the parent node. 

The merging criteria are based on both the Bhattacharya distance~\cite{hennig2010methods} between the two respective Gaussian distributions, and the cosine similarity between the descriptors of the nodes $\langle \mathbf{z}_i \cdot \mathbf{z}_j\rangle$ (which will be explained in the following part). For two multivariate Gaussian distributions ($G_i, G_j$), the Bhattacharya distance $D_B(G_i,G_j)$ is given by: \vspace{-5pt}
\begin{equation}\label{eq:db}
    \begin{split}
    D_B(G_i, G_j) = &\frac{1}{8}(\mathbf{\mu}_i-\mathbf{\mu}_j)^{T}\mathbf{\Sigma}^{-1}(\mathbf{\mu}_i-\mathbf{\mu}_j) \\ &+\frac{1}{2}\ln(\frac{\det\mathbf{\Sigma}}{\sqrt{\det\mathbf{\Sigma}_i \det\mathbf{\Sigma}_j}}), \\[-5pt]
    \end{split}
\end{equation}
where $\mathbf{\Sigma}=\frac{\mathbf{\Sigma_i}+\mathbf{\Sigma_j}}{2}$. The tree structure simply is formed by merging every two nodes whose descriptors are similar and whose Bhattacharya distance is lower than a predefined threshold. Algorithm~\ref{alg:psudocode} shows pseudocode of the proposed GMT.

When constructing the GMT, we do not directly use the Gaussian primitives (\ie, parameters obtained through Gaussian splatting optimization) at the leaf nodes, which would be computationally expensive given as there can be a huge number of them in the 3D scene. Instead, we first group the Gaussian primitives into local clusters and use each cluster as the leaf node of the GMT. 
For locally grouping the Gaussian primitives, two additional parameters are introduced for each Gaussian during the optimization, which are the surface normal vector at the center location of the Gaussian and a learnable vector called \emph{plane descriptor} which can be used to identify distinct 3D planes. 

\begin{minipage}[h!]{.92\columnwidth}
\vspace{-5pt}
\begin{algorithm}[H]
\small
\LinesNumberedHidden
\caption{\small Hierarchically Merging Gaussians}
\label{alg:psudocode}
\SetAlgoLined
\SetKwInOut{Input}{Input}
\SetKwInOut{Output}{Output}
\Input{$\{G_i(\mathbf{\mu}_i, \mathbf{\Sigma}_i, \mathbf{n}_i, \mathbf{z}_i)\}_{i=1}^{N_s}$, \,and $N_s$ leaf nodes}
\Output{ $p(\mathbf{x}|\mathbf{G}^s)=\sum_{k=1}^{L} \pi_k \cdot p(\mathbf{x}|\mathbf{G}_k^s)$, \;\; $\texttt{where}\; x\in \mathbb{R}^3$}
  
Initialize: $L=0, G^{s}=\emptyset$
 
\For{i=1:$N_s$}{
    \If{i is NOT descendant of  $\{G_{k}^{s}\}_{k=1}^{L}$}{ 
        \For{j=i+1:$N_s$}{
            \If{j is NOT descendant of  $\{G_{k}^{s}\}_{k=1}^{L}$}{
                \If{$D_B(G_i,G_j) \leq \epsilon_B$ and $1-\langle \mathbf{z}_i \cdot \mathbf{z}_j\rangle  \leq \epsilon_z$}{
                    $G_{p}^{ij}(\mathbf{\mu}_p, \mathbf{\Sigma}_p)\leftarrow G_i + G_j$ \tcp{(\ref{eq:merge})}
                    $G_i \leftarrow G_{p}^{ij}$
                }
            }
        }
    $G^{s}.insert(G_i)$  \tcp{Add to plane nodes}
    $L\, \leftarrow L+1$
    }
 }
\end{algorithm}
\vspace{-5pt}
\end{minipage}

\vspace{-5pt}
\subsection{Learning Plane Descriptors}\label{sec:lpd}
\vspace{-5pt}
To organize Gaussian primitives within the GMT hierarchy, two additional parameters for each Gaussian primitive are introduced: a normal vector $n \in \mathbb{R}^3$ and a plane descriptor $z \in \mathbb{R}^k$ (\eg, $k=3$).

\textbf{Lifting 2D normal maps to 3D}: To learn the normal vectors in 3D field, we employ an off-the-shelf network that predicts the normal map for the 2D training images; specifically, we use the Omnidata model~\cite{eftekhar2021omnidata}. To lift the normal vectors from 2D to 3D, similar to Eq.~\ref{eq:color_rendering}, the normal vectors are rendered for the camera view and compared with the off-the-shelf network's prediction ($\mathbf{n}_{uv}$) at pixel position $(u,v)$, using the cosine distance. The normal loss is defined as:\vspace{-5pt}
\begin{equation}\label{eq:normal_rendering}
\mathcal{L}_{\text{n}}=\sum_{(u,v)} \Bigl(1-\bigl\langle\hat{\mathbf{n}}_{uv}\cdot \mathbf{n}_{uv}\bigr\rangle \Bigr), \; \hat{\mathbf{n}}_{uv}=\sum_{i}^{N} \mathbf{n}_{i}\alpha_{i}\prod\limits_{j}^{i-1} (1-\alpha_{j}).\\[-3pt]
\end{equation}

\textbf{Lifting 2D SAM masks to 3D}:
To learn the plane descriptors, we leverage the 2D masks generated by the Segment Anything Model (SAM)~\cite{kirillov2023segment}. SAM segments the input 2D images into object parts. We prompt SAM with 32$\times$32 regular grid points. For each point, SAM predicts a set of masks that may correspond to different parts of objects. 
SAM incorporates ambiguity-aware modeling: if a point lies on a part or subpart, SAM may return the subpart, the part, or the entire object. The image segments with high variance in their normal vectors are ignored and are considered as invalid regions for the current camera frame.  

In order to learn the plane descriptors in the 3D Gaussian field, the valid 2D image segments need to be lifted to 3D. However, the masks from SAM have no semantic information and 2D-3D lifting is not straightforward due to two challenges: (1) the mask associations across different views are unknown, and (2) the number of segments is variable in each image and the maximum number of segments is unknown. In the following, we discuss how we handle these issues and utilize the 2D segments to learn the 3D plane descriptors. 

We restrict the descriptor $\mathbf{z}$ to have a vector norm equal to one ($\left\lVert \mathbf{z} \right\rVert_2=1$).  
To address the aforementioned challenge of lifting the 2D segments to create 3D descriptors, we propose a linear regression approach (with closed-form solution) to predict the indices of 2D segments based on $\mathbf{z}$, for each individual training image. More specifically, given the camera pose, we first render a 2D descriptor image where each pixel is computed by blending the 3D descriptors of Gaussian field for given camera view. Next, we use a linear layer ($\mathbf{W}$) to map each pixel in this descriptor image to a one-hot vector $\mathbf{y}$ that encodes a segment. $\mathbf{W}$ can be solved analytically. In matrix form, this is specified as follows. \vspace{-5pt}
\begin{equation}\label{eq:linear_solver}
    \begin{split}
    &\mathbf{Y} = [\mathbf{Z}|\mathbf{1}]\!\cdot\! \mathbf{W} \; \rightarrow \; \hat{\mathbf{W}}=(\mathbf{Z}^T\!\cdot\! \mathbf{Z})^{-1}\! \cdot\! \mathbf{Z}^T\! \cdot\! \mathbf{Y}, \\
    &\mathcal{L}_{seg} = \sum_{i}{\left\lVert y_i-\hat{y}_i \right\rVert_1}, \\[-5pt]
    \end{split}
\end{equation}
where $\mathbf{y}_i\in \{0,1\}^m$ and $\sum_{j=1}^{m}{y}_{ij}=1$. The $\mathbf{Y}$ and $\mathbf{Z}$ denote the matrix form of the labels and descriptors, the loss is computed by comparing the prediction $\hat{\mathbf{y}}= [\mathbf{z}|\mathbf{1}]\cdot \hat{\mathbf{w}}$ with the labels $\mathbf{y}$. 

\begin{figure}[t!]
\centering
\begin{adjustbox}{width=\columnwidth}
\begin{tabular}{cccc}
\includegraphics[width=.33\linewidth]{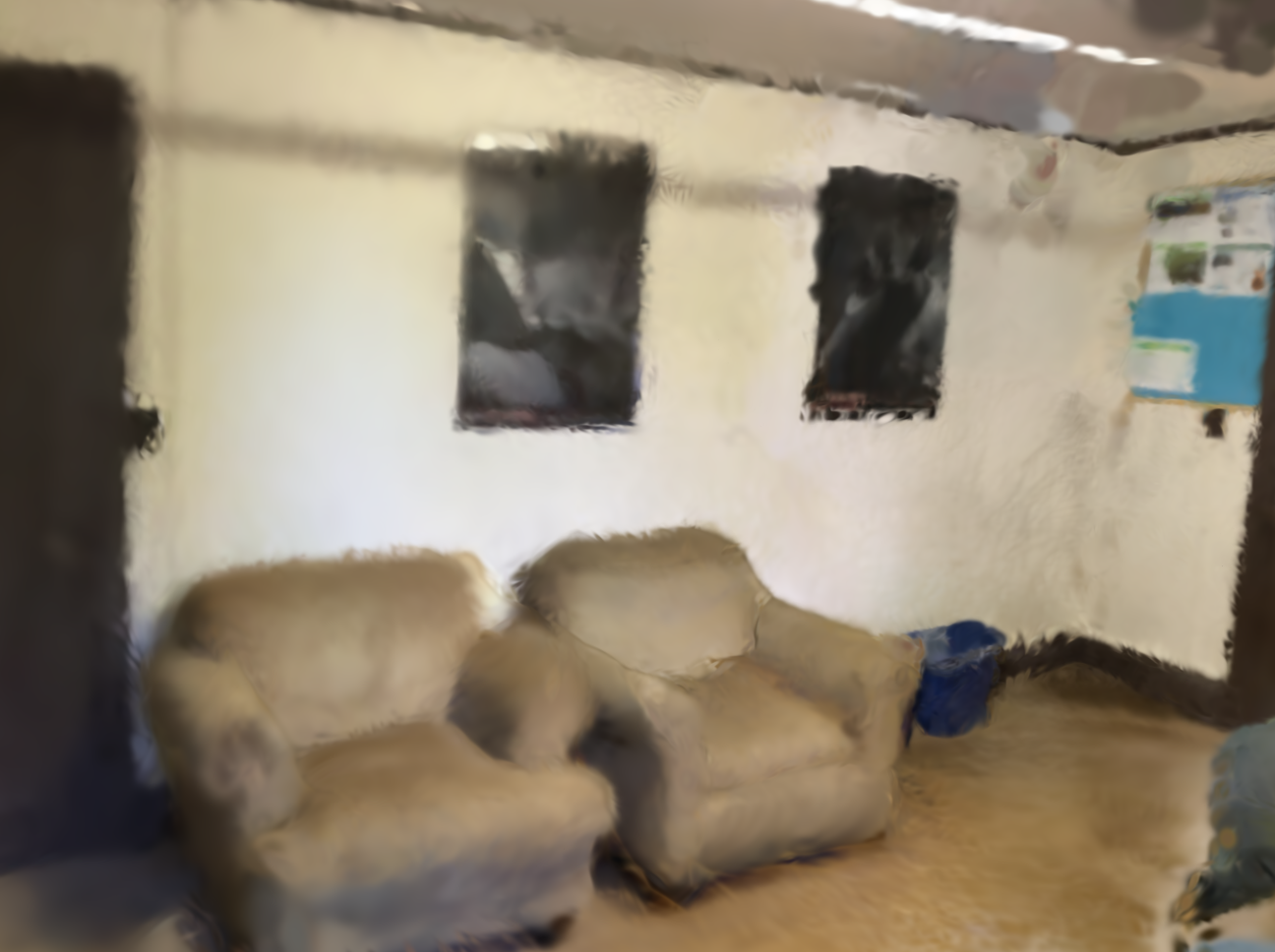}&
\includegraphics[width=.33\linewidth]{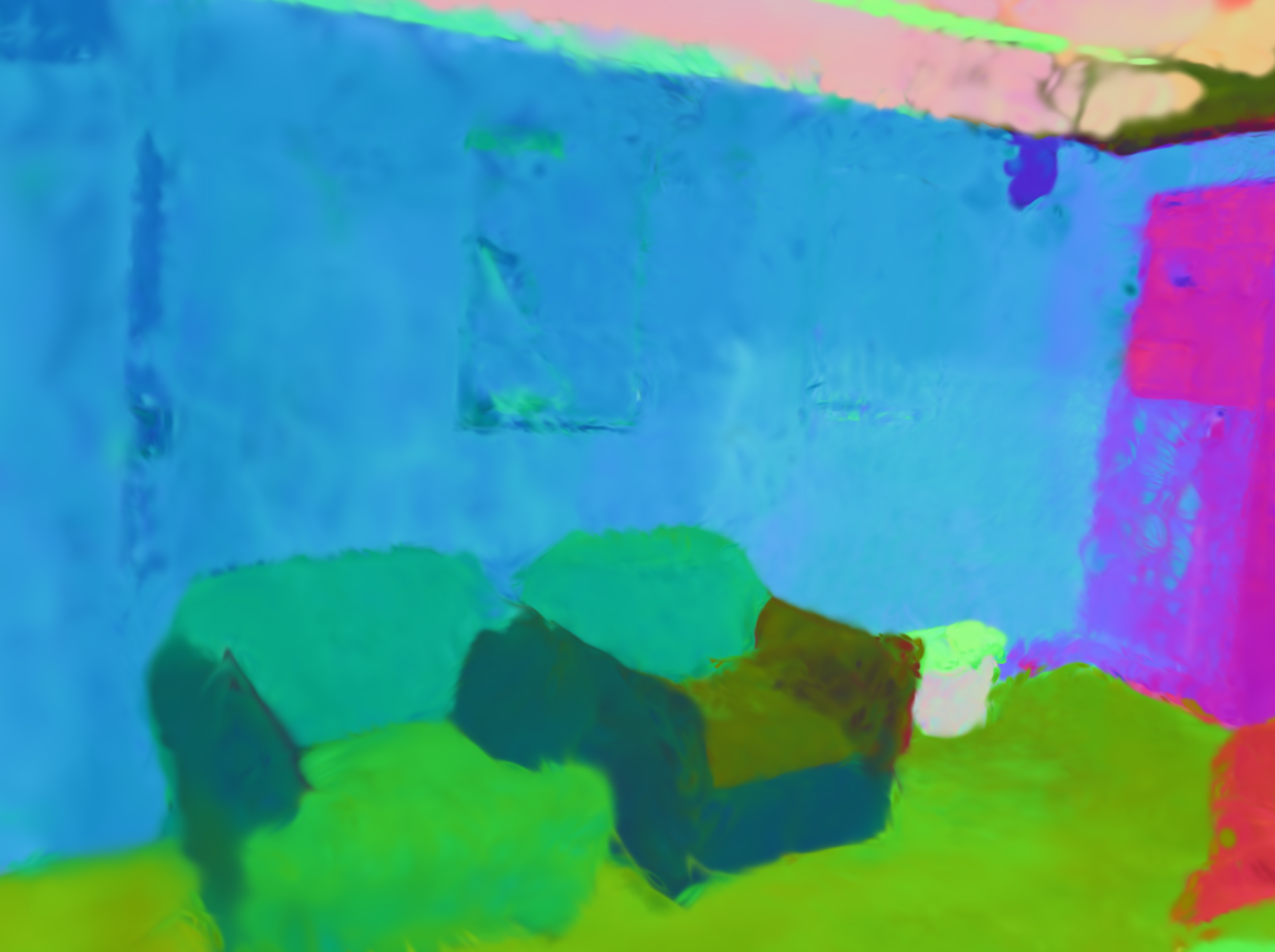}&
\includegraphics[width=.33\linewidth]{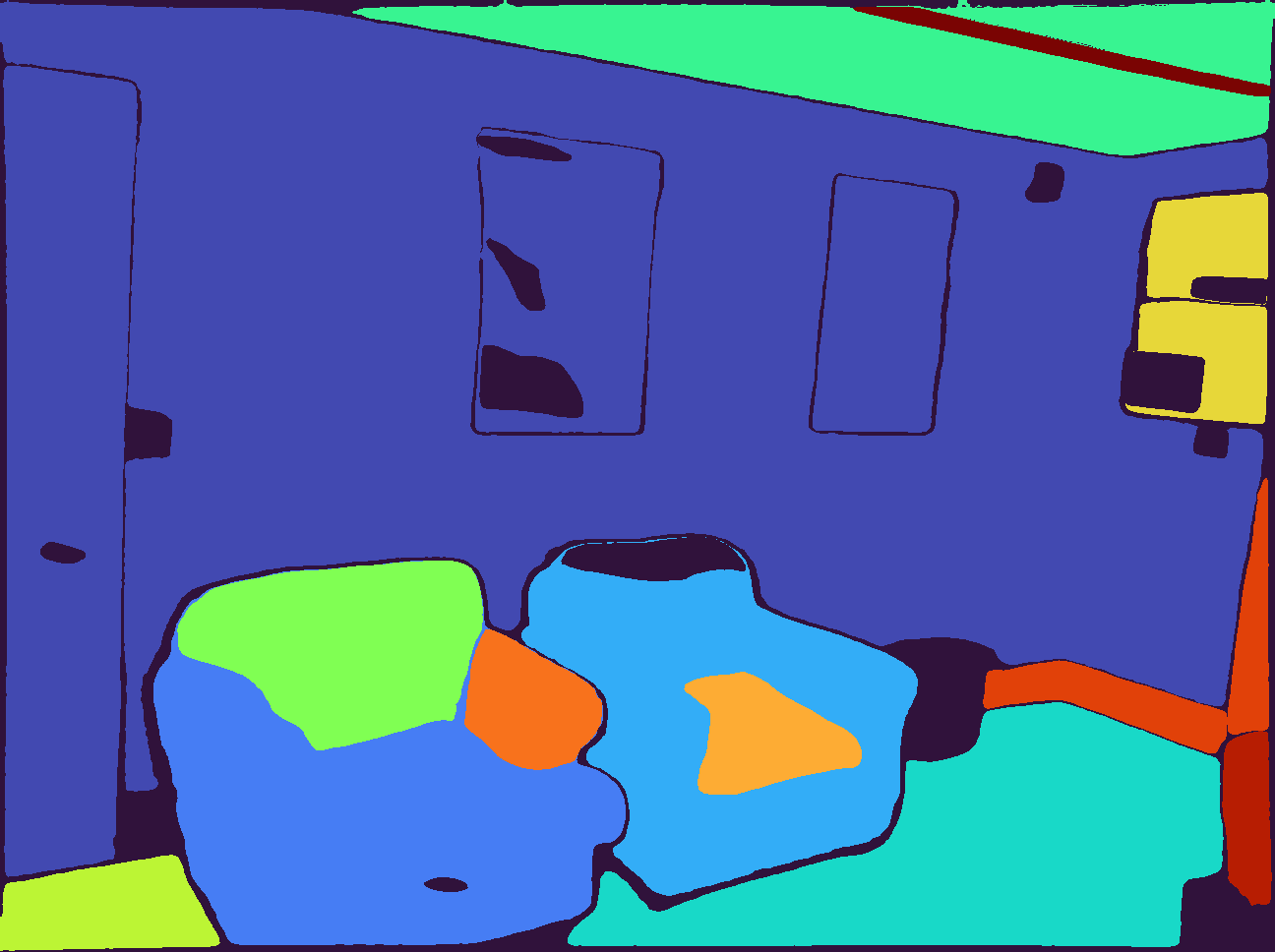}&
\includegraphics[width=.33\linewidth]{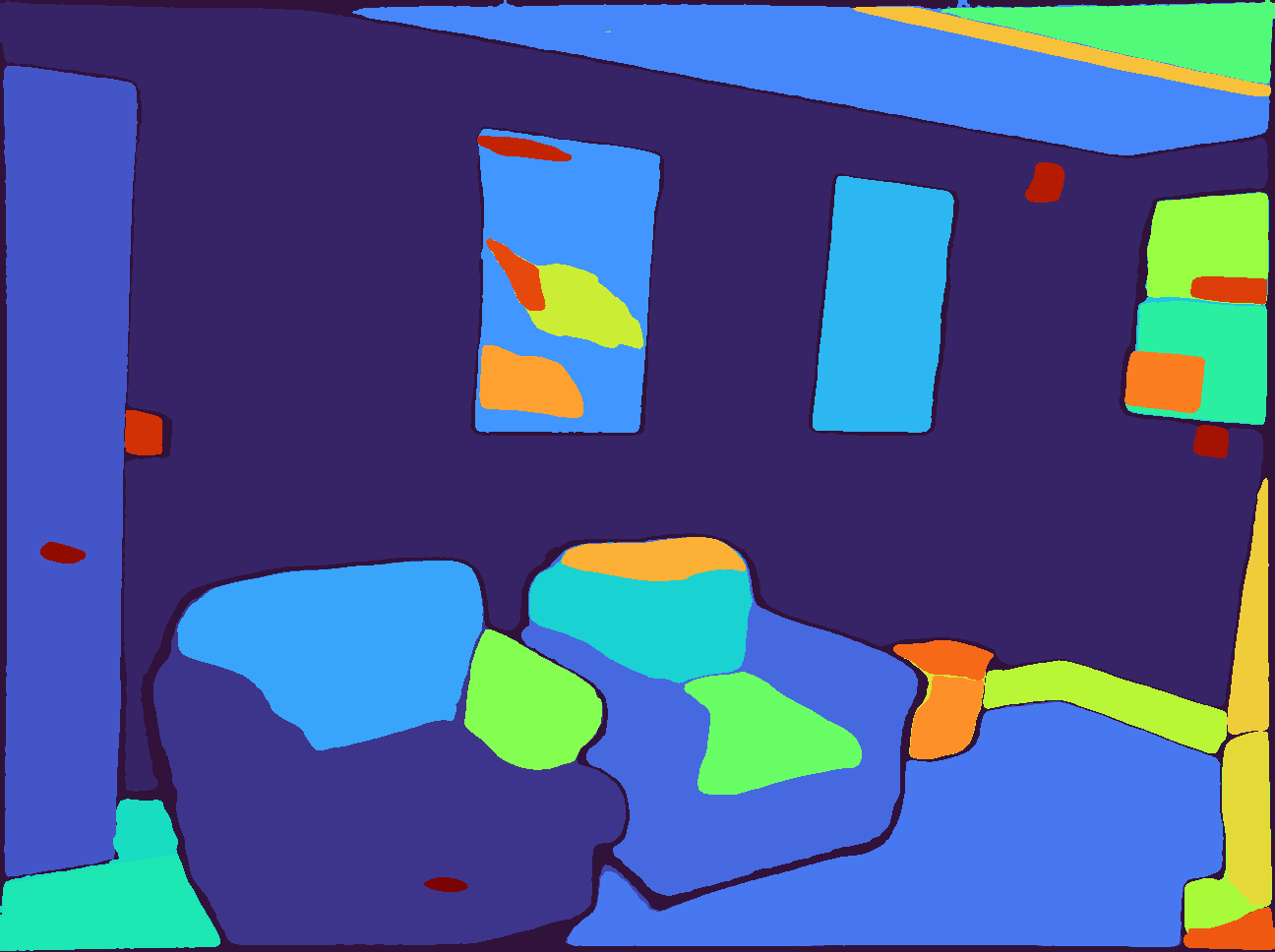}\\
\includegraphics[width=.33\linewidth]{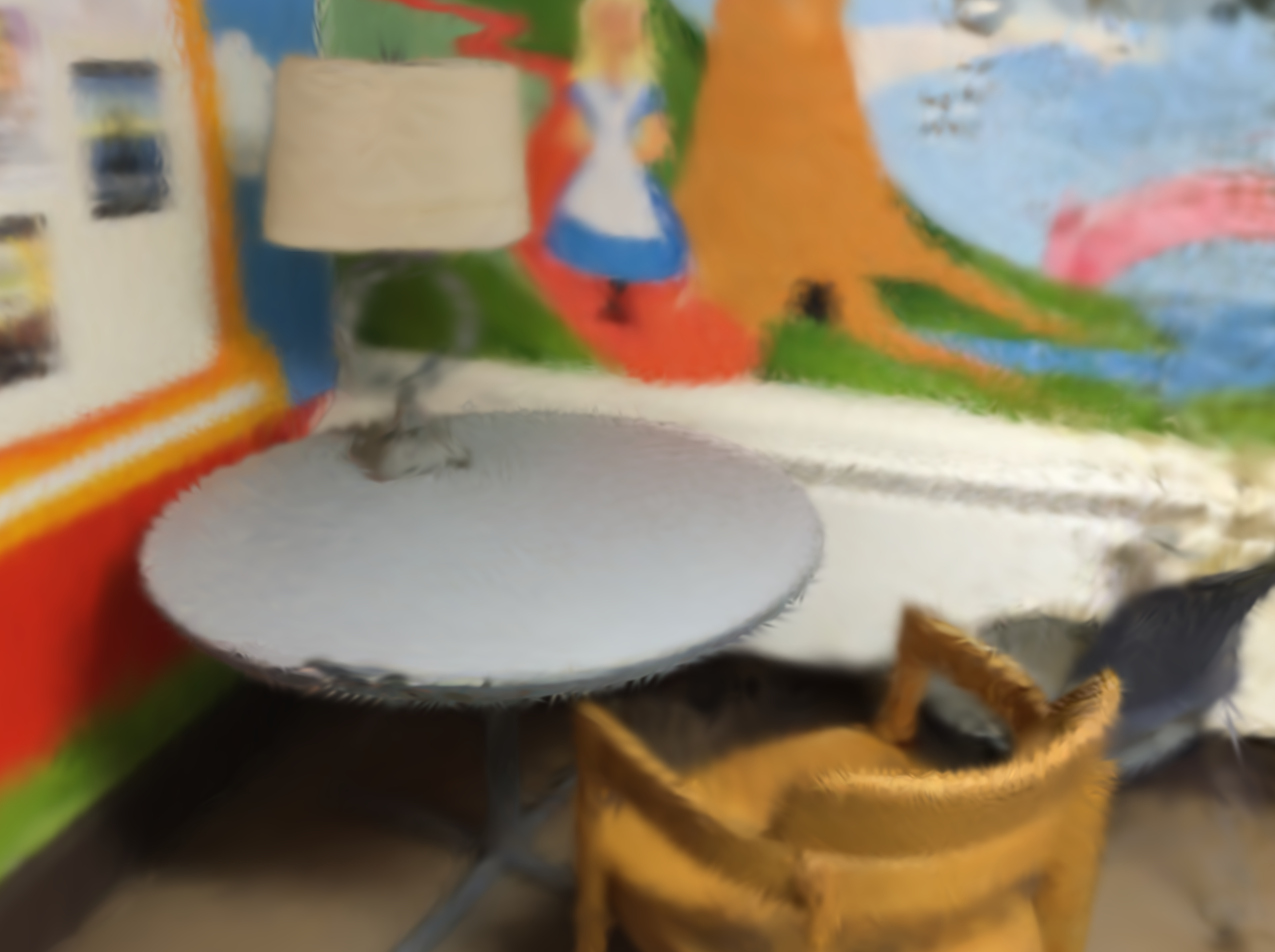}&
\includegraphics[width=.33\linewidth]{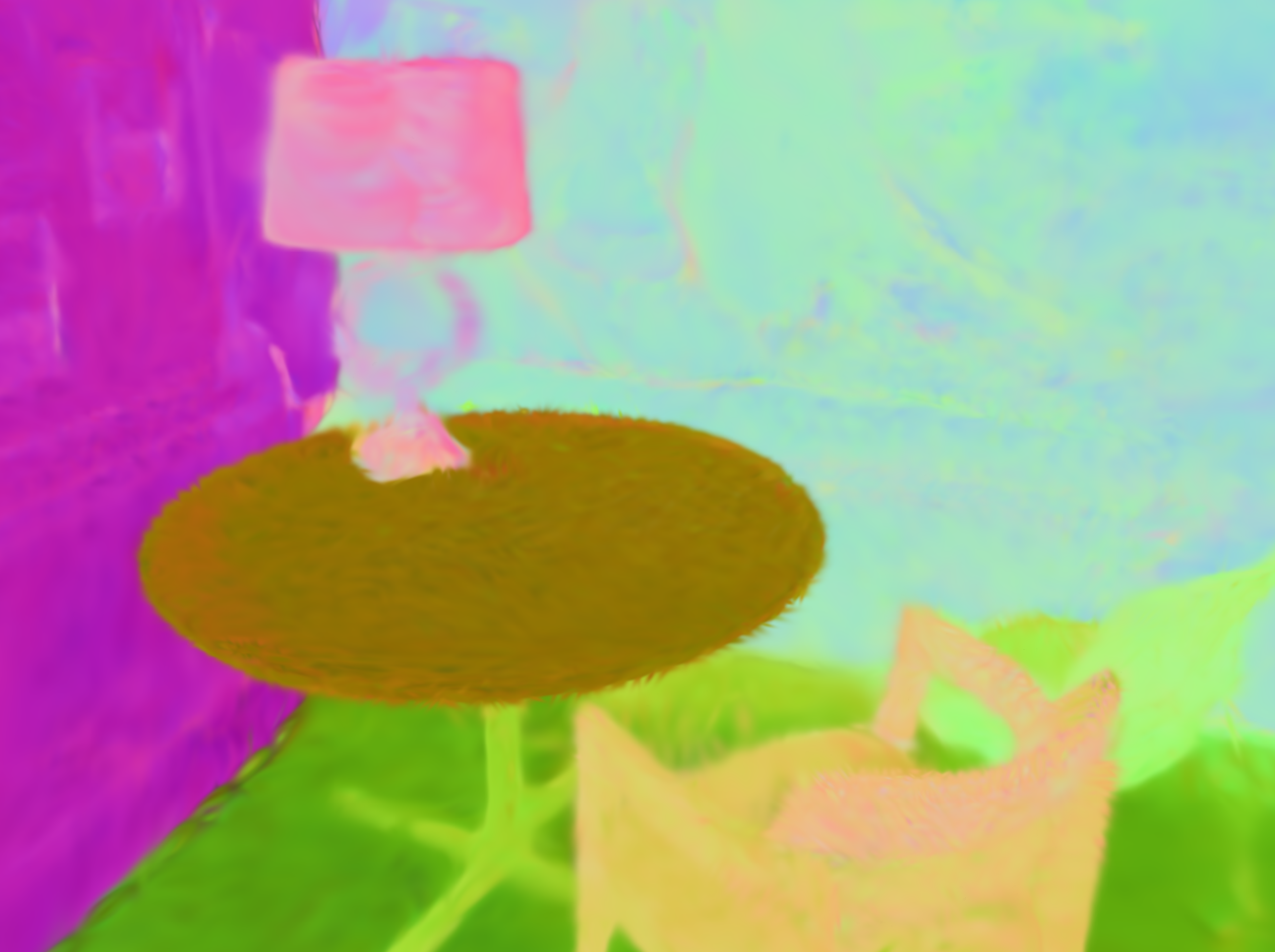}&
\includegraphics[width=.33\linewidth]{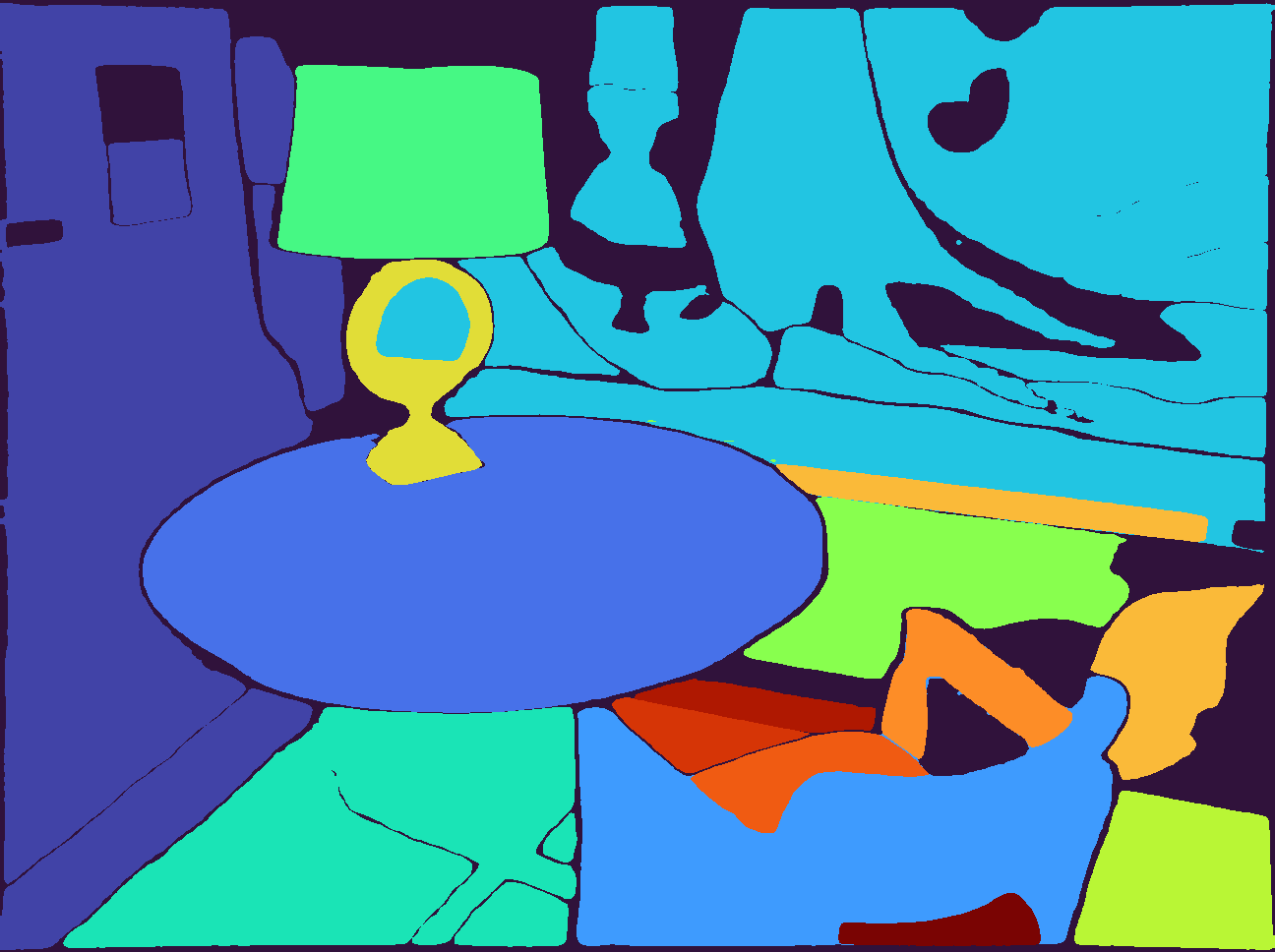}&
\includegraphics[width=.33\linewidth]{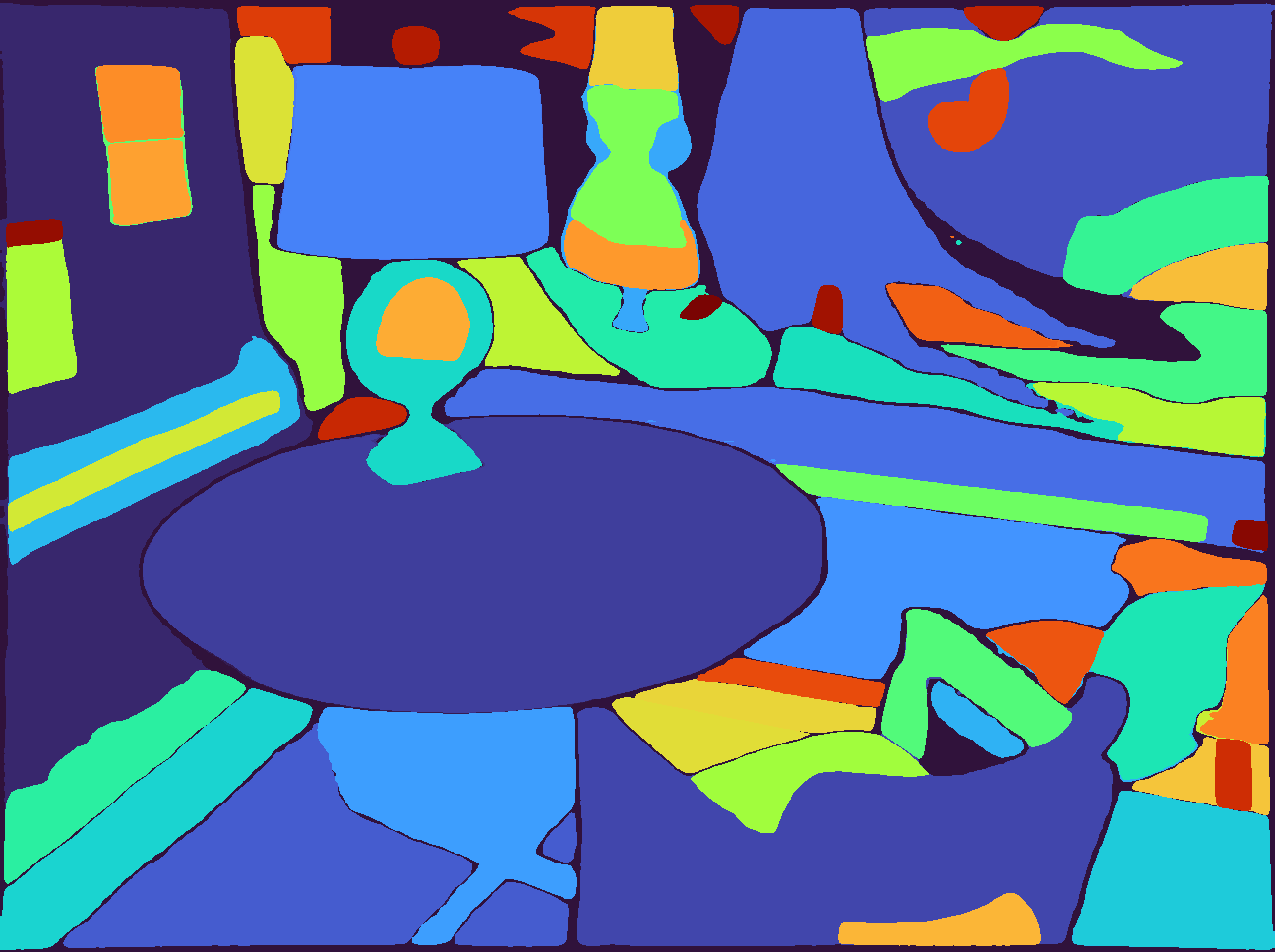}\\

\includegraphics[width=.33\linewidth]{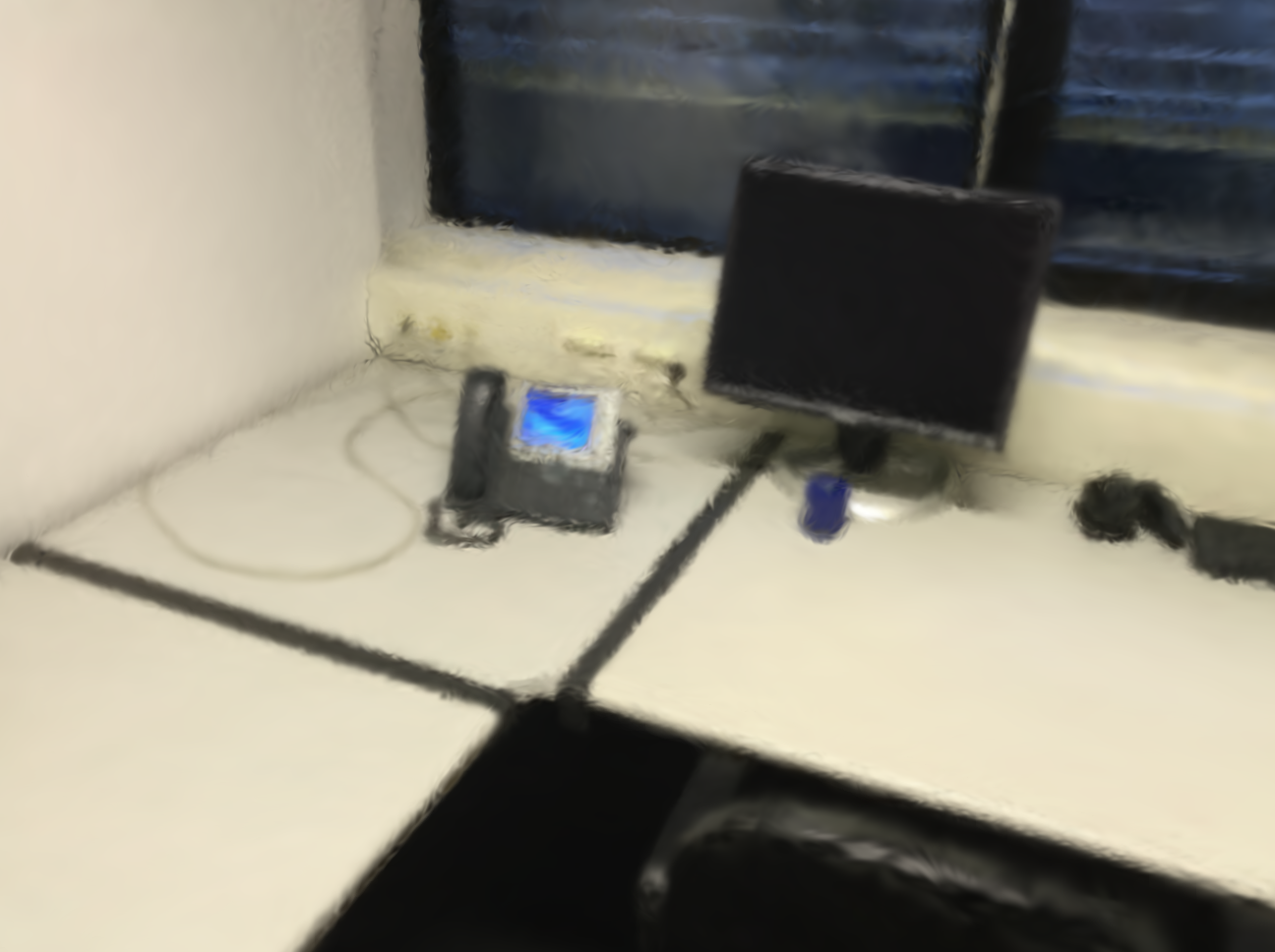}&
\includegraphics[width=.33\linewidth]{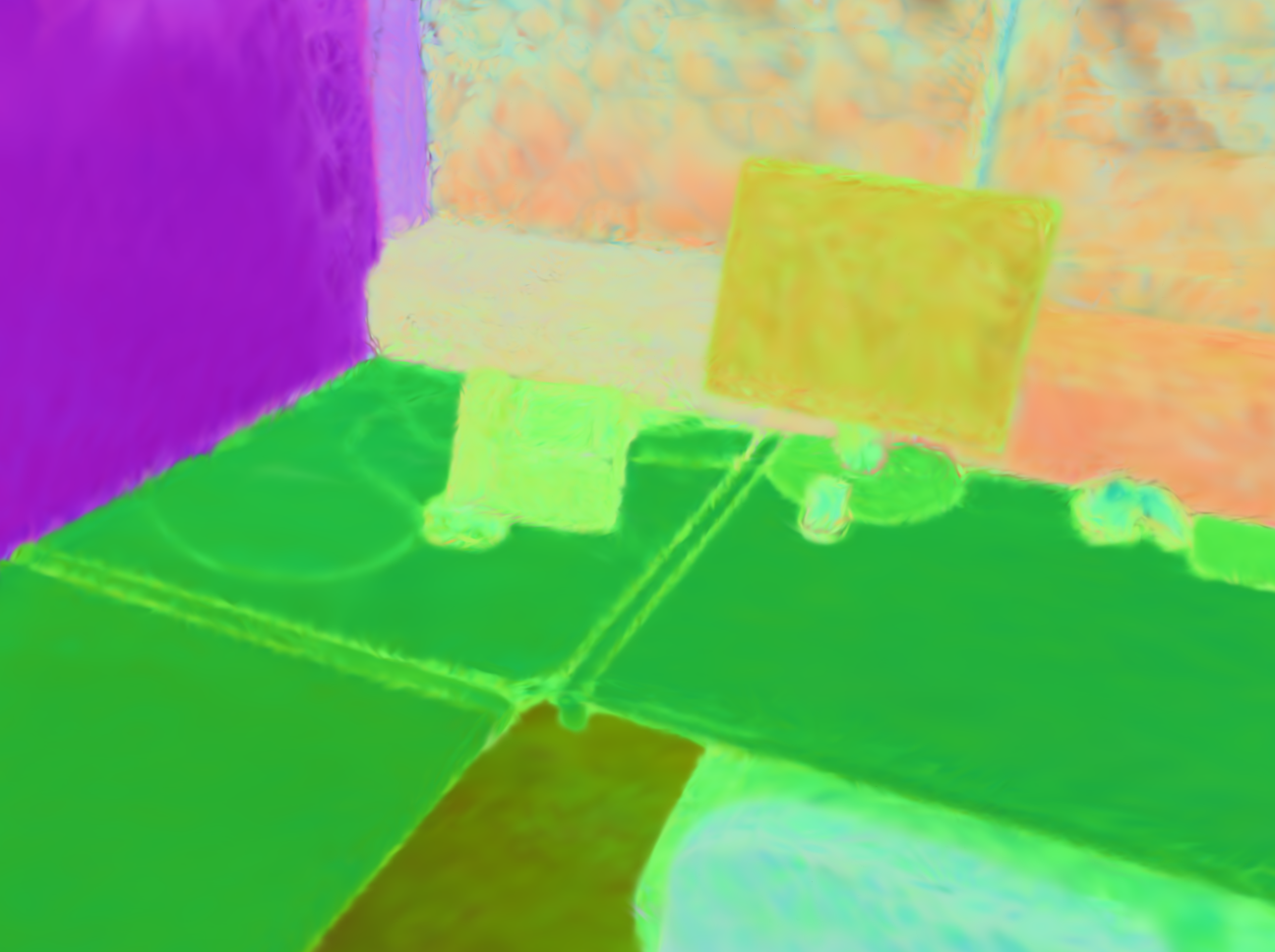}&
\includegraphics[width=.33\linewidth]{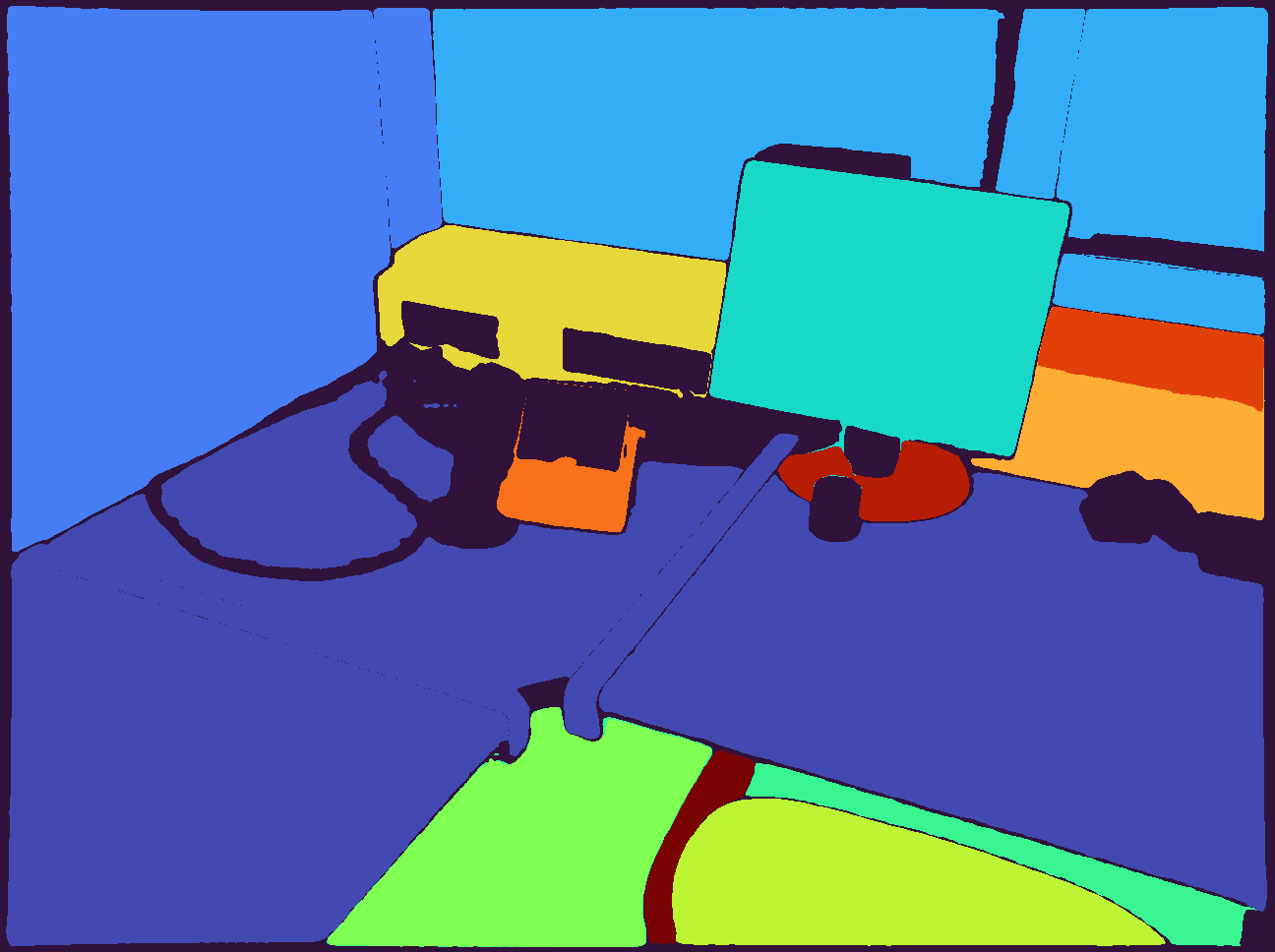}&
\includegraphics[width=.33\linewidth]{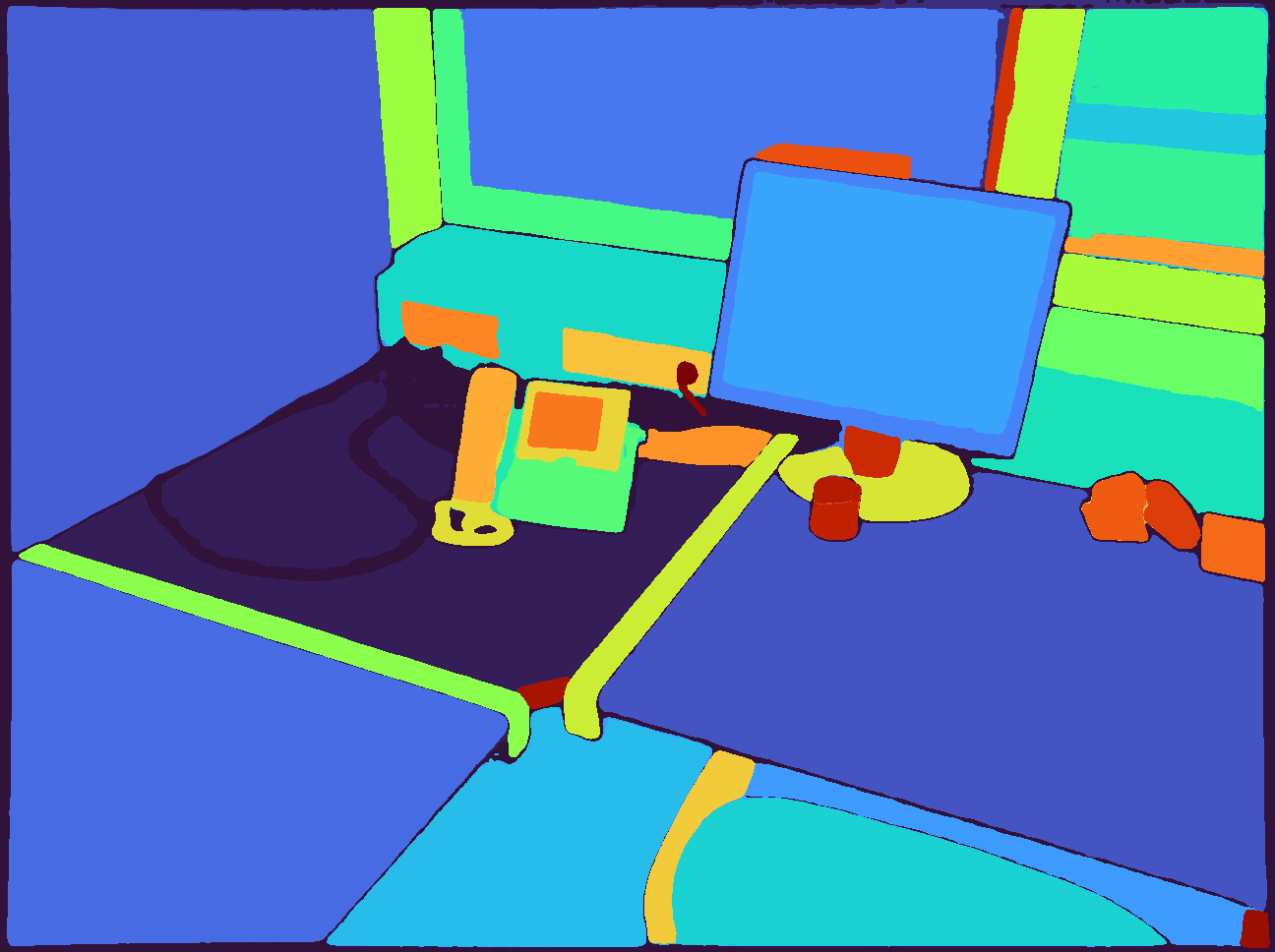}\\

\includegraphics[width=.33\linewidth]{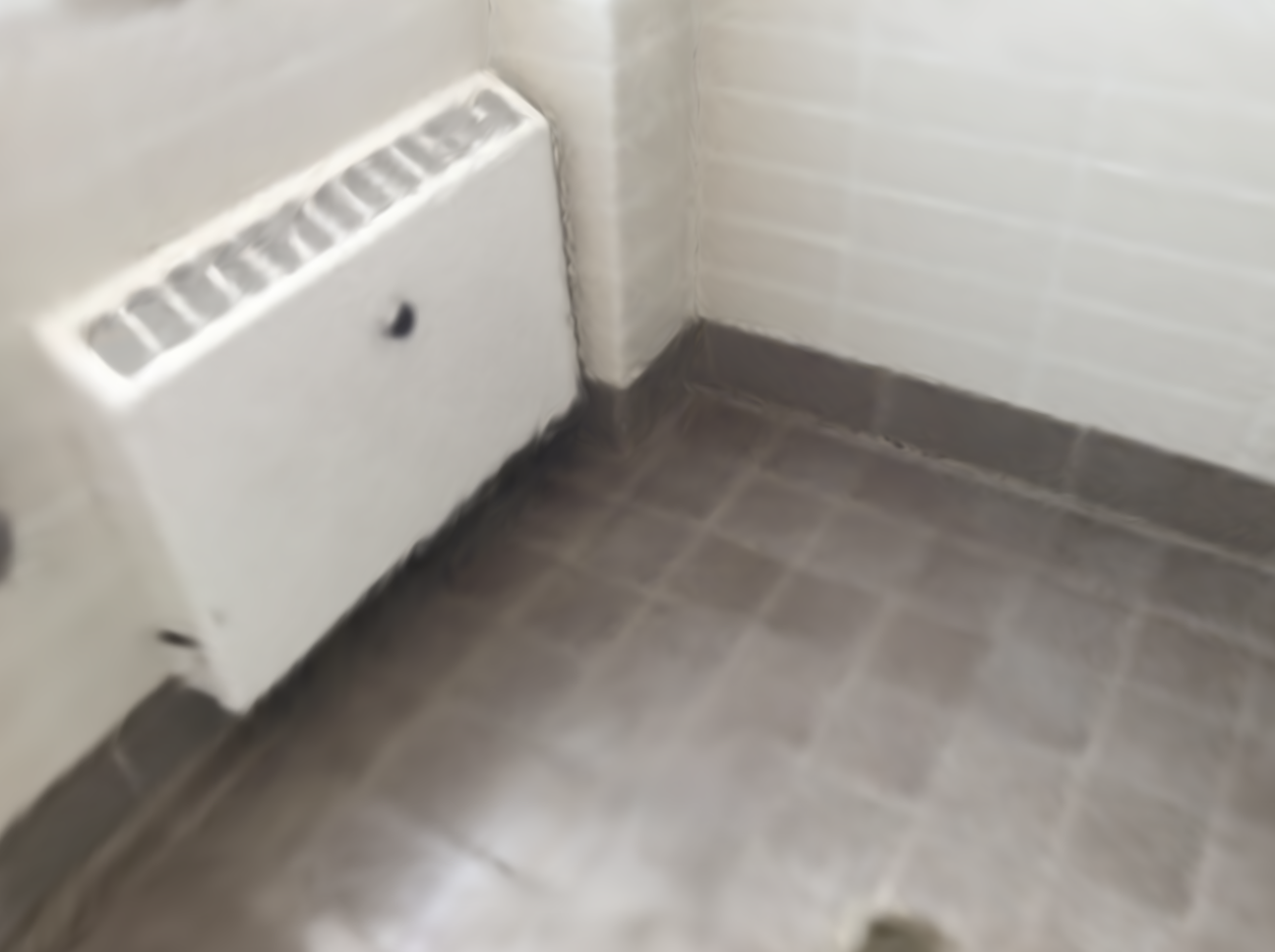}&
\includegraphics[width=.33\linewidth]{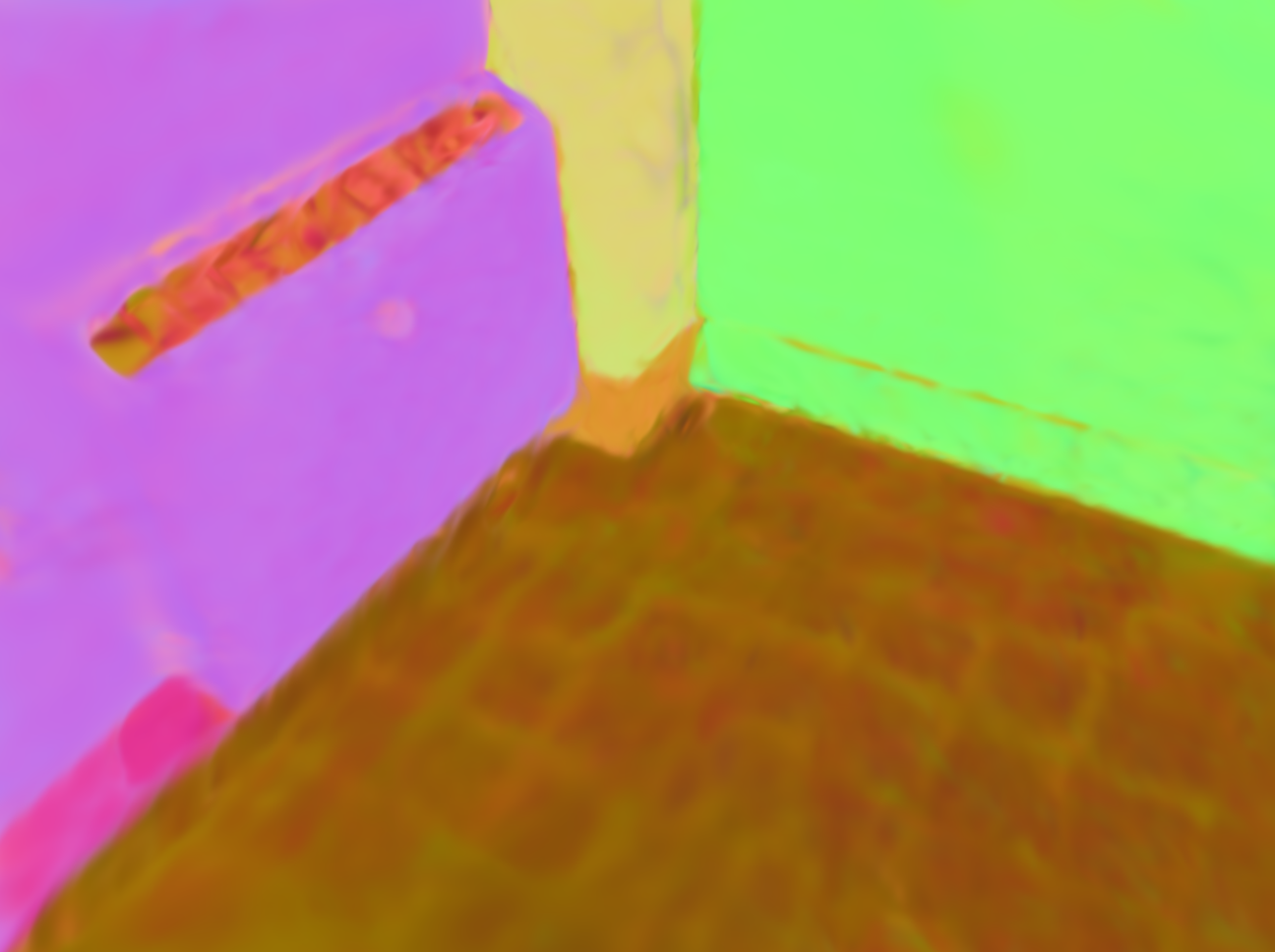}&
\includegraphics[width=.33\linewidth]{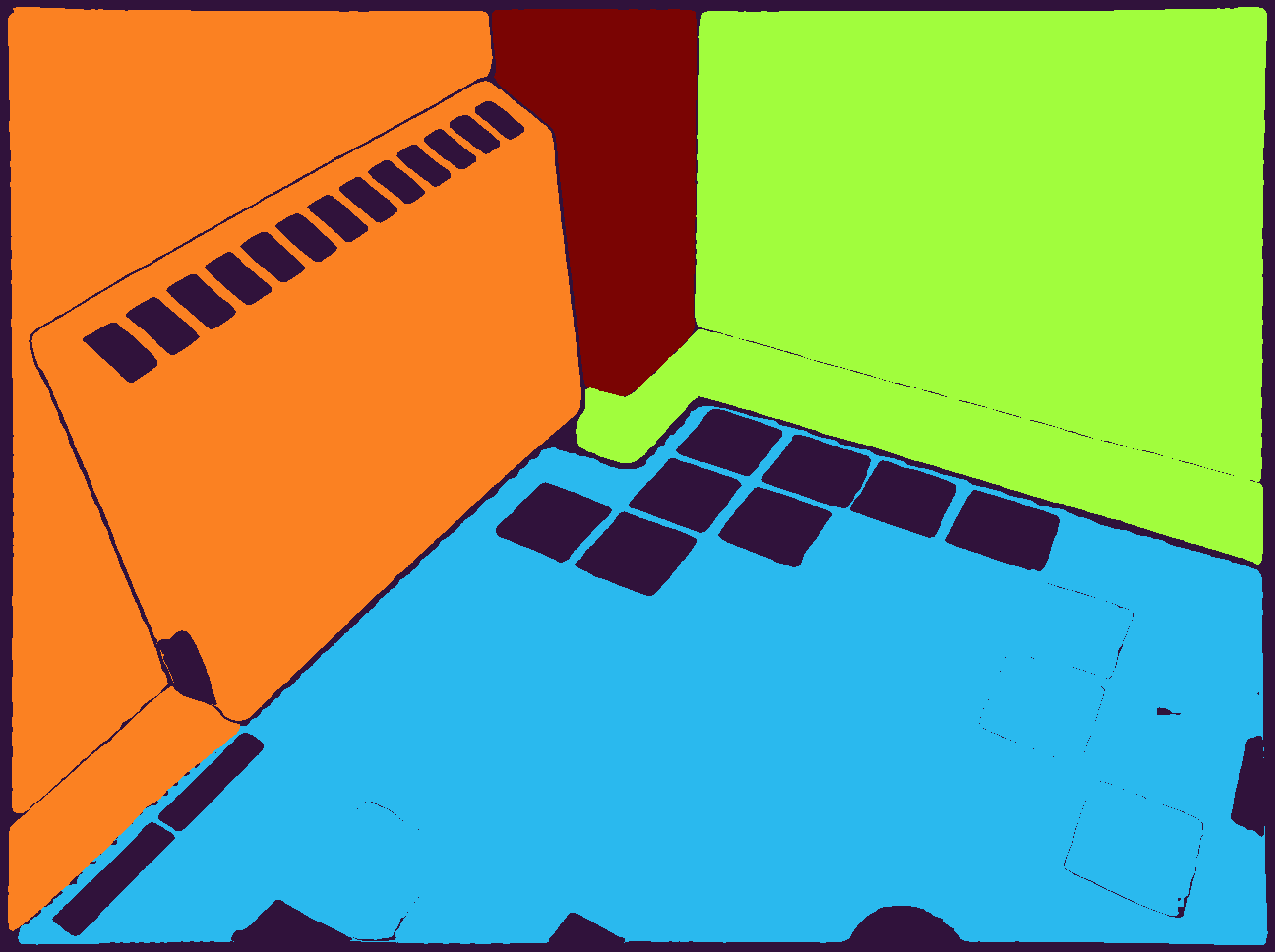}&
\includegraphics[width=.33\linewidth]{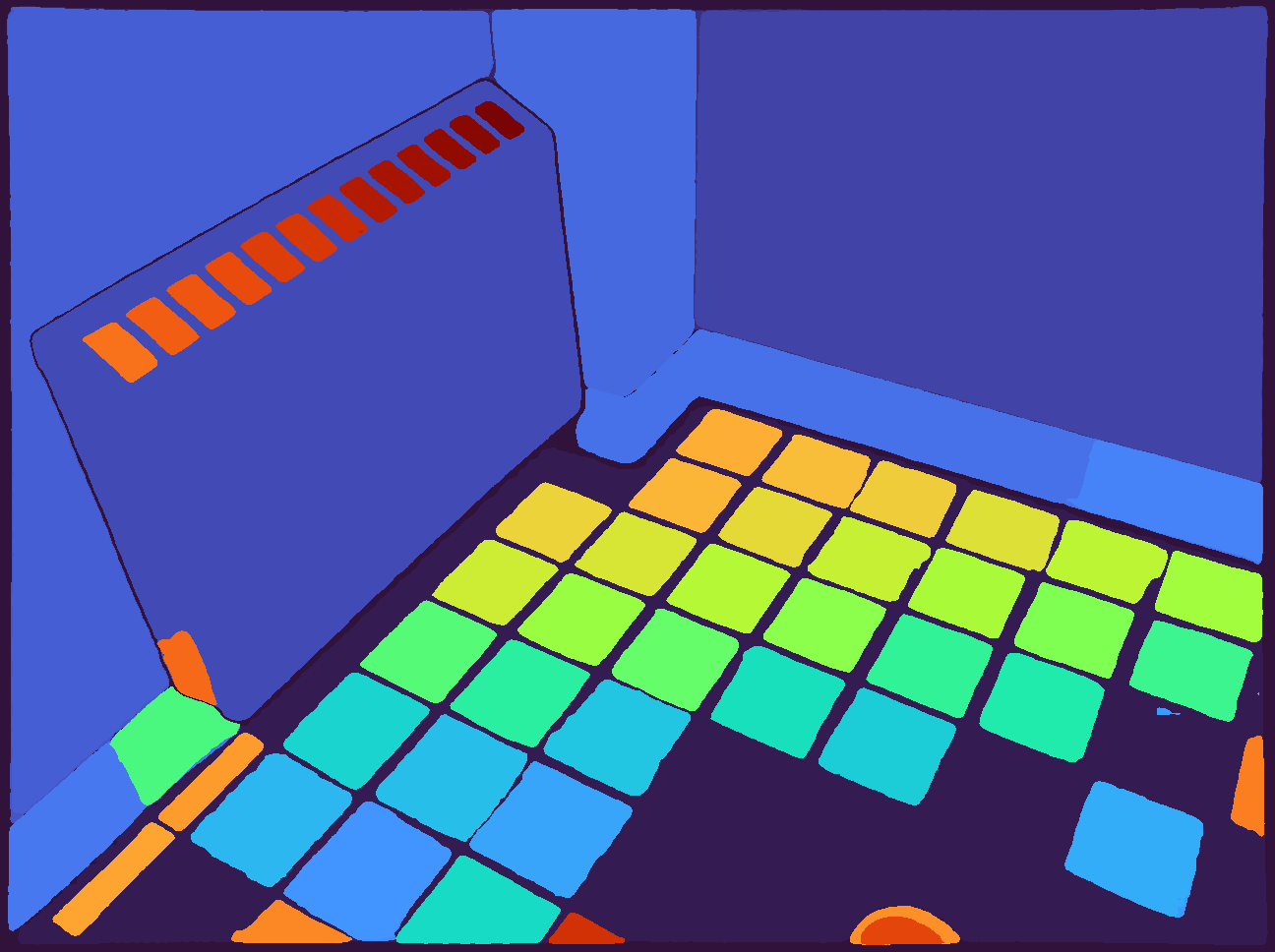}\\

\includegraphics[width=.33\linewidth]{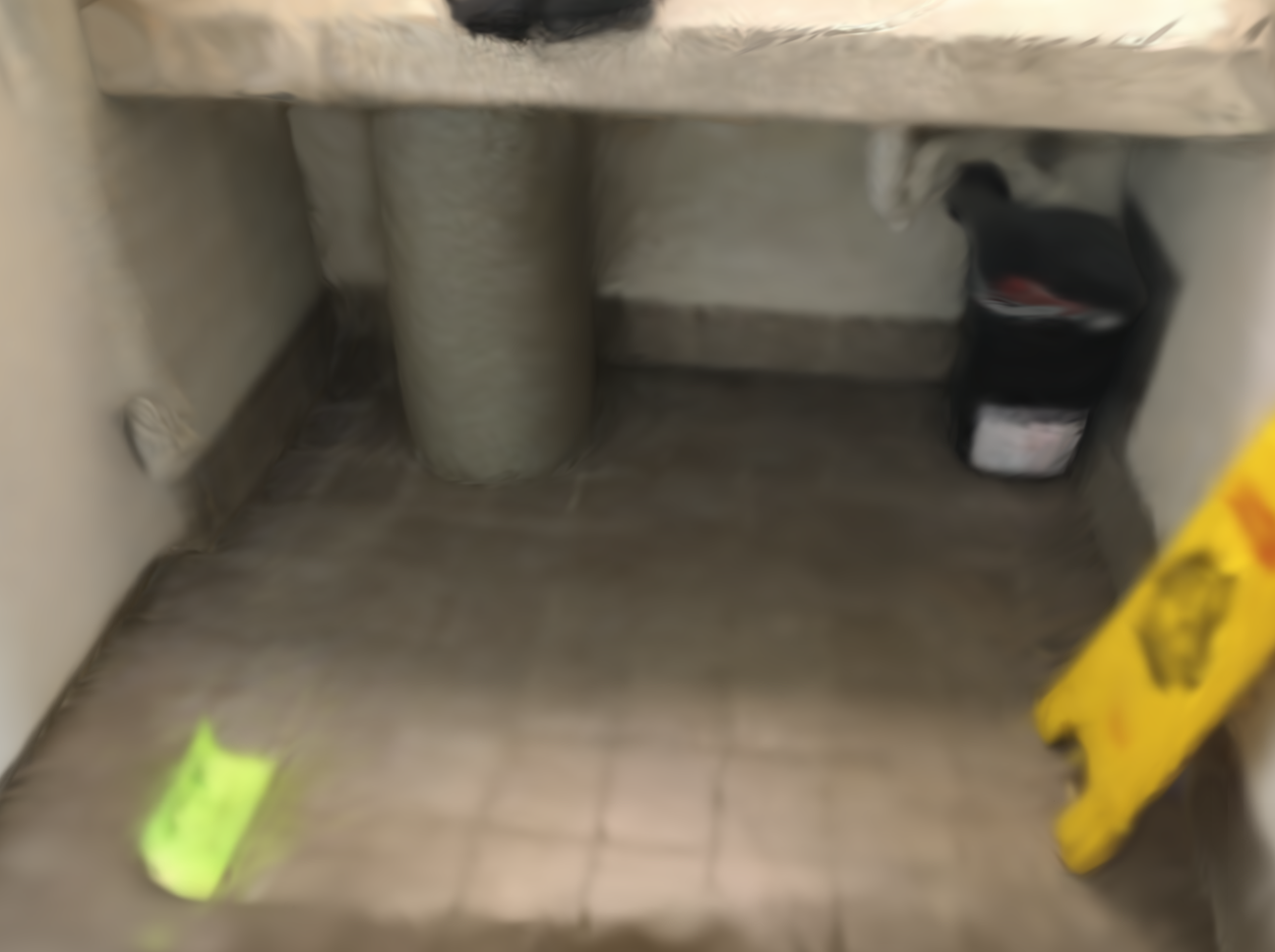}&
\includegraphics[width=.33\linewidth]{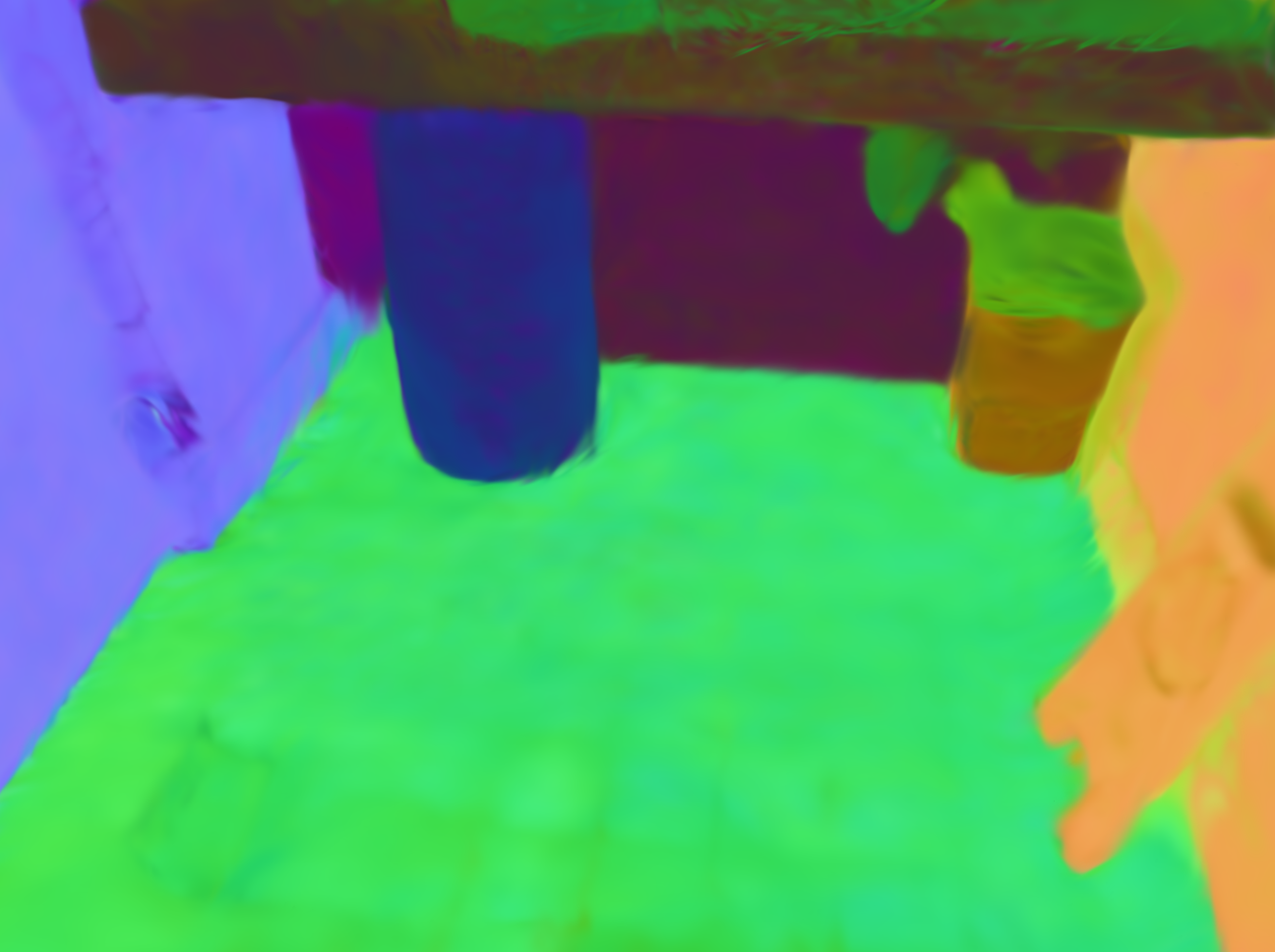}&
\includegraphics[width=.33\linewidth]{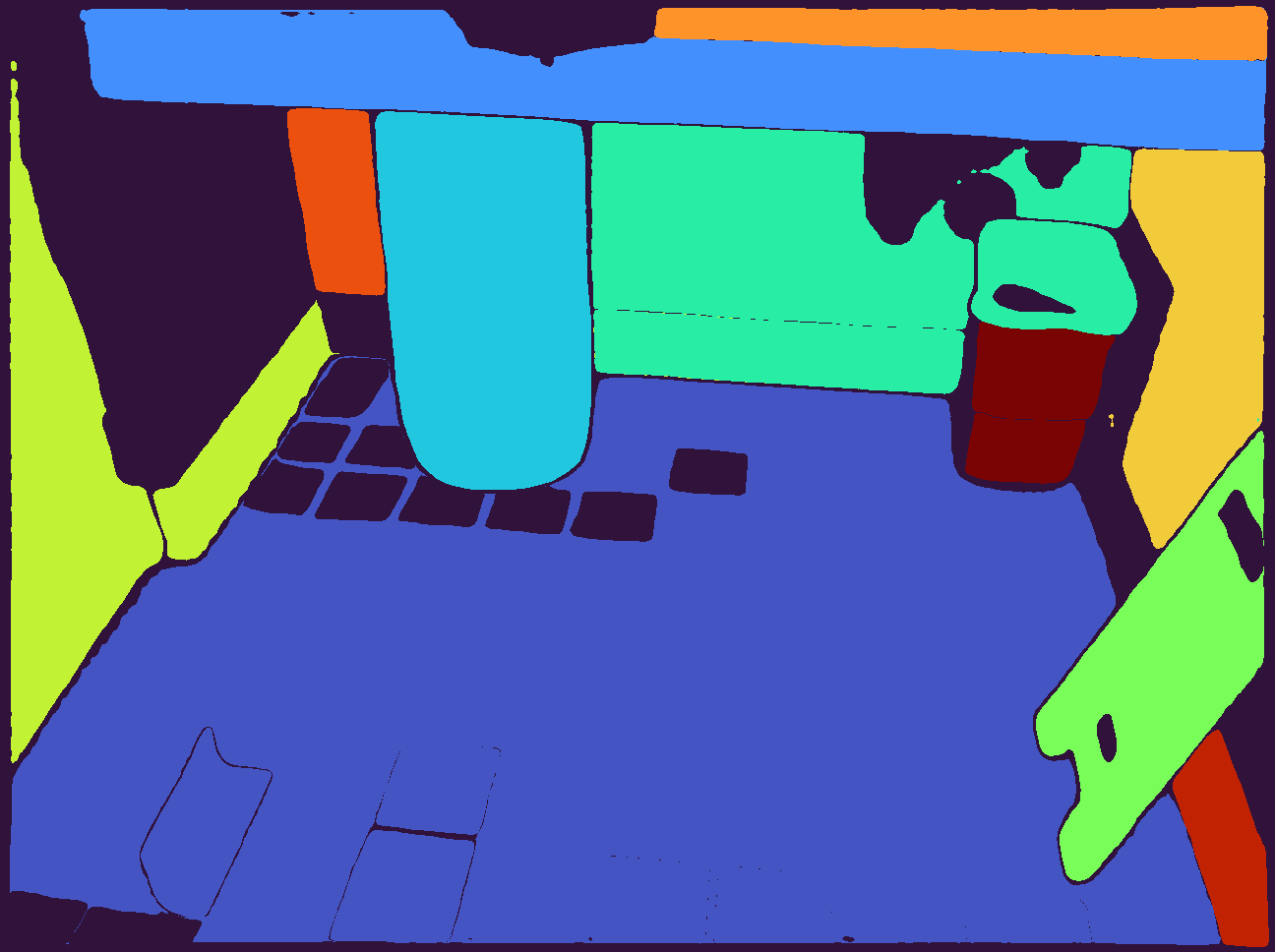}&
\includegraphics[width=.33\linewidth]{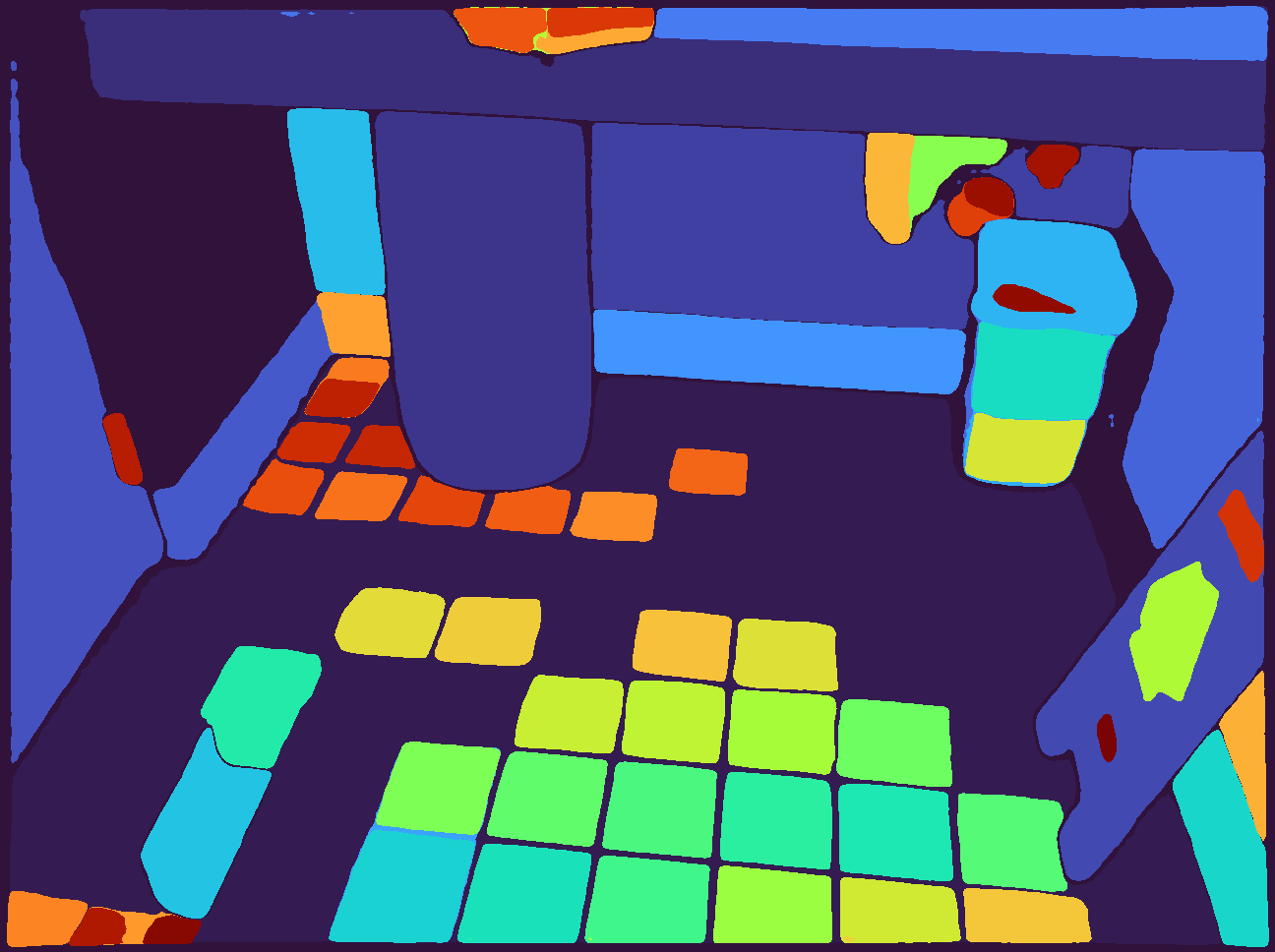}\\

\includegraphics[width=.33\linewidth]{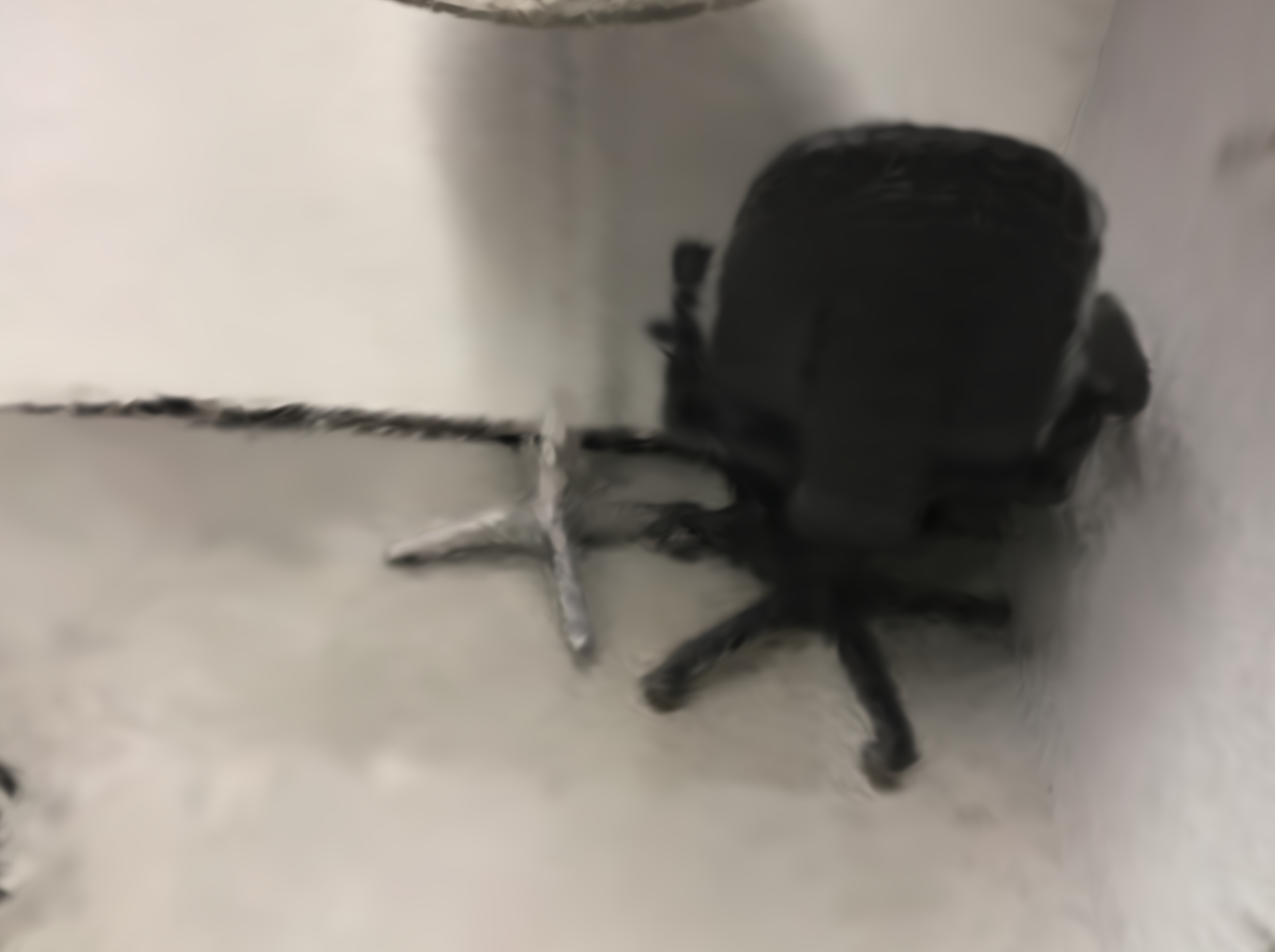}&
\includegraphics[width=.33\linewidth]{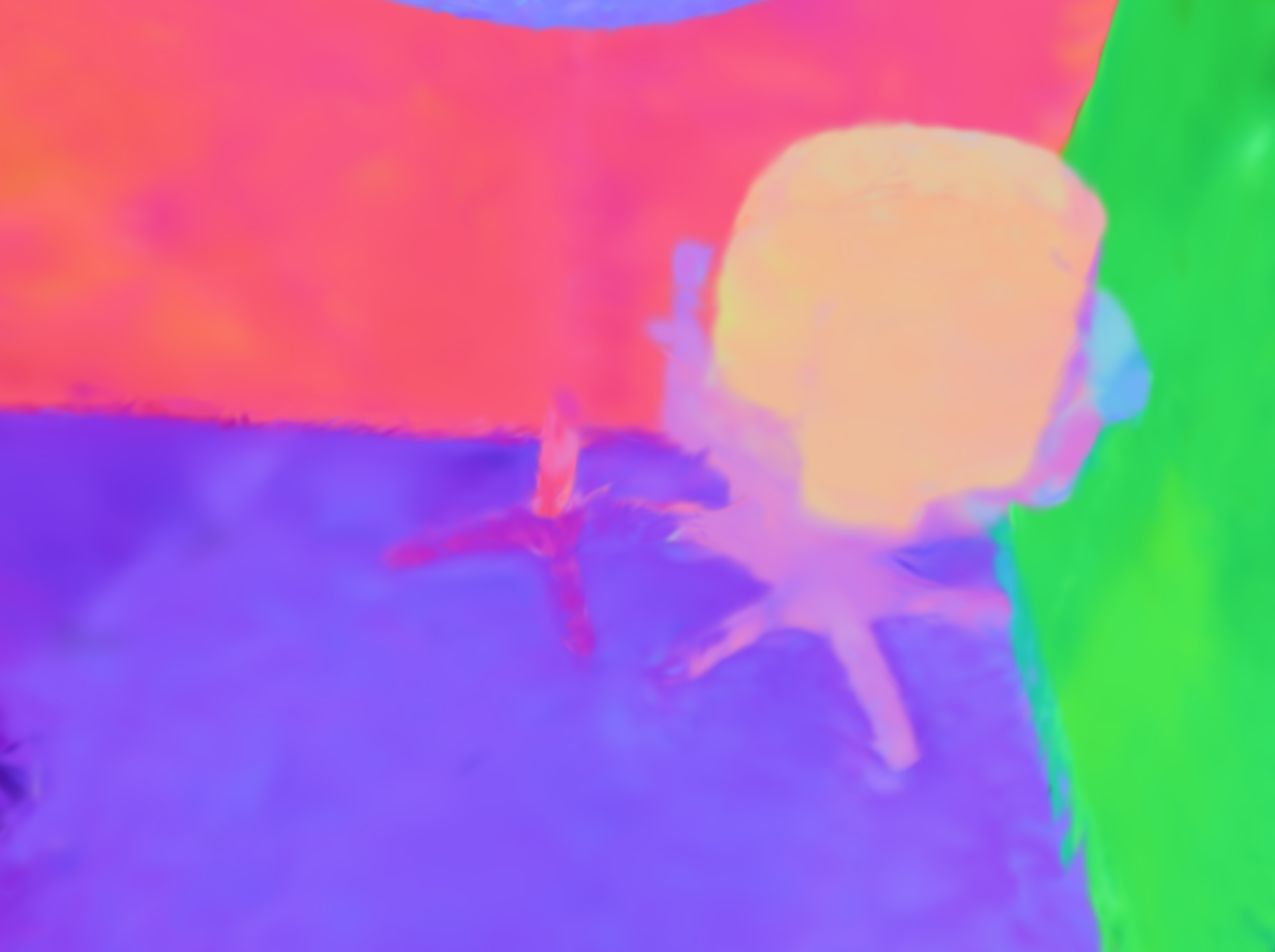}&
\includegraphics[width=.33\linewidth]{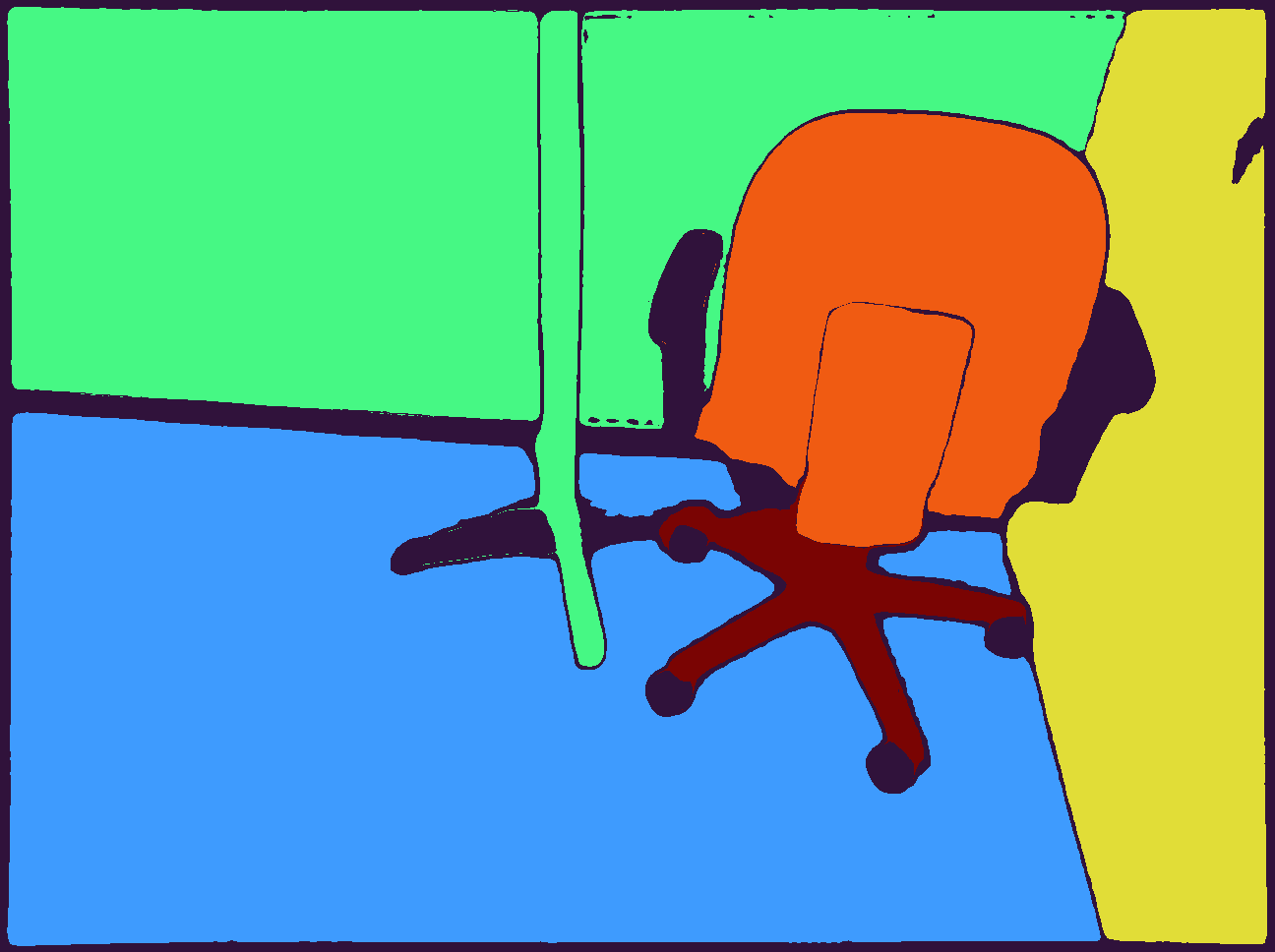}&
\includegraphics[width=.33\linewidth]{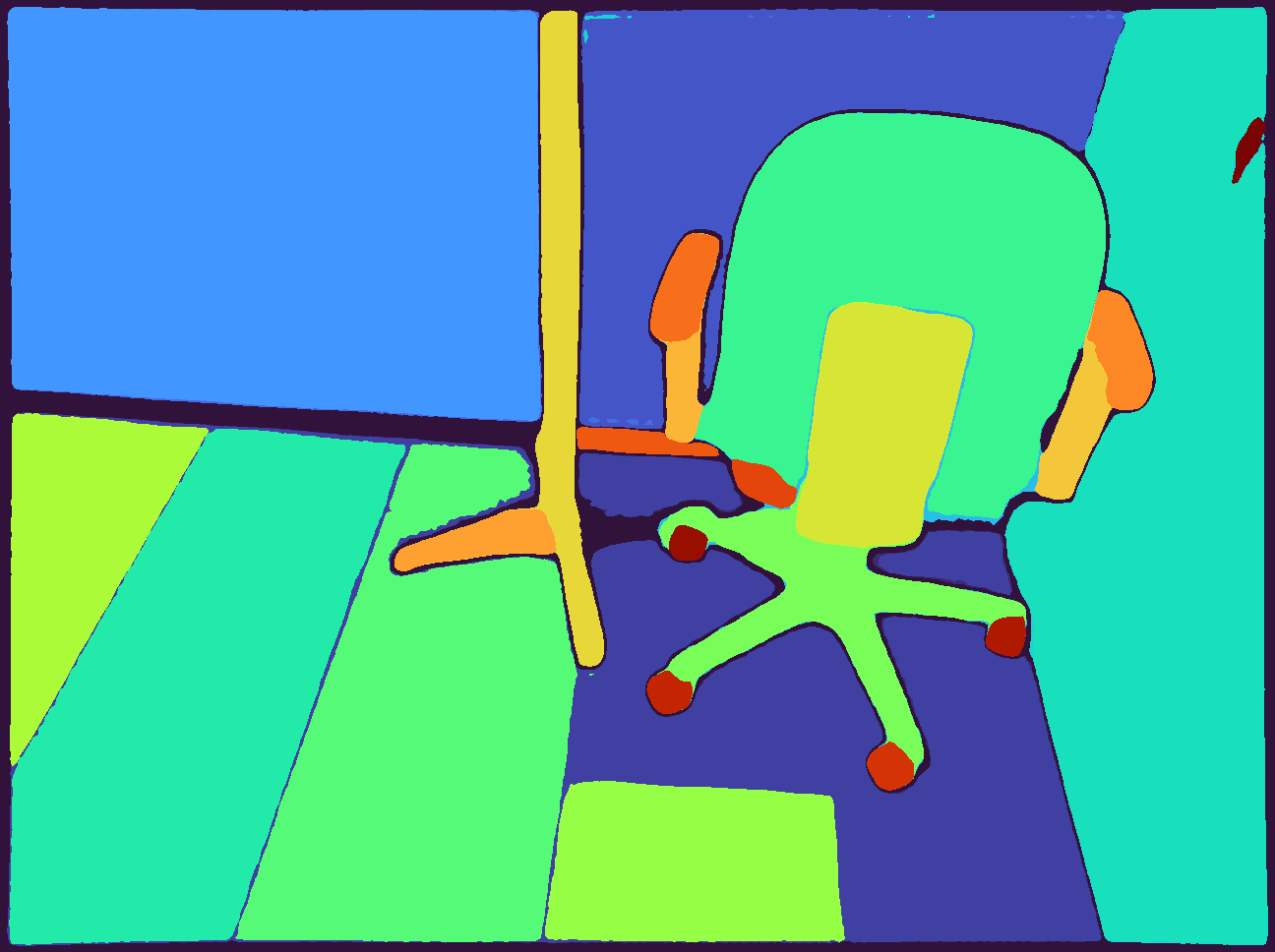}\\

(a) & (b) & (c) & (d)\\
\end{tabular}
\end{adjustbox}
\vspace{-8pt}
\caption{\small Visualizing the steps of generating plane descriptors:  (a) rendered color image, (b) rendered plane descriptors from 3D Gaussian field in camera view, (c) merged SAM masks by partitioning the Region Adjacency Graph, (d) masks by SAM.}
\label{fig:RAG}
\vspace{-10pt}
\end{figure}

Eq.~\ref{eq:linear_solver} optimizes the descriptors $\mathbf{z}$ for the Gaussians, taking advantage of the fast rendering of 3DGS. Note that the linear regression of Eq.~\ref{eq:linear_solver} is recomputed for each given camera view and the length $m$ of the target vector $\mathbf{y}_i$ can be variable in each image, depending on the number of segments in $\mathbf{Y}$. Figure~\ref{fig:RAG} provides visual examples of the learned descriptors and SAM segmentation masks.

Since the descriptors need to represent distinct 3D planes of the scene, we cannot directly use the SAM segments as the label $\mathbf{Y}$ as there can be multiple segments on the same plane. 2D segments belonging to the same planar surface should be merged. Since we do not have access to 3D or 2D plane annotations, we perform this merging using a \textbf{Region Adjacency Graph (RAG)}. 
The nodes of RAG represent the segments given by SAM and the edges connect nodes whose corresponding regions are adjacent in the image. 
We partition the RAG by cutting the edges that connect nodes from two different planes and keep the ones from the same plane connected. In order to do that, we use the normal vector of each node as well as a \emph{planar distance}.

The normals can be obtained by rendering as specified in Eq.~\ref{eq:normal_rendering}. However, partitioning the RAG solely based on the dissimilarity of normal vectors between neighboring nodes is insufficient; two nodes with similar normals may actually belong to two different planes, \eg, two planes at different heights in the scene. 
To resolve this ambiguity, we additionally assign a planar distance, $d_{p}$, to each node of the RAG. To compute the planar distance for each segment, we assume every pixel belongs to one planar surface in the scene. The corresponding 3D point $\mathbf{p} \in \mathbb{R}^3$ and the plane satisfy the point-normal equation: $\mathbf{n}\cdot\mathbf{p}+d_{p}=0$, where $\mathbf{n}=(n_1,n_2,n_3)^\top$ is the normal vector. As such, planar distance $d_{p}$ at pixel position $(u,v)$ in the image is obtained as follows:
\begin{equation}\label{eq:d_p}
d_{p}=d_{(u,v)}\cdot\bigl(\frac{n_1}{f_x}(u_0-u)+\frac{n_2}{f_y}(v_0-v)-n_3\bigr),
\end{equation}
where $(f_x, f_y)$ are the x and y focal lengths of the pinhole camera, and $(u_0, v_0)$ is the principal point, and the depth $d_{(u,v)}$ is computed by rendering the 3D Gaussian field.  

The average normal and planar distance values within each SAM segment is assigned to the corresponding node in the RAG. Finally, by thresholding these values, we can cut the edges accordingly in the RAG. Figure~\ref{fig:RAG}~(c) shows visual examples of merging SAM segments by partitioning the RAG.

\subsection{Local Planar Alignment of 3DGS}\label{sec:lpa}
\vspace{-4pt}
The original 3DGS method~\cite{kerbl20233d} excels in photo-realistic novel view synthesis through volume rendering and alpha-blending technique. However, it operates without any explicit constraints on the spatial arrangement of the Gaussian primitives, which makes it not directly applicable for learning geometric features. More specifically, the alpha-blending integration of Gaussians (as described in Eq.~\ref{eq:color_rendering}), which are depth-sorted relative to the camera, results in applying the 2D supervision to the overall integration of Gaussians, rather than to each instance individually. For example, occluded Gaussians receive weak supervision, leading to suboptimal parameter optimization. Although this is not critical for rendering RGB images, it poses challenges to learning 3D normal vectors and surface descriptors, which are additional parameters in our proposed PGS. To address this, we enforce the centers of Gaussians to lie exactly on the surfaces of objects. This can be achieved through the alignment process which involves projecting the centers of Gaussian on their local tangent planes. This requires first computing the K Nearest Neighbors (KNN) of Gaussians and then computing the covariance matrices of KNN Gaussian centers in the scene. Given the local covariance $3 \times 3$ matrices, the local tangent planes are specified by the two eigenvectors, corresponding to the two largest eigenvalues, of covariance matrices, using singular value decomposition. Estimating the KNN indices also allows us to apply a Laplacian smoothing on the learnt normal and descriptor features by averaging over the features of neighbouring Gaussian primitives.  

\vspace{-0pt}
\subsection{Holistic Separability of Gaussian Descriptors}\label{sec:meanshift}
\vspace{-4pt}
The minimization of the segmentation loss term in Eq.~\ref{eq:linear_solver} results in a discriminative representation of descriptors denoted as $\mathbf{z}$ in the current image. This representation helps identify Gaussians that belong to distinct 3D plane instances. In order to maintain a holistic separability of descriptors across all planar surfaces in the scene (including surfaces that have never been seen jointly in any camera view), a recurrent mean-shift update~\cite{kong2018recurrent} is applied to the entire Gaussian field, the matrix form of which is given as follows.\vspace{-5pt}
\begin{equation}\label{eq:mean-shift}
    \mathbf{Z} \leftarrow \mathbf{Z}\cdot(\eta \cdot \mathbf{K} \cdot \mathbf{D}^{-1} + (1-\eta)\cdot \mathbf{I}),\\[-5pt]
\end{equation}
where $\mathbf{K}=e^{(\gamma\cdot\mathbf{Z}^T\cdot\mathbf{Z})}$ is von Mises-Fisher (vMF) distribution of $\mathbf{z}$ on sphere and $\mathbf{D}=diag(\mathbf{K}^T\cdot \mathbf{I})$ is the diagonal matrix. $\eta$ is the rate of update and $\gamma$ is the kernel bandwidth which determines the smoothness of the kernel density estimation. 

\begin{figure}[t!]
\centering
\begin{adjustbox}{width=\columnwidth}
\begin{tabular}{ccc}
\includegraphics[width=.33\linewidth, trim={7cm 8cm 8cm 6cm}]{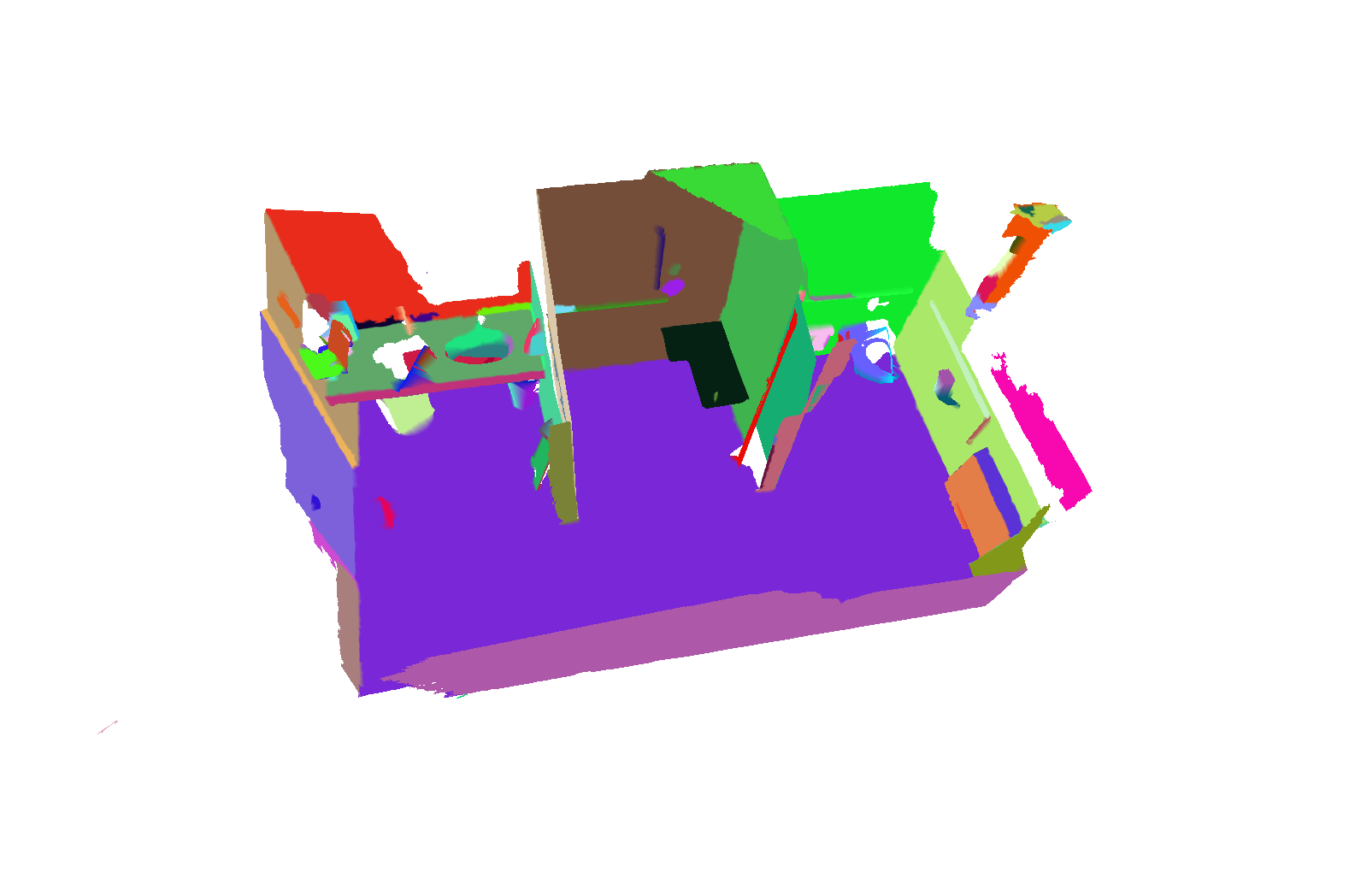}&
\includegraphics[width=.33\linewidth, trim={6cm 3cm 5.5cm 3cm}]{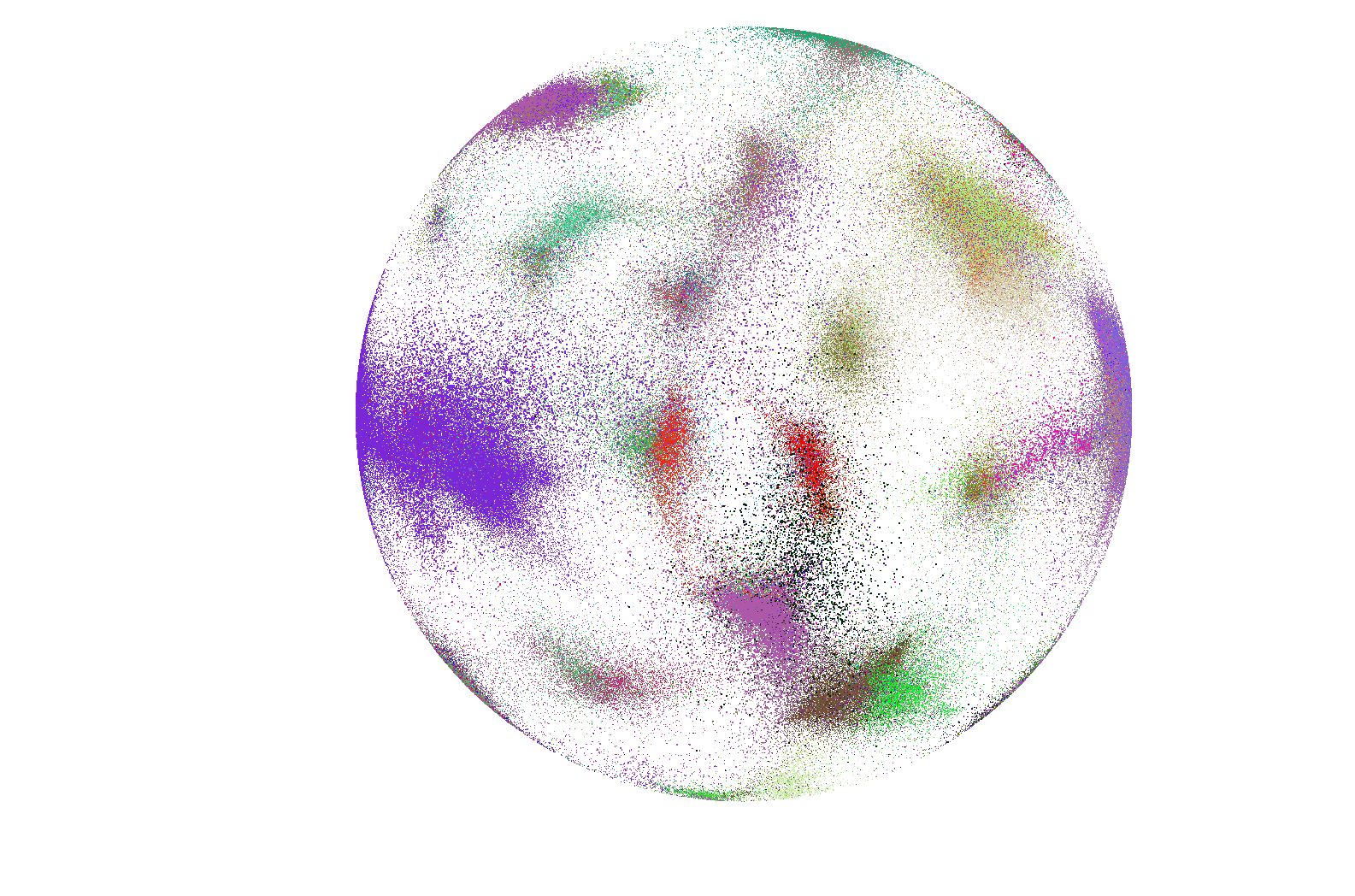}&
\includegraphics[width=.33\linewidth, trim={6cm 3cm 5.5cm 3cm}]{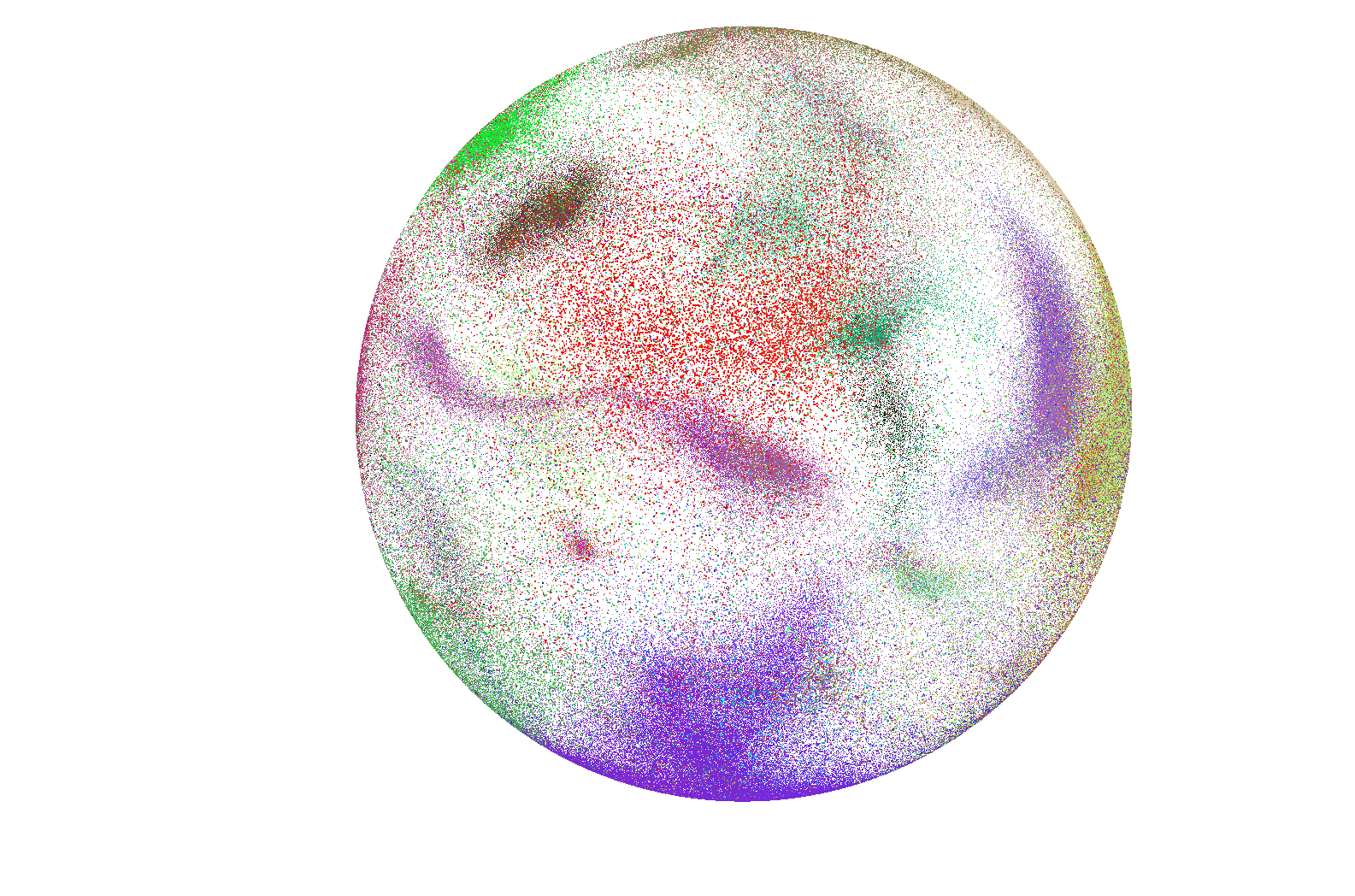}\\
(a) & (b) & (c)\\
\end{tabular}
\end{adjustbox}
\vspace{-5pt}
\caption{Effect of holistic separation. (a) Ground-truth plane labels of an example scene from ScanNet dataset, (b) learned descriptors with applying the holistic separability which results in a more compact and separable representation, (c) learned descriptors without the holistic constraint.}
\label{fig:spherical}
\vspace{5pt}
\end{figure}

By applying such updates, we improve separability of the descriptors of all the 3D plane instances. Figure~\ref{fig:spherical} visualize the impact of applying Eq.~\ref{eq:mean-shift} in the training of PGS. In practice we run updates on $\mathbf{z}$ as specified in Eq.~\ref{eq:mean-shift} for a few steps at every $N$ iterations in the optimization process. Since the number of Gaussians in the scene can be very large, we use an efficient way to  compute Eq.~\ref{eq:mean-shift}; more details on this are provided in the Appendix.

\vspace{-0pt}
\section{Experiments}
\vspace{-5pt}
We conduct experiments to evaluate the 3D plane instance segmentation of PGS, as well as compare with existing competitive approaches. We further perform ablation study to analyze various design choices in the proposed approach.

\textbf{Datasets.} We perform evaluation on 3D planar reconstruction benchmarks commonly used by prior works~\cite{xie2022planarrecon, zanjani2024neural}, including ScanNetv2~\cite{dai2017scannet} and Replica\cite{replica19arxiv}. ScanNetv2 contains RGB videos taken by a mobile device from indoor scenes with the camera pose
information associated with each frame. We run our experiments on 10 scenes; 4 of them are the same as in~\cite{guo2022neural}. Replica is a synthetic dataset featuring a diverse set of indoor scenes. Each scene is equipped with high-quality geometry and photo-realistic textures, allowing one to render high-fidelity images from arbitrary camera poses. 

\textbf{Baselines.} There are only a few existing works that focus on learning-based multi-view 3D planar reconstruction. We compare PGS with two types of approaches: (1) specialized 3D planar reconstruction methods, \eg, PlanarRecon~\cite{xie2022planarrecon}, which is trained with 3D geometry and 3D plane supervisions, and NMF~\cite{zanjani2024neural} which is an optimization-based approach using depth and normal supervision, (2) dense 3D reconstruction methods like NeuralRecon\cite{sun2021neuralrecon} with 3D geometry supervision and 3DGS~\cite{kerbl20233d} with 2D RGB supervision, followed by Sequential RANSAC to extract planes~\cite{fischler1981random}. We denote them as NeuralRecon\texttt{++} and 3DGS\texttt{++}.

\textbf{Metrics.} Similar to prior works~\cite{liu2019planercnn, tan2021planetr, xie2022planarrecon, zanjani2024neural}, we evaluate the performance of 3D plane instance segmentation by measuring the Rand Index (RI), Variation of Information (VOI), and Segmentation Covering (SC).

\begin{figure*}[t!]
\centering
\begin{adjustbox}{width=\linewidth}
\begin{tabular}{cccc}
\includegraphics[width=.33\linewidth]{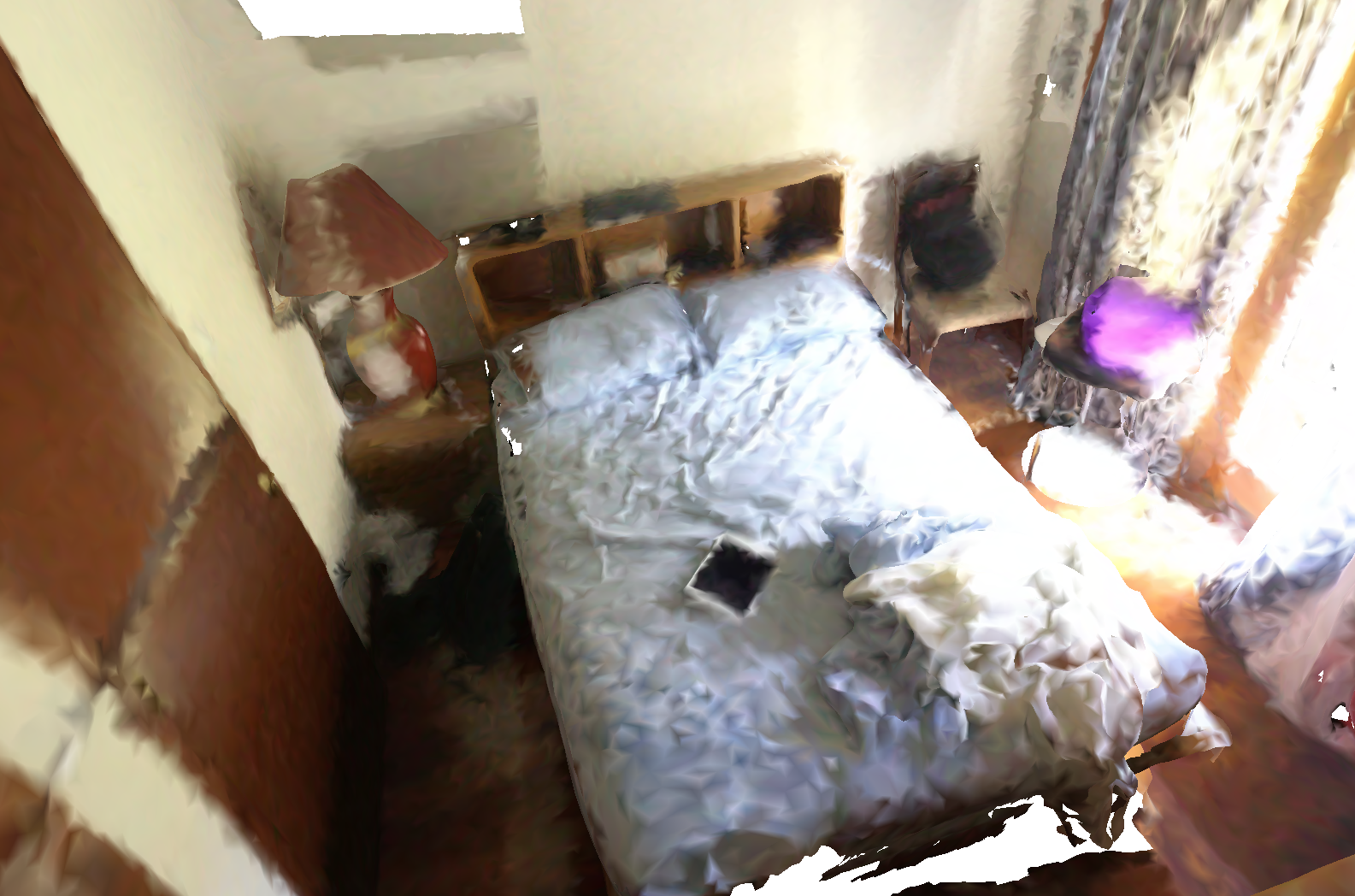}&
\includegraphics[width=.33\linewidth]{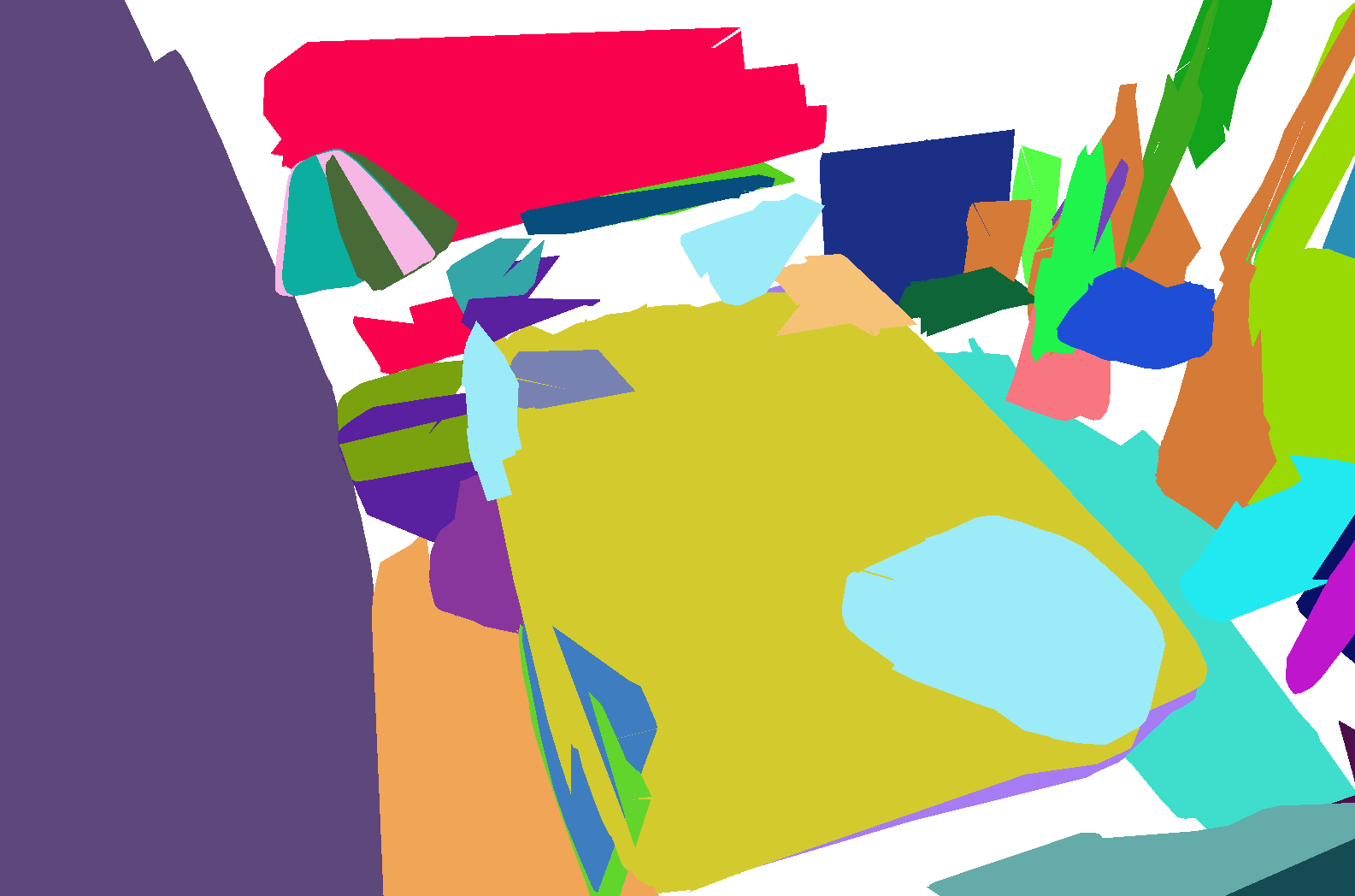}&
\includegraphics[width=.33\linewidth]{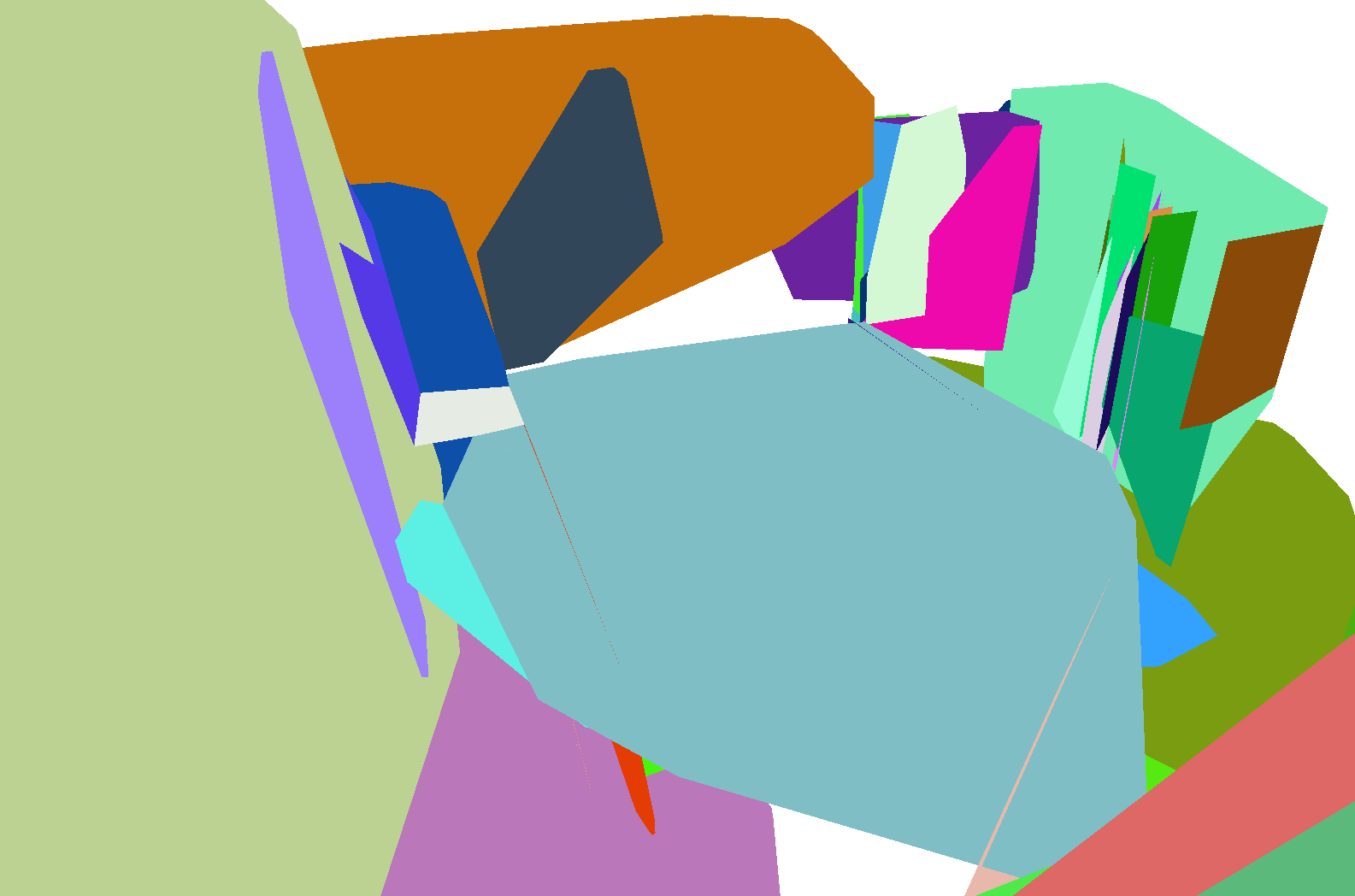}&
\includegraphics[width=.33\linewidth]{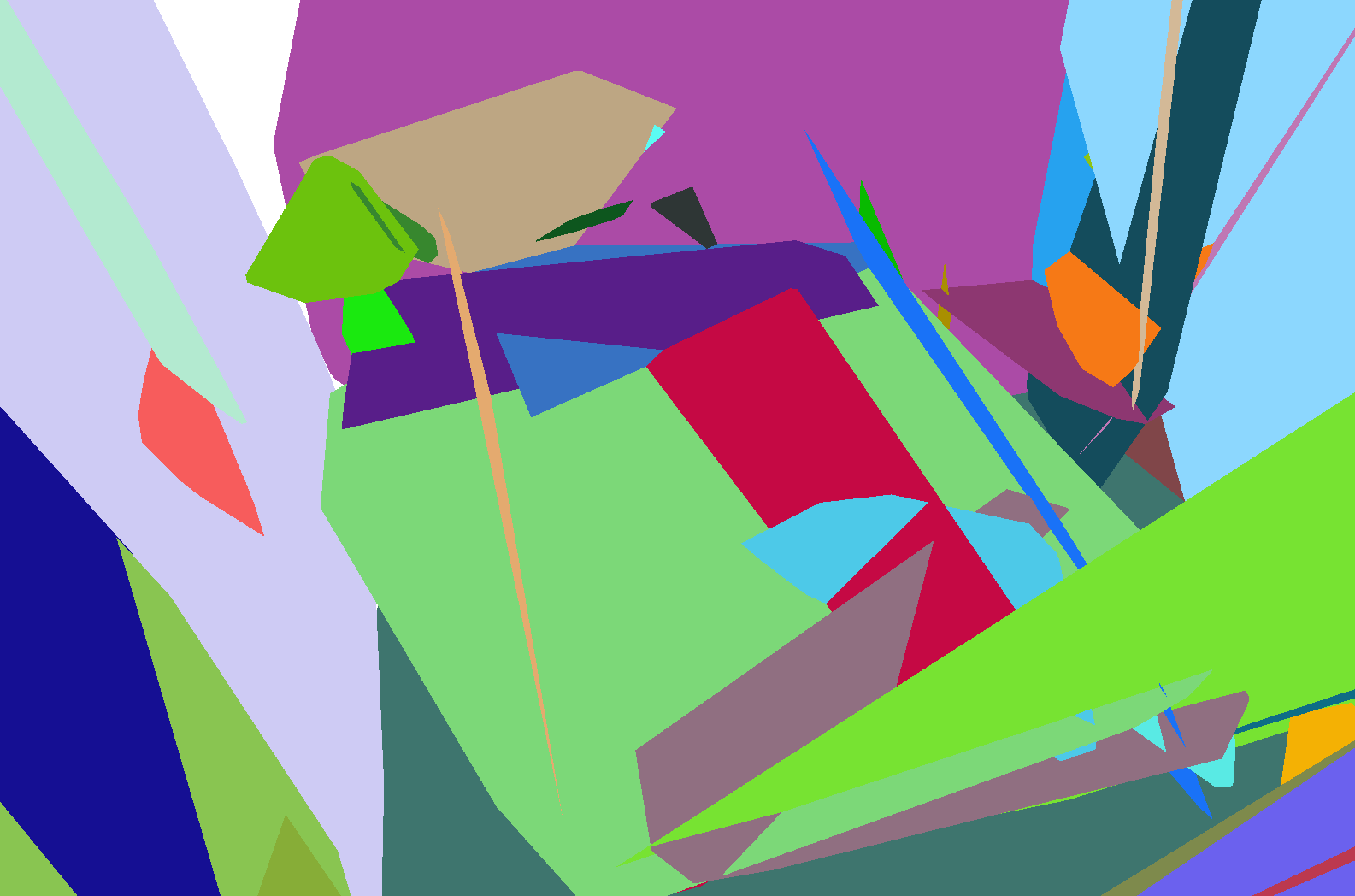}\\
\includegraphics[width=.33\linewidth]{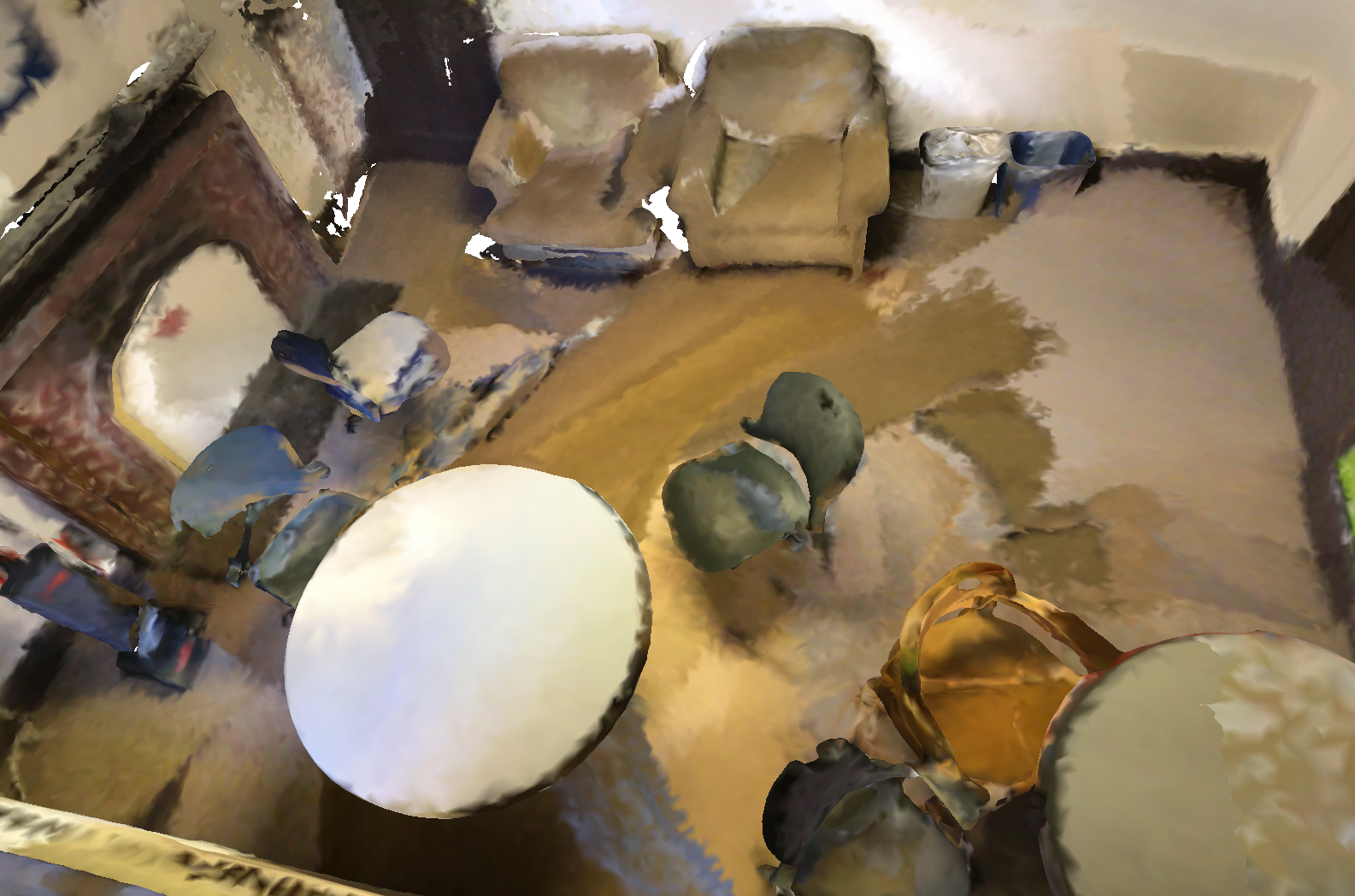}&
\includegraphics[width=.33\linewidth]{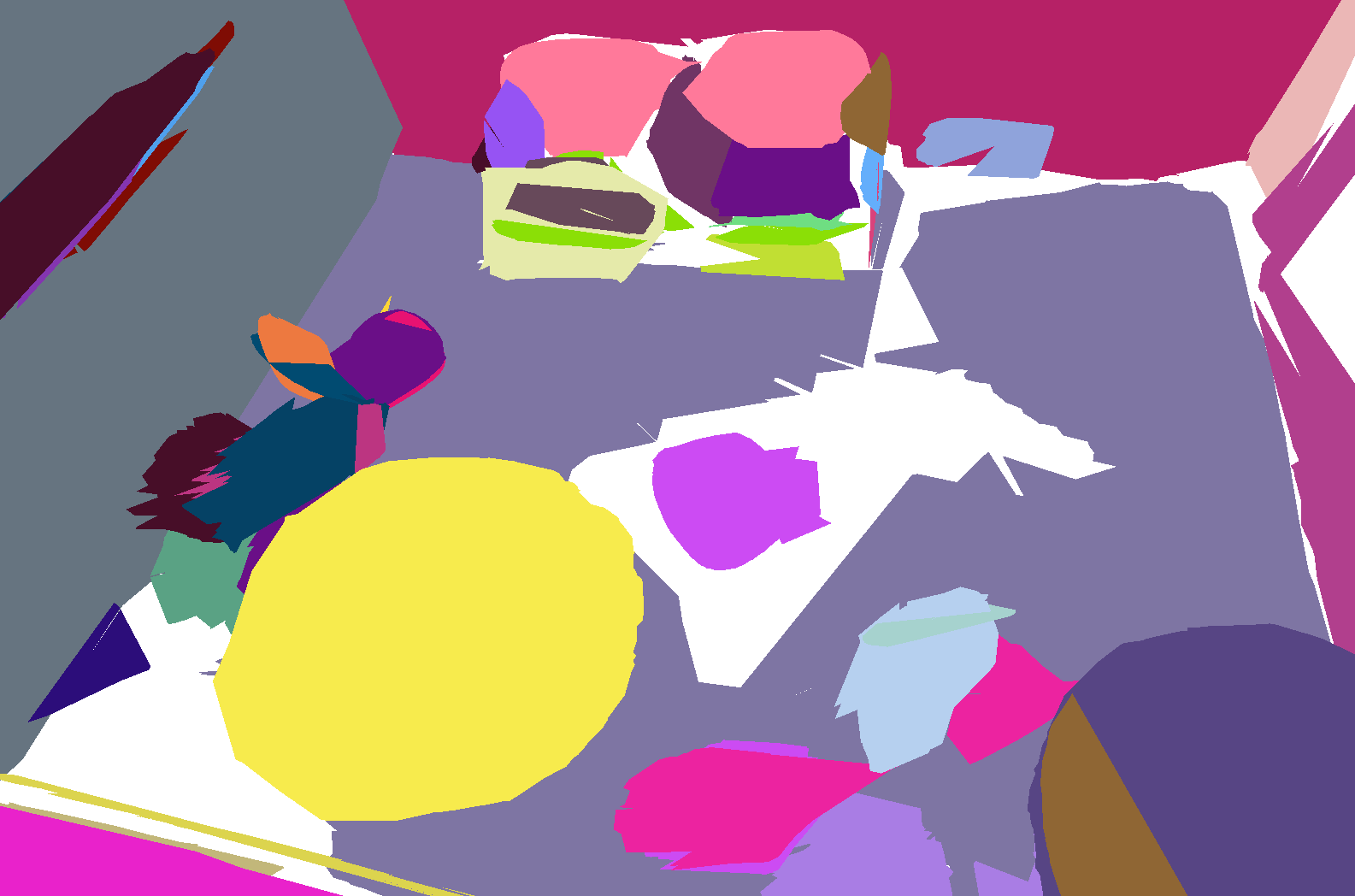}&
\includegraphics[width=.33\linewidth]{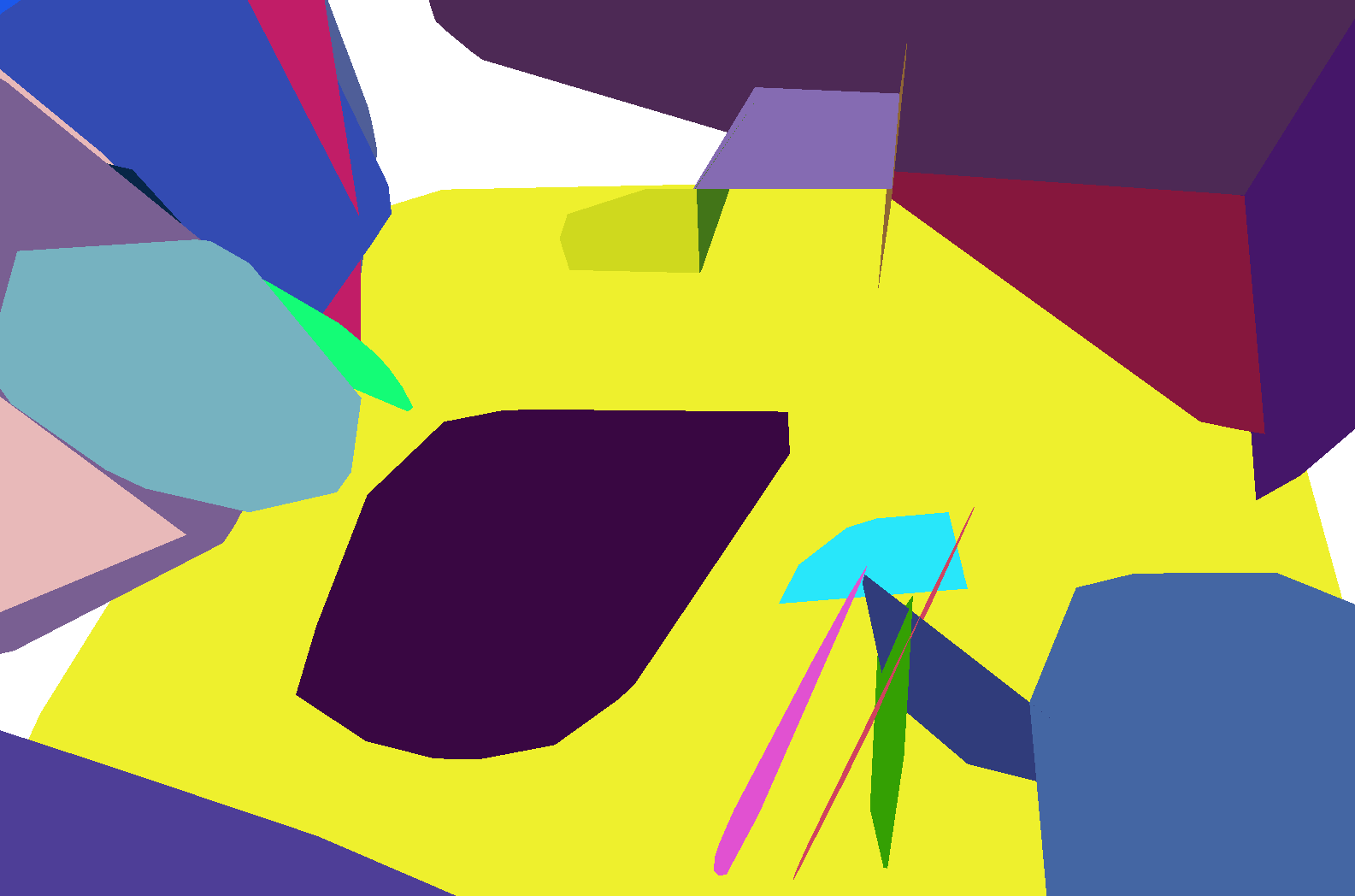}&
\includegraphics[width=.33\linewidth]{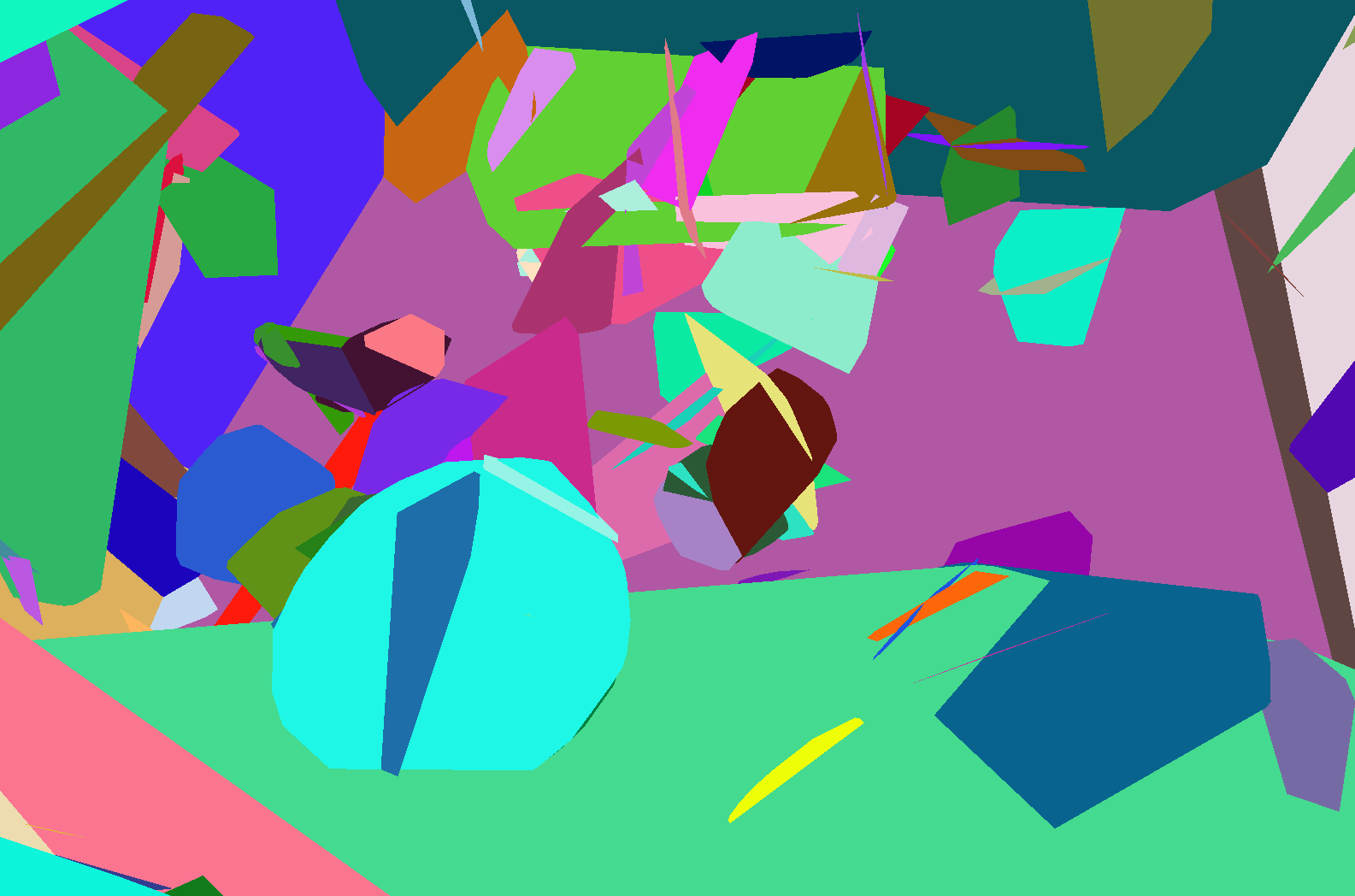}\\
\includegraphics[width=.33\linewidth]{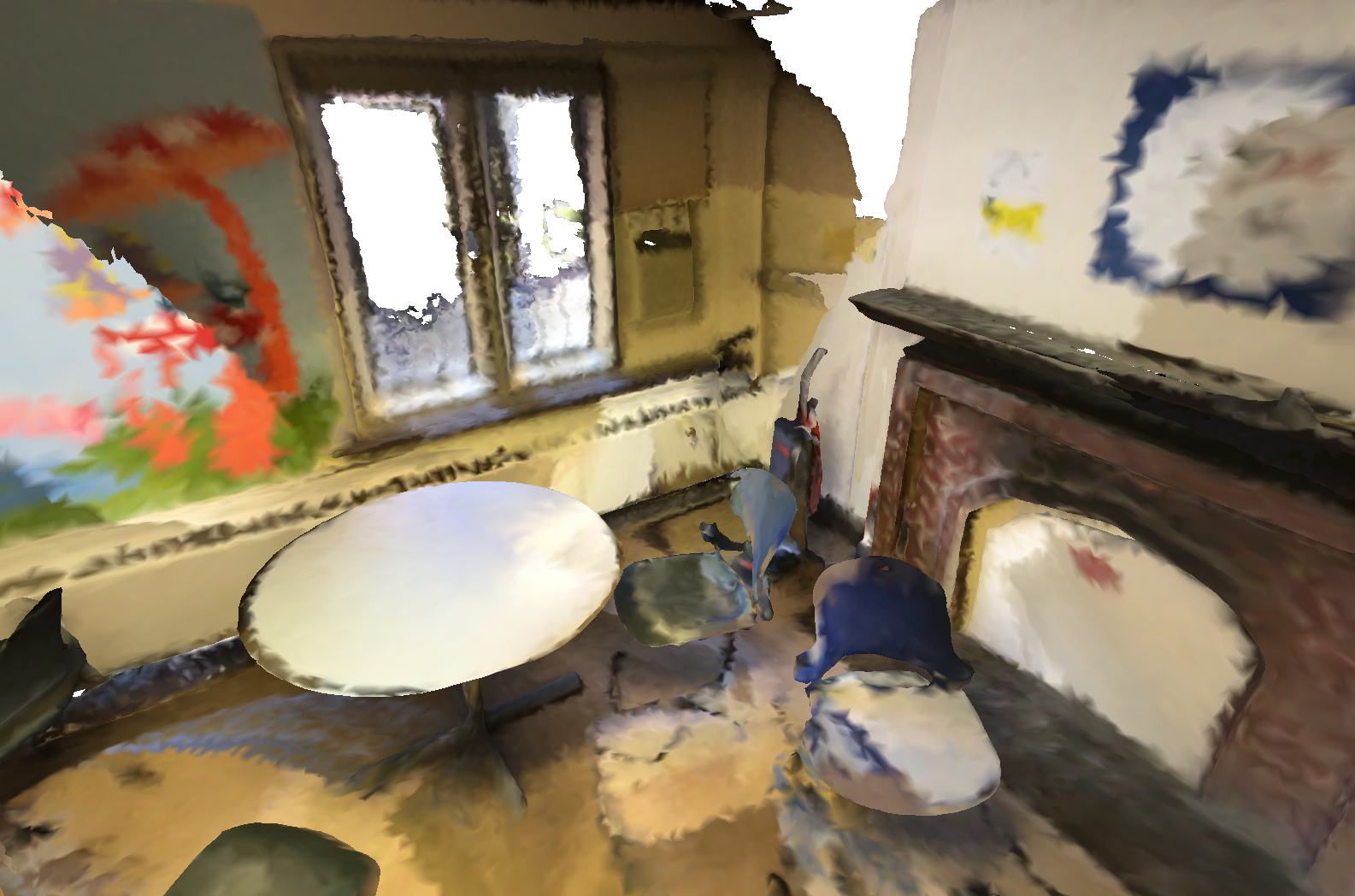}&
\includegraphics[width=.33\linewidth]{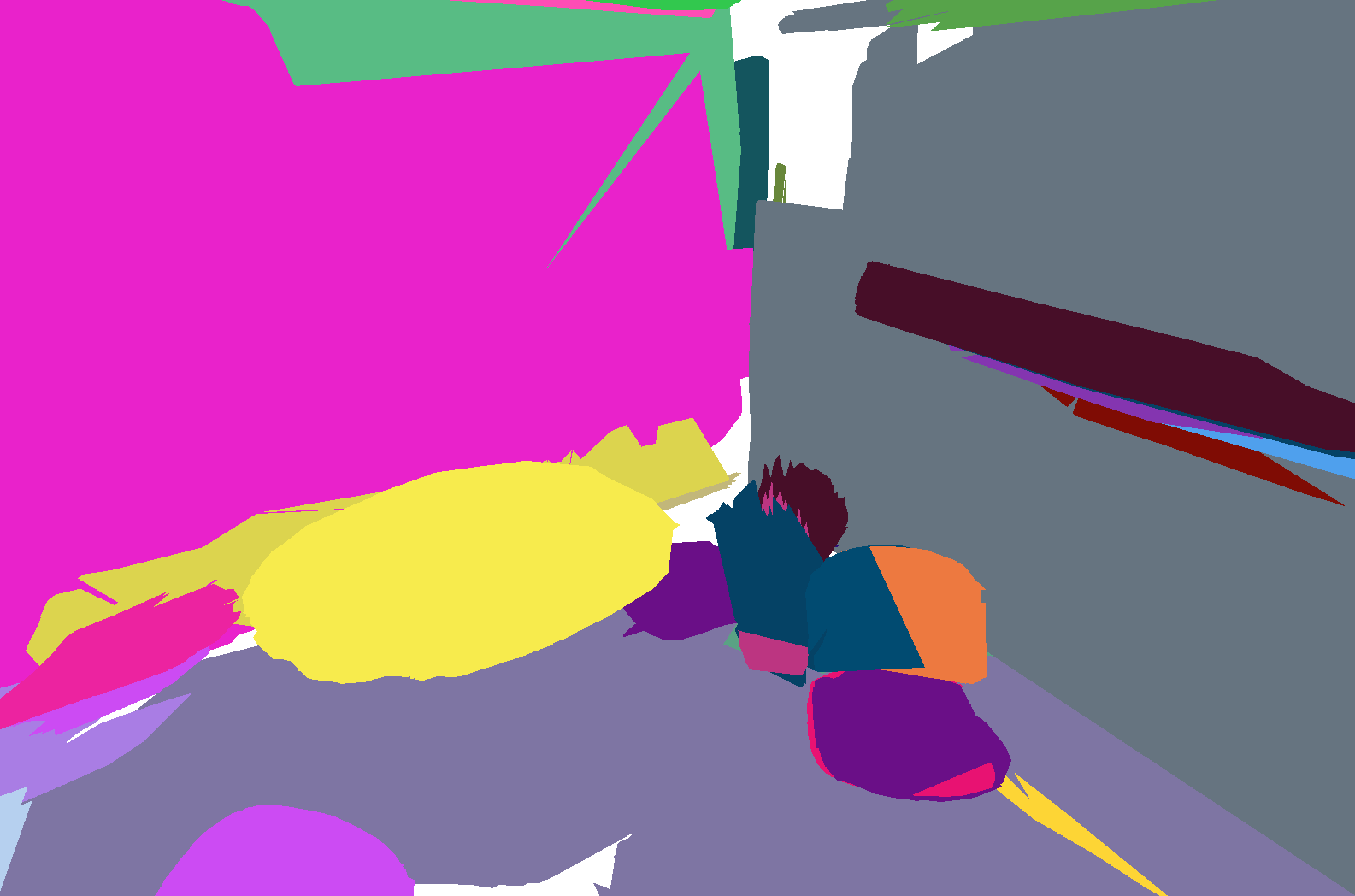}&
\includegraphics[width=.33\linewidth]{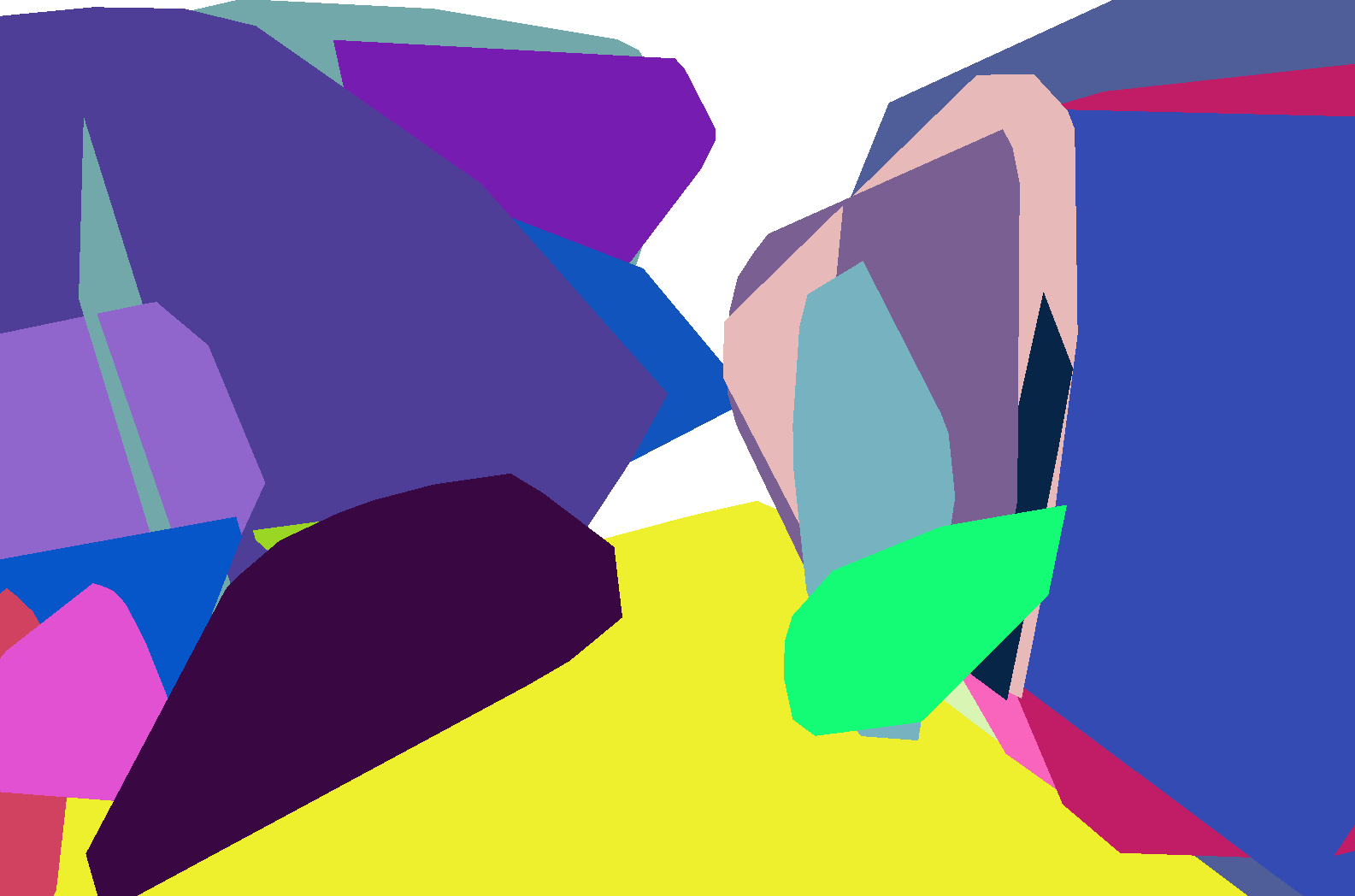}&
\includegraphics[width=.33\linewidth]{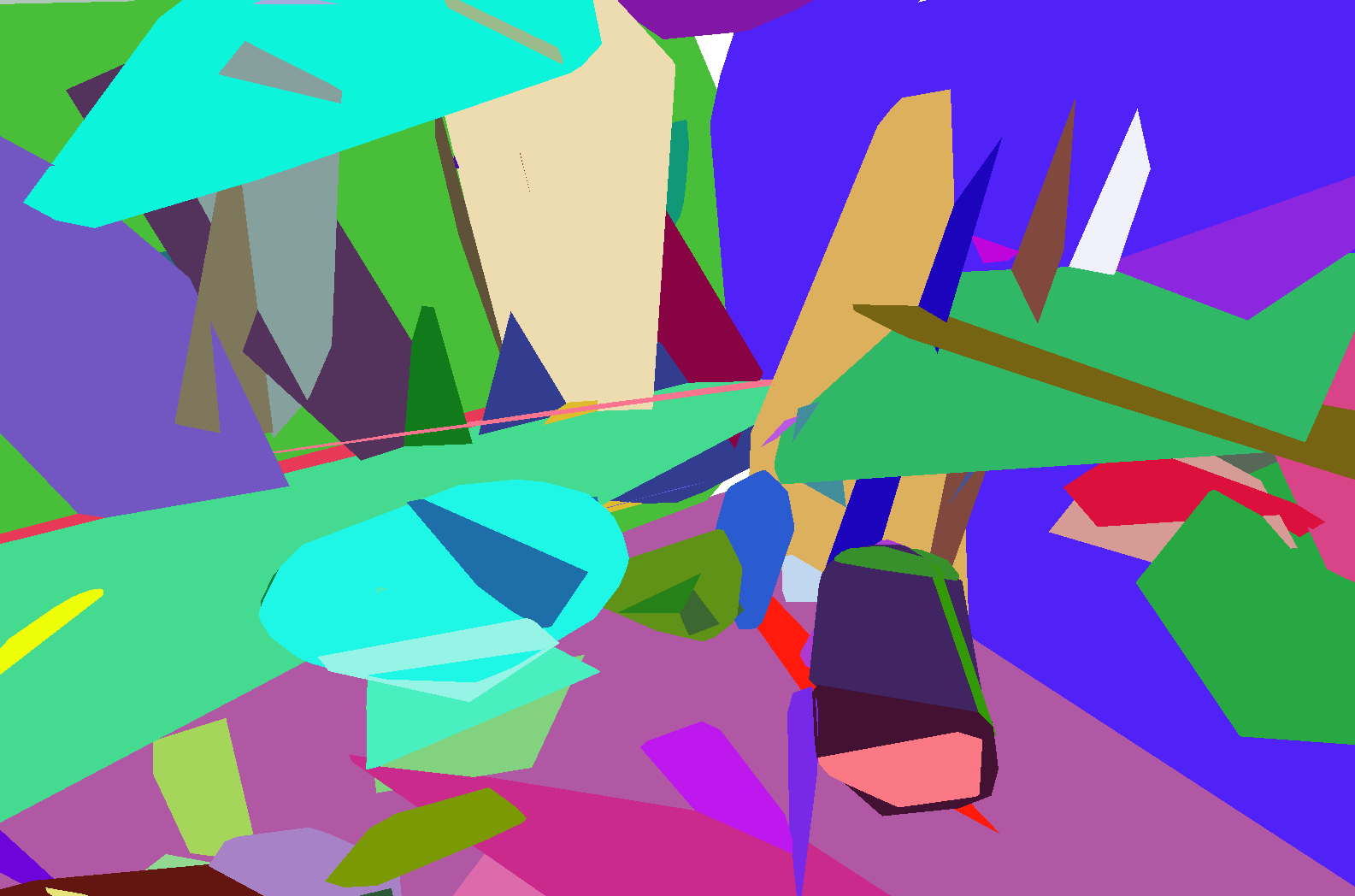}\\
\includegraphics[width=.33\linewidth]{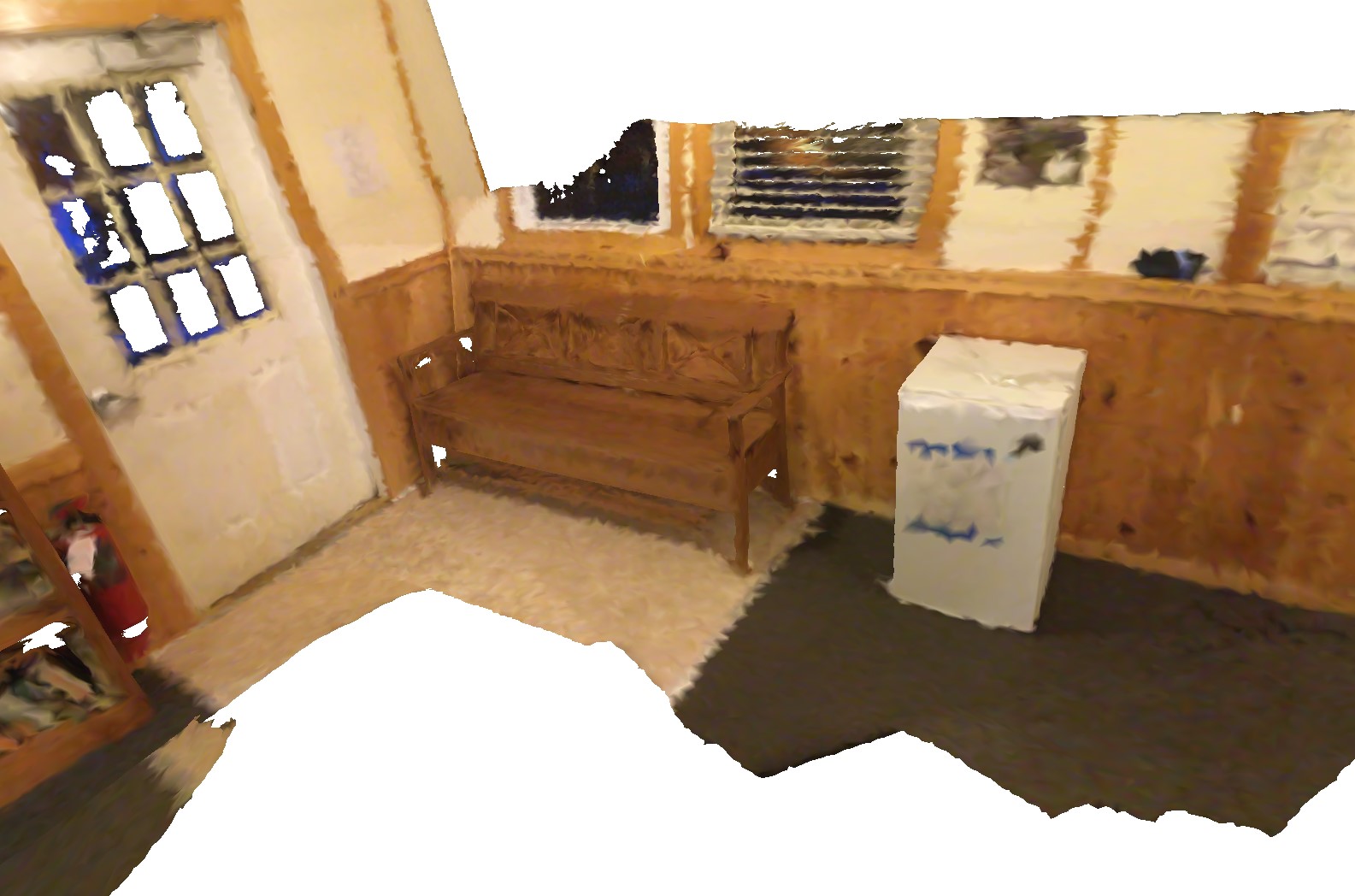}&
\includegraphics[width=.33\linewidth]{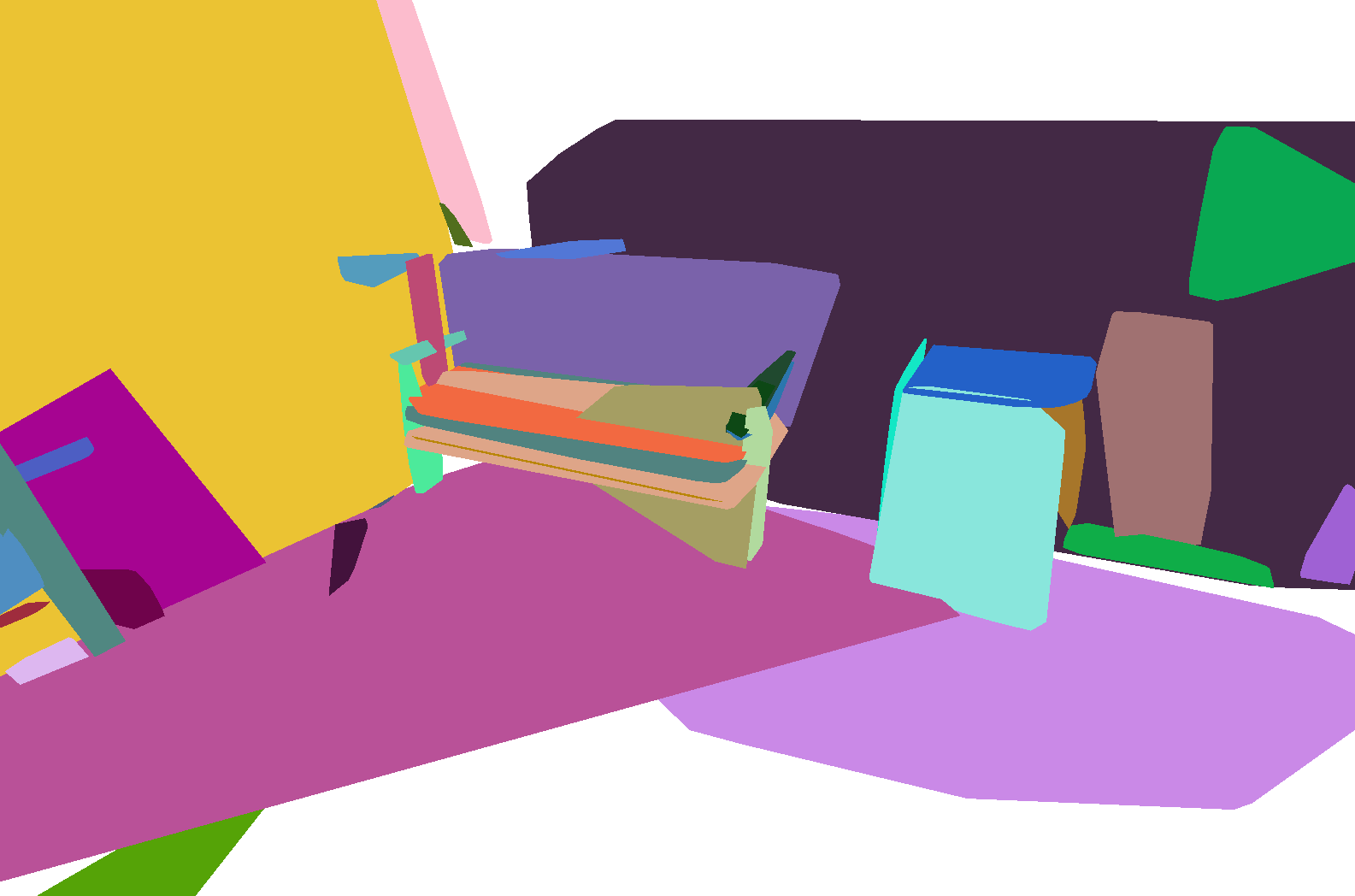}&
\includegraphics[width=.33\linewidth]{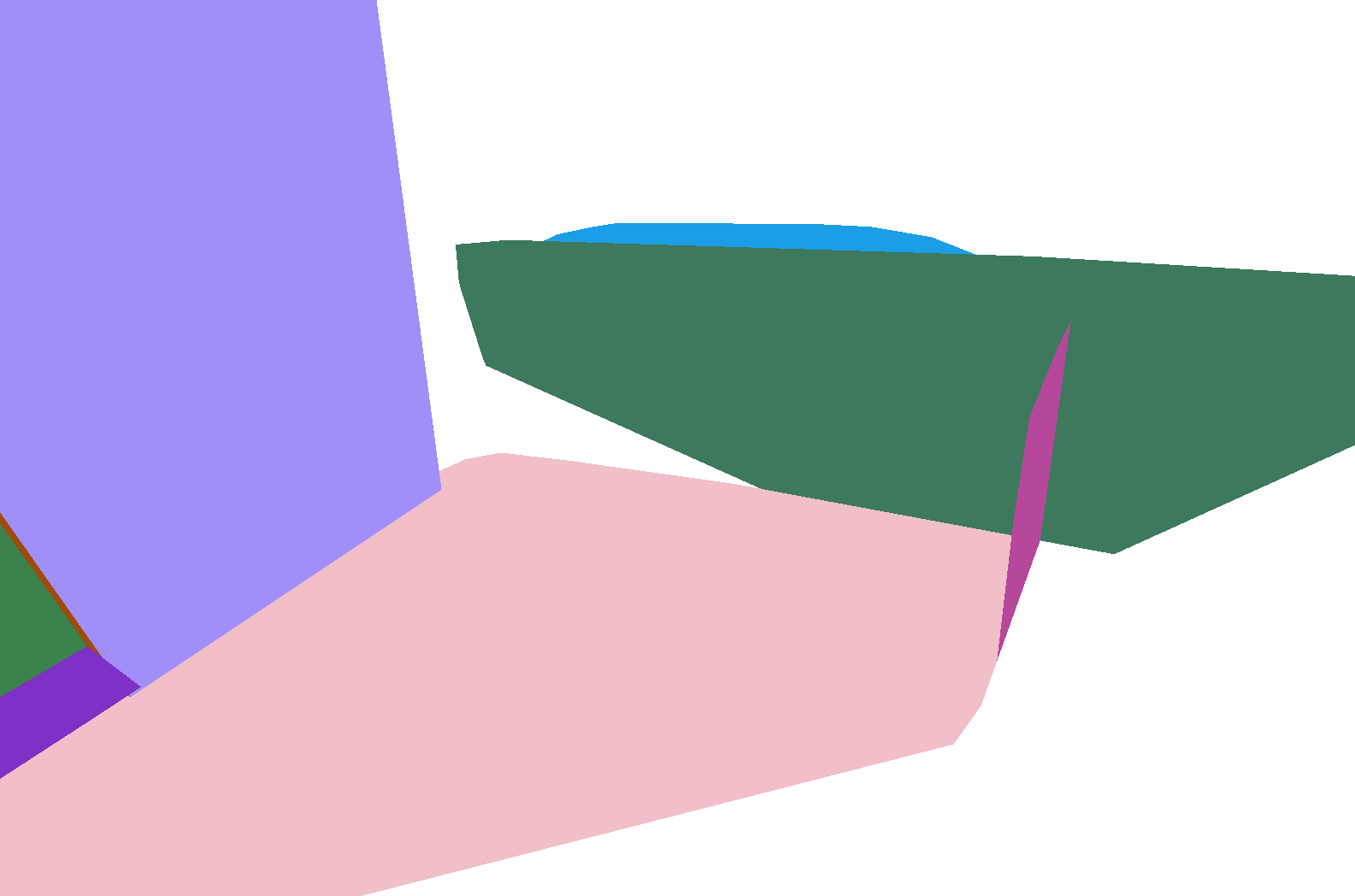}&
\includegraphics[width=.33\linewidth]{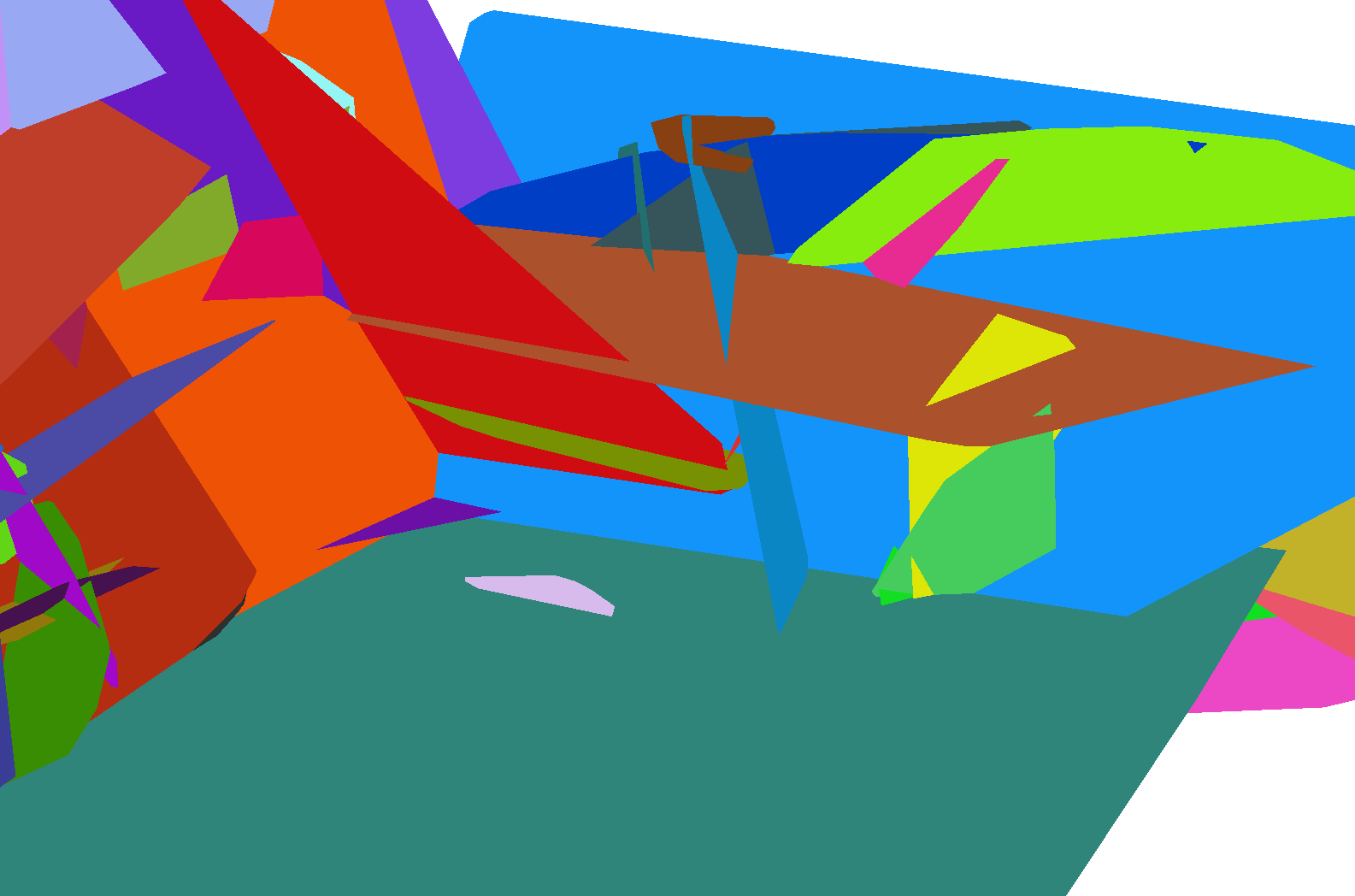}\\
\includegraphics[width=.33\linewidth]{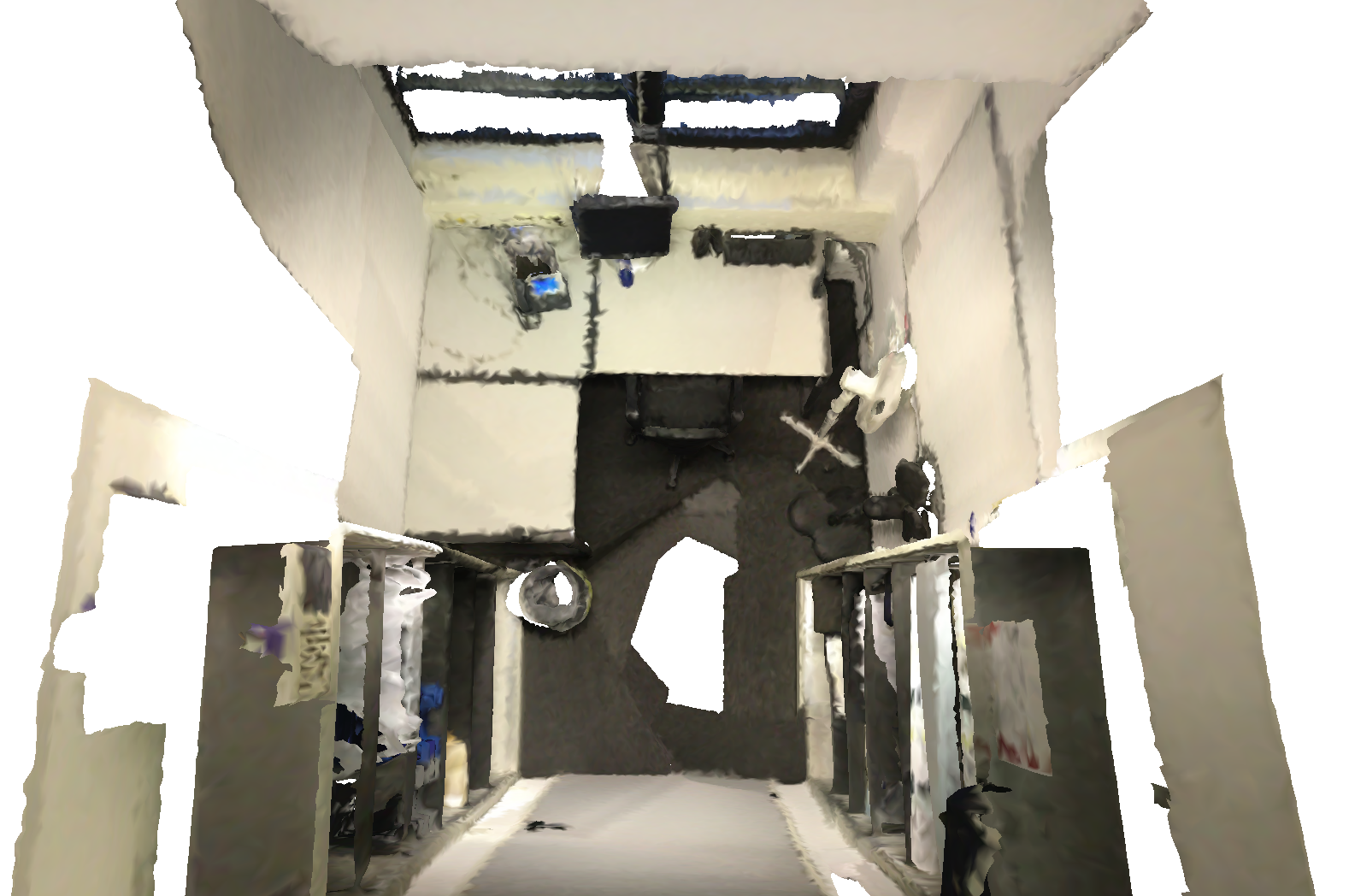}&
\includegraphics[width=.33\linewidth]{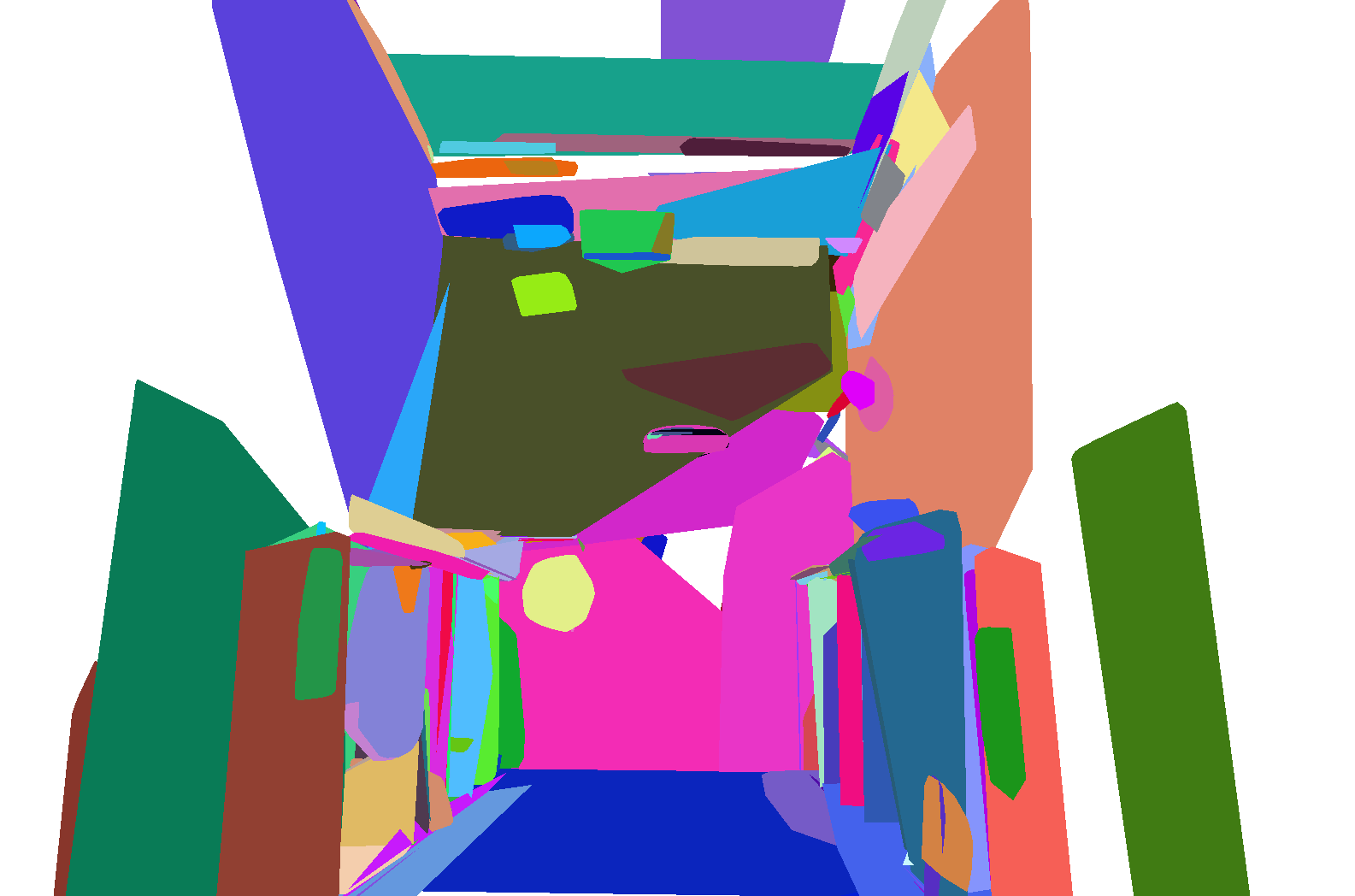}&
\includegraphics[width=.33\linewidth]{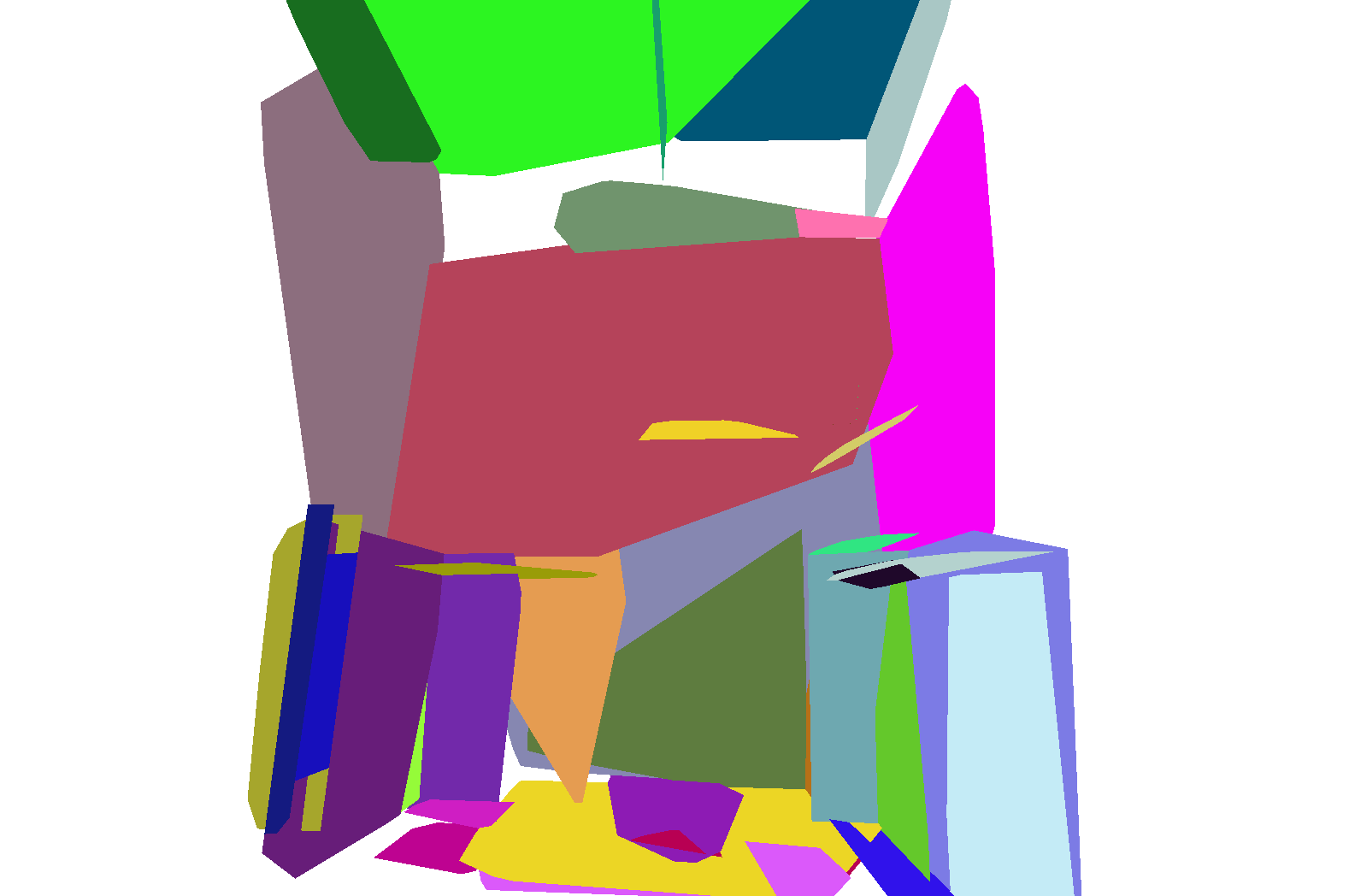}&
\includegraphics[width=.33\linewidth]{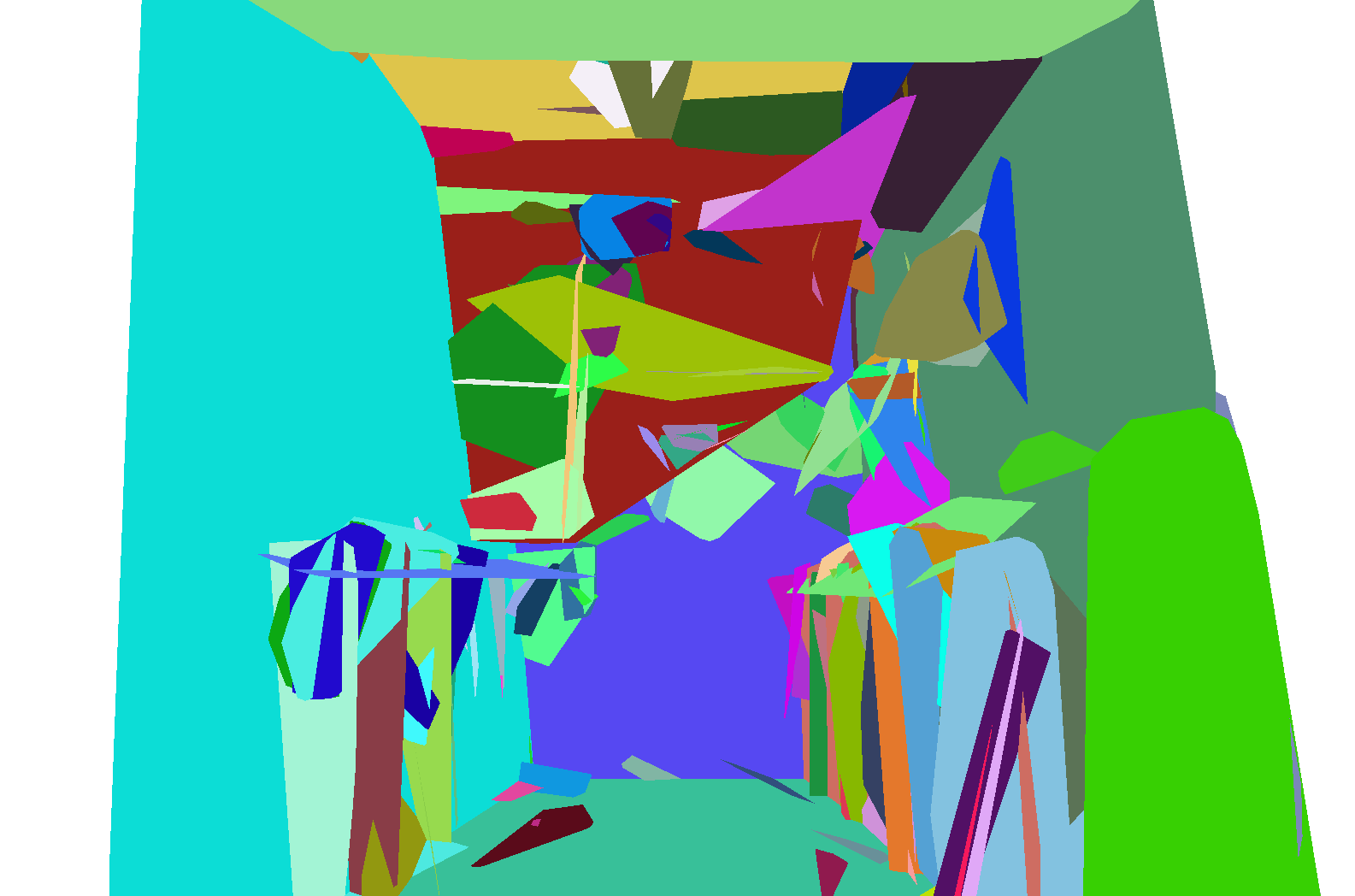}\\
(a) & (b) & (c) & (d) \\
\end{tabular}
\end{adjustbox}
\vspace{-5pt}
\caption{\small Examples of 3D planar reconstruction on ScanNet. (a) Ground-truth textured meshes (with holes on floor due to unseen regions in the videos), (b) our proposed Planar Gaussian Splatting (PGS), (c) PlanarRecon~\cite{xie2022planarrecon}, and (d) 3DGS\texttt{++}. We see that our proposed PGS produces more accurate 3D planes, in terms of precise shapes and recall. PlanarRecon misses many planes and produces incorrect shapes, and 3DGS\texttt{++} results are noisy and inaccurate.}
\label{fig:vis_scannet}
\end{figure*}

\begin{table}[t!]
\center
\caption{3D plane instance segmentation results on ScanNet. The symbol \texttt{++} indicates that the method only produces a 3D mesh reconstruction and a post-processing by Sequential RANSAC is performed to extract the 3D planes.}\label{tab:scannet_seg}
\vspace{-10pt}
\begin{adjustbox}{width=\columnwidth ,keepaspectratio}
\begin{tabular}{l|c|c|c|c|c|c|l}
\toprule
\multirow{2}{*}{Method} & \multirow{2}{*}{VOI $\downarrow$}  & \multirow{2}{*}{RI $\uparrow$} & \multirow{2}{*}{SC$\uparrow$} & \multicolumn{3}{|c|}{\textit{Supervision}} & Inference\\ 
 & & & & RGB & Geo. & Planes &  time\\
\midrule
\makecell{NeuralRecon\texttt{++} ~\cite{sun2021neuralrecon}} & 5.540 &  0.696 &  0.139 & \checkmark & \checkmark & - & 2 min\\
PlanarRecon  ~\cite{xie2022planarrecon} & 3.458 &  0.861 &  0.359  & \checkmark & \checkmark & \checkmark & realtime\\
3DGS\texttt{++}~\cite{kerbl20233d} & 5.056 & 0.850 & 0.306 & \checkmark & - & - & 16 min.\\
NMF~\cite{zanjani2024neural} & 3.253 &  0.880 & 0.381 & \checkmark & - & - & 40 min.\\
Planar GS (ours) & $\mathbf{3.045}$ &  $\mathbf{0.901}$ &  $\mathbf{0.430}$ & \checkmark & - & - & 16 min.\\
\bottomrule
\end{tabular}
\end{adjustbox}
\vspace{-5pt}
\end{table}

\subsection{Main Results}
\vspace{-4pt}
\textbf{On ScanNet}: Table~\ref{tab:scannet_seg} shows 3D planar segmentation results on ScanNet data. We see that our proposed PGS achieves significantly better performance across all the evaluation metrics. For instance, it has significant higher segmentation covering score as compared to existing methods (more than 11\% improvement). Although both NeuralRecon and PlanarRecon are trained on ScanNet, their 3D plane instance segmentation scores are considerably worse. We also see that naively using Sequential RANSAC to extract 3D planes from 3DGS reconstruction does not yield good accuracy. In terms of runtime, our proposed PGS is significantly faster as compared to the latest state-of-the-art optimization-based 3D plane segmentation method of NMF, with more than 60\% less runtime.

Figure~\ref{fig:vis_scannet} shows sample 3D planar reconstruction results on ScanNet. It can be seen that our proposed PGS generates more accurate planes. For instance, in the 2nd row, our PGS captures the shape of the round table top~(b) while PlanarRecon fails to recover the shape~(c). In addition, PGS has higher recalls. We see that PlanarRecon misses a lot of planes in the scene, especially for smaller objects like chairs, while PGS better identifies planes on those objects. 
However, the PGS also makes mistake on small subset of the estimated 3D planes by duplicating plane instances, which can be observed at the intersections of planes in our visualizations.
The results by 3DGS\texttt{++} are very noisy, showing that it is suboptimal to simply add a post-processing step like Sequential RANSAC to extract 3D planes from a reconstruction, as compared to an end-to-end, holistically designed pipeline.

\begin{table}[t!]
\center
\caption{3D plane instance segmentation results on Replica. The symbol \texttt{++} indicates that the method only produces a 3D mesh reconstruction and a post-processing by Sequential RANSAC is performed to extract the 3D planes. Note that ScanNet-trained NeuralRecon fails to produce valid meshes on Replica}\label{tab:replica_seg}
\vspace{-10pt}
\begin{adjustbox}{width=\columnwidth,keepaspectratio}
\begin{tabular}{l|c|c|c|c|c|c}
\toprule
\multirow{2}{*}{Method} & \multirow{2}{*}{VOI $\downarrow$}  & \multirow{2}{*}{RI $\uparrow$} & \multirow{2}{*}{SC$\uparrow$} & \multicolumn{3}{c}{\textit{Supervision}}\\
 & & & & RGB & Geo. & Planes \\
\midrule 
NeuralRecon\texttt{++}~\cite{sun2021neuralrecon} & - &  - &  - & \checkmark & \checkmark & - \\
PlanarRecon\cite{xie2022planarrecon} & 4.676 &  0.829 &  0.148  & \checkmark & \checkmark & \checkmark\\
3DGS\texttt{++}~\cite{kerbl20233d} & 4.401 &  0.904 &  0.179 & \checkmark & - & -\\
NMF~\cite{zanjani2024neural} & 4.311 &  0.891 &  0.188 & \checkmark & - & -\\
Planar GS (ours) & $\mathbf{4.168}$ & $\mathbf{0.943}$ & $\mathbf{0.209}$ & \checkmark & - & -\\
\bottomrule
\end{tabular}
\end{adjustbox}
\vspace{-0pt}
\end{table}

\textbf{On Replica}: Table~\ref{tab:replica_seg} shows the evaluation results on Replica. Both supervised methods of NeuralRecon and PlanarRecon trained on ScanNet cannot generalize to the new dataset, with NeuralRecon failing to produce valid meshes and PlanarRecon generating poor planar reconstruction results. On the other hand, our proposed PGS works well on Replica, with higher 3D plane segmentation scores and a lower inference time when comparing to the existing SOTA optimization-based method of NMF. 

\vspace{-2pt}
\subsection{Ablation Study}
\vspace{-4pt}
We analyze different aspects of our proposed design. The ablation experiments are performed on two ScanNet scenes. More specifically, we study the effectiveness of 
(1)~utilizing SAM masks for learning plane descriptors, (2)~2D normal maps supervision, (3)~local planar alignment (Section~\ref{sec:lpa}), (4)~applying holistic separability by using the recurrent mean-shift layer (Section~\ref{sec:meanshift}), and (5)~Laplacian smoothing of geometric features including normal and plane descriptor smoothing (Section~\ref{sec:lpa}). 
Table~\ref{tab:ablation} shows the 3D plane segmentation performance on different variants of the proposed PGS, where we deactivate one component at each time. 
We observe a significant drop in performance when SAM masks are not used. This is expected, as the PGS learns plane descriptors using SAM, which are later used for grouping Gaussian nodes in tree and parsing plane instances. The absence of plane descriptors leads to high ambiguity in parsing individual small surfaces close to larger planar regions.
Moreover, dropping normal vectors results in less
degradation, as the plane descriptors using SAM masks can mainly resolve grouping ambiguity between Gaussian nodes in the tree structure.
Furthermore, we see that the local planar alignment shows a high impact on the performance. This is because by enforcing the Gaussians to locate on local tangent planes of surfaces, it improves the learning of correct geometric features, such as normal vectors and plane descriptors, through rendering. While dropping such alignment results in a 3D reconstruction which point cloud  is scattered around the surfaces.
The Laplacian smoothing includes local averaging over both the normal and descriptor features of Gaussian primitives in training time, which further improves the performance. It can be seen that encouraging holistic separability also helps, providing effective performance improvement on plane segmentation.

\begin{table}[t!] 
\center
\caption{\small Ablation Study on two ScanNet scenes.}\label{tab:ablation}
\vspace{-10pt}
\begin{adjustbox}{width=.9\columnwidth,keepaspectratio}
\begin{tabular}{l|c|c|c}
\toprule
Experiment & VOI$\downarrow$ & RI $\uparrow$ & SC$\uparrow$\\ 
\midrule 
W/o SAM masks  & \quad 3.914 \quad& \quad 0.873 \quad& \quad 0.349 \\
W/o local planar alignment  & \quad3.655 \quad& \quad0.901 \quad& \quad0.374 \\
W/o normal vectors & \quad 3.326 \quad& \quad 0.904 \quad& \quad 0.390 \\
W/o holistic separability & \quad3.240 \quad& \quad0.905 \quad& \quad0.401 \\
W/o Laplacian smoothing & \quad3.151 \quad& \quad 0.908 \quad & \quad 0.393 \\
Full Planar GS & \quad \textbf{3.024} \quad & \quad \textbf{0.919} \quad & \quad \textbf{0.415} \\
\bottomrule
\end{tabular}
\end{adjustbox}
\vspace{-8pt}
\end{table}  

\vspace{-2pt}
\section{Conclusions}
\vspace{-4pt}

In this paper, we proposed Planar Gaussian Splatting (PGS), which leverages two fundamental concepts: a probabilistic, hierarchical Gaussian mixture approach and a foundational vision model. Our approach represents the scene using a set of 3D planes, each defined by merging local geometries represented through 3D Gaussian distributions. To address ambiguity during the merging process, we introduce additional parameters to the Gaussians, including the normal vector and a plane descriptor. 
Learning plane descriptors without access to 2D/3D plane annotations involves utilizing a vision foundation model, specifically SAM. We learn 3D plane descriptors by constructing and partitioning a region adjacency graph based on SAM segments. Additionally, we address the challenges posed by variable-length and non-corresponding mask proposals across images via a linear regression approach. Experiments demonstrate that the proposed PGS outperforms existing competitive approaches in 3D planar reconstruction.

\textbf{Limitations}: The proposed method has some limitations. Dark regions in the image usually suffers under-reconstruction due to sparse assignment of Gaussian points. This can affect computing both KNN and the statistics such as mean and covariance of the point distributions on the plane. 
Additionally, very large planes in the scene might be split into two or several pieces as the likelihood of learning a compact descriptor for a large area containing a huge number of Gaussians primitives decreases. The proposed recurrent mean-shift updates mitigate this issue to some extent but does not resolve it completely.

{\small
\bibliographystyle{ieee_fullname}
\bibliography{main}
}

\newpage
\appendix
\onecolumn
\begin{center}
\LARGE{Planar Gaussian Splatting \\ --- Supplementary Material --- }
\linebreak
\linebreak
\\
\author{\large{Farhad G. Zanjani \qquad Hong Cai \qquad Hanno Ackermann \qquad Leila Mirvakhabova \qquad Fatih Porikli} \linebreak
\\
\large{Qualcomm AI Research\footnote{Qualcomm AI Research is an initiative of Qualcomm Technologies, Inc.}}\\
{\tt\small \{fzanjani, hongcai, hackerma, lmirvakh, fporikli\}@qti.qualcomm.com}
}
\linebreak
\end{center}

This supplementary document includes further implementation details of the proposed Planar Gaussian Splatting method, which is discussed in Section A. Additionally, we provide an evaluation of geometry reconstruction and present more quantitative results in Section B.

\section{Implementation Details}

We initialize the Gaussian locations and color with the sparse point cloud obtained by running SfM on training images. We train the model on a single NVIDIA GeForce RTX 2080 Ti for 15K iterations. We follow almost all the hyperparameters introduced in~\cite{kerbl20233d}, except that we increase the densification threshold on the gradient to 0.001 in order to limit the number of cloning/splitting Gaussians and using only two degrees of spherical harmonics because a high-quality rendering is not the primary objective of this work. 
Cutting edges of RAG is performed by setting a threshold of 10~cm on planar distances and 20 degrees for the surface normal cosine distance. The recursive mean-shift is run once after 2k iterations of training and repeated every 100 training iterations, and the number of update steps (Eq.~\ref{eq:mean-shift}) performed each time is set to 10 with $\gamma$ equal to 60. The number of neighbours ($K$) is equal to 30 and the KNN is recomputed after applying any refinement to the 3D Gaussian field, including cloning, splitting, and culling of the Gaussians. Local planar alignment is applied every 500 iterations.

\subsection{Efficient implementation of holistic separability}
Applying recurrent mean-shift to the descriptor vectors of all Gaussian points in the scene enhances the separability among descriptors corresponding to distinct planes. The mean-shift update computation involves recursively evaluating Eq.~(\ref{eq:mean-shift}) for a specified number of steps: 
\begin{align*}
    \mathbf{Z} \leftarrow \mathbf{Z}\cdot(\eta \cdot \mathbf{K} \cdot \mathbf{D}^{-1} + (1-\eta)\cdot \mathbf{I}),\\[-15pt]
\end{align*}
where $\mathbf{Z} \in \mathbb{R}^{N\times d}$ denotes the matrix of descriptors of length $d$ (e.g., $d=3$ in our experiments) for the entire scene with $N$ number of Gaussian points (typically in the order of millions). 
To compute pairwise distances using the von Mises-Fisher kernel, we encounter a challenge: the computation of $\mathbf{K} \in \mathbb{R}^{N \times N}$ is quadratic with respect to the number of Gaussian points in the scene. This becomes infeasible in terms of computational memory.

In our implementation, we tackle this issue by initially estimating the kernel on a random subset of samples with $M$ points (where $M\ll N$). As a result, the estimated kernel $K \in \mathbb{R}^{M \times M}$ and the update vectors (right-hand side of the equation) are propagated and shared across all KNN samples within the chosen set. The random sampling process (without replacement) continues until an update vector has been computed for all samples or their neighbors. Estimating the mean-shift updates for a large number of Gaussian points in the scene using this sampling strategy takes approximately one second on the GPU, and the required memory can be accommodated within the available resources. Notably, in our experiments, holistic separability occurs every $N=$100 iterations, introducing only minor overhead to the training time.

\begin{figure}[H]
\centering
\begin{adjustbox}{width=.95\textwidth}
\begin{tabular}{ccc}
\includegraphics[width=.33\linewidth]{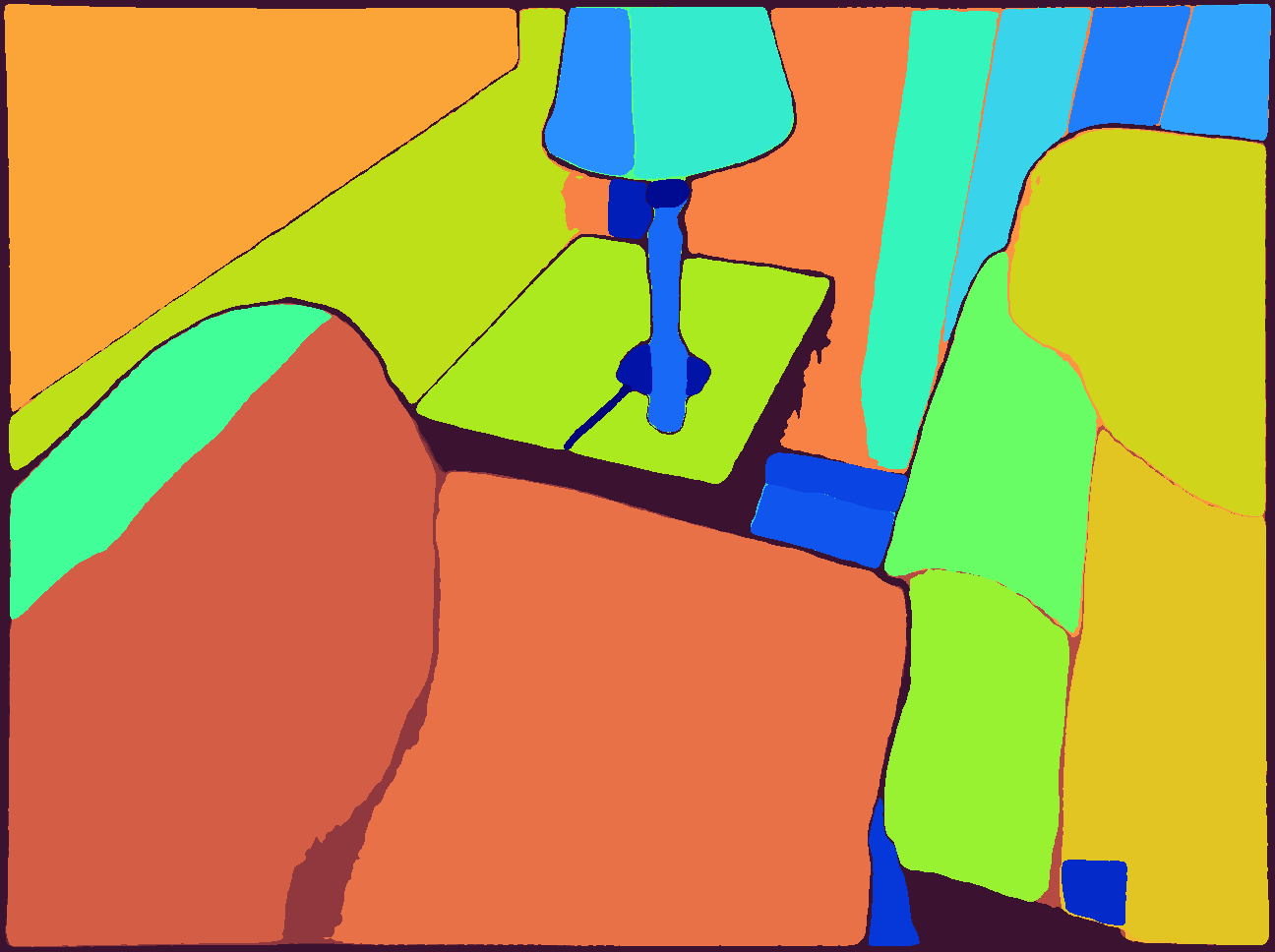}&
\includegraphics[width=.33\linewidth]{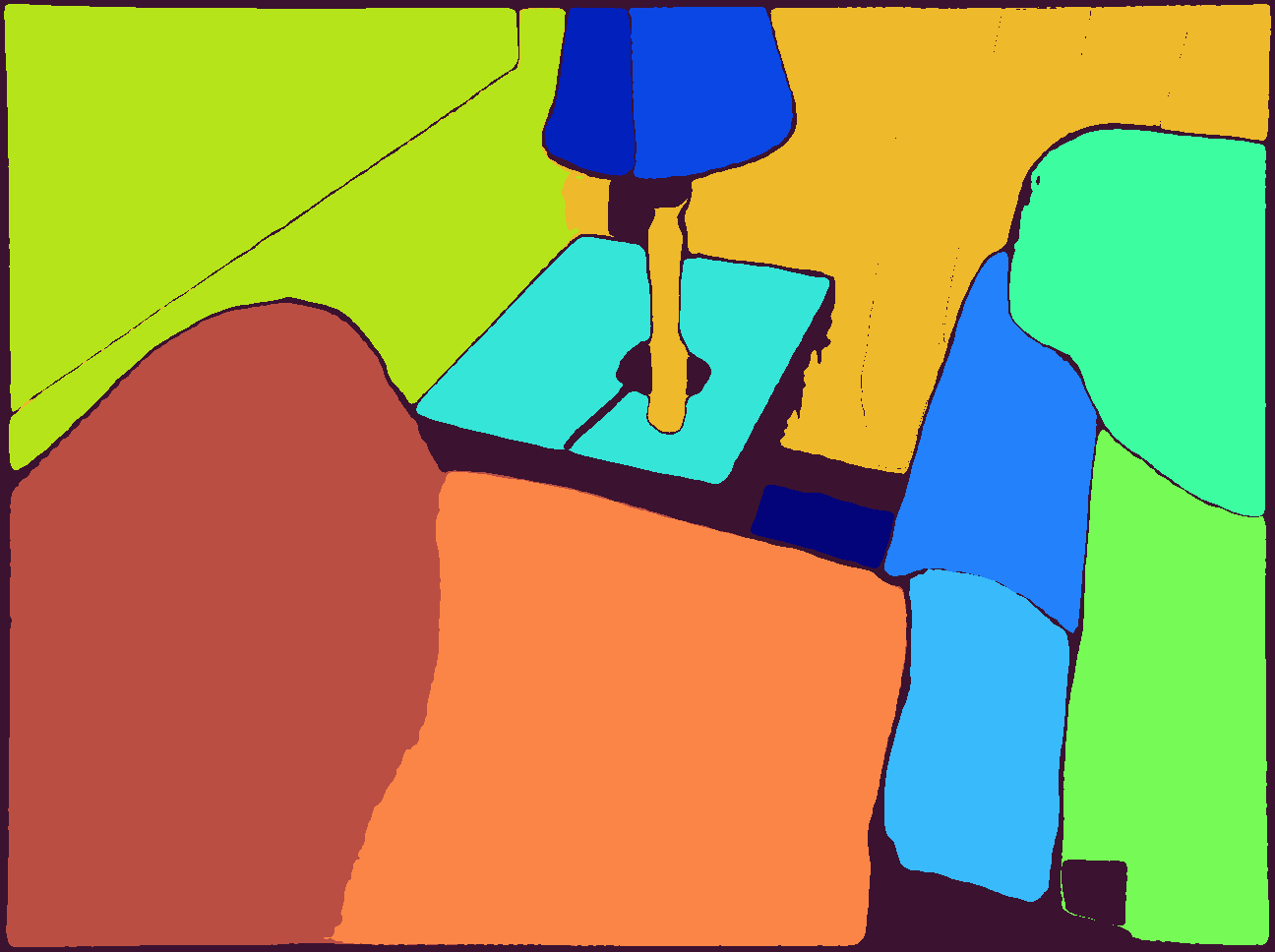}&
\includegraphics[width=.33\linewidth]{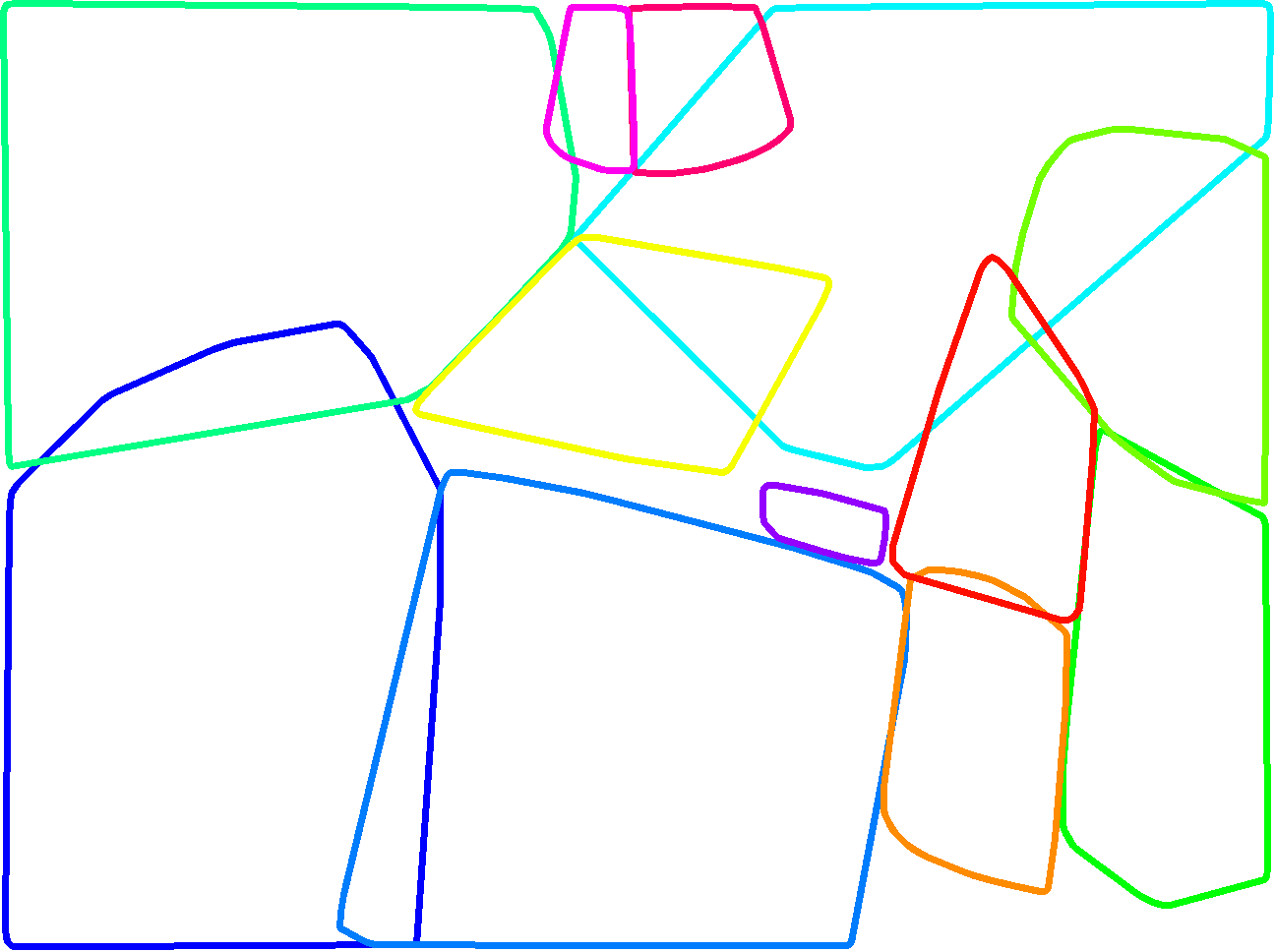}\\
(a) & (b) & (c)\\
\end{tabular}
\end{adjustbox}
\vspace{-5pt}
\caption{\small Example of merging SAM proposals and lifting their boundary points into 3D. The boundary points of each segments is used for estimating a Gaussian distribution in 3D as a leaf node of GMT.}
\label{fig:grouping}
\vspace{-5pt}
\end{figure}

\subsection{Local grouping of Gaussians primitives}
In Section~\ref{sec:GMT}, we address the construction of the Gaussian Mixture Tree (GMT). Rather than directly utilizing the Gaussian points at the leaf nodes — a computationally expensive approach, especially when dealing with a large number of points in a 3D scene — we adopt an alternative strategy. The GMT is formed by computing a set of Gaussian distributions over point clusters derived from all training images. Each cluster corresponds to points belonging to SAM proposal masks after refinement (achieved through merging via RAG).

Figure~\ref{fig:grouping} illustrates this process: (a) SAM Proposals represent distinct regions; (b) the refined version merges the proposals on the same planar surface; (c) boundary points of each mask proposal are lifted into 3D space using the rendered depth map and camera parameters. For each set of boundary 3D points, the mean and covariance matrix of a Gaussian distribution is computed. By estimating the Gaussian distributions from all training images collectively, the leaf nodes of the Gaussian Mixture Tree are determined. These nodes are subsequently used to construct the tree, as explained in Section~\ref{sec:GMT}.

\section{Additional Evaluations}
This section provides more quantitative and qualitative evaluation of the Planar Gaussian Splatting (PGS). 

\begin{table}[H] 
\center
\caption{\small Performance of geometric reconstruction on ScanNet.}\label{tab:geometry}
\begin{adjustbox}{width=.4\textwidth,keepaspectratio}
\begin{tabular}{l|c|c}
\toprule
Method & Accuracy$\downarrow$ & Completeness $\downarrow$\\ 
\midrule
PlanarRecon~\cite{xie2022planarrecon}  & \quad 0.154 \quad& \quad 0.187 \\
PGS (ours)  & \quad 0.137 \quad& \quad0.118 \\
\bottomrule
\end{tabular}
\end{adjustbox}
\vspace{-5pt}
\end{table}  

\subsection{Geometric reconstruction}
In addition to evaluating 3D plane instance segmentation, we assess the performance of both PlanarRecon and the proposed Planar Gaussian Splatting on geometric reconstruction. This evaluation involves measuring the distances between the ground-truth 3D meshes and the surfaces of 3D planes.  
We adhere to common practice and report reconstruction quality using Accuracy and Completeness metrics (see~\cite{murez2020atlas} for their mathematical definitions). The accuracy quantifies the mean distance of the reconstruction points from the ground truth.
Completeness measures the extent to which the ground truth is recovered and is defined as the mean distance of the ground truth points to the reconstruction points.
In Table~\ref{tab:geometry}, we present the measured distances for the ScanNet scenes. Notably, the predicted planes by the proposed PGS exhibit lower distances compared to the ground truth mesh.

\subsection{Visualization of plane descriptors of Gaussian field}
Figure~\ref{fig:vis_scannet_desc} shows samples scenes of ScanNet and Replica, and the learnt plane descriptors. The visualized points are the centers of Gaussians where are colorized by the learnt descriptors. The distinct colors assigned to different 3D plane instances illustrate the effectiveness of our learnt descriptors in 3D Gaussian field.  

\begin{longtable}{cc}
\includegraphics[width=.45\linewidth]{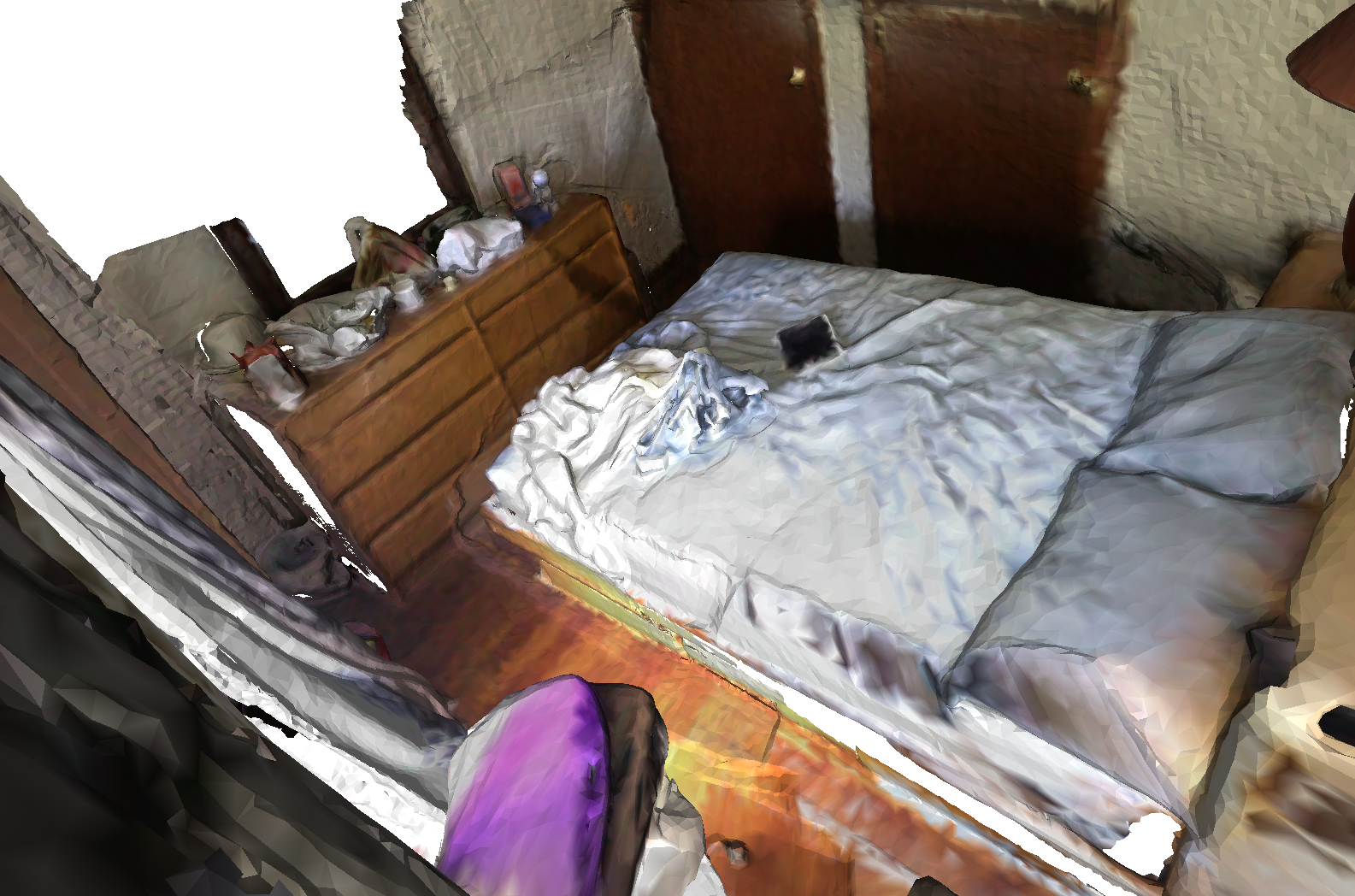}&
\includegraphics[width=.45\linewidth]{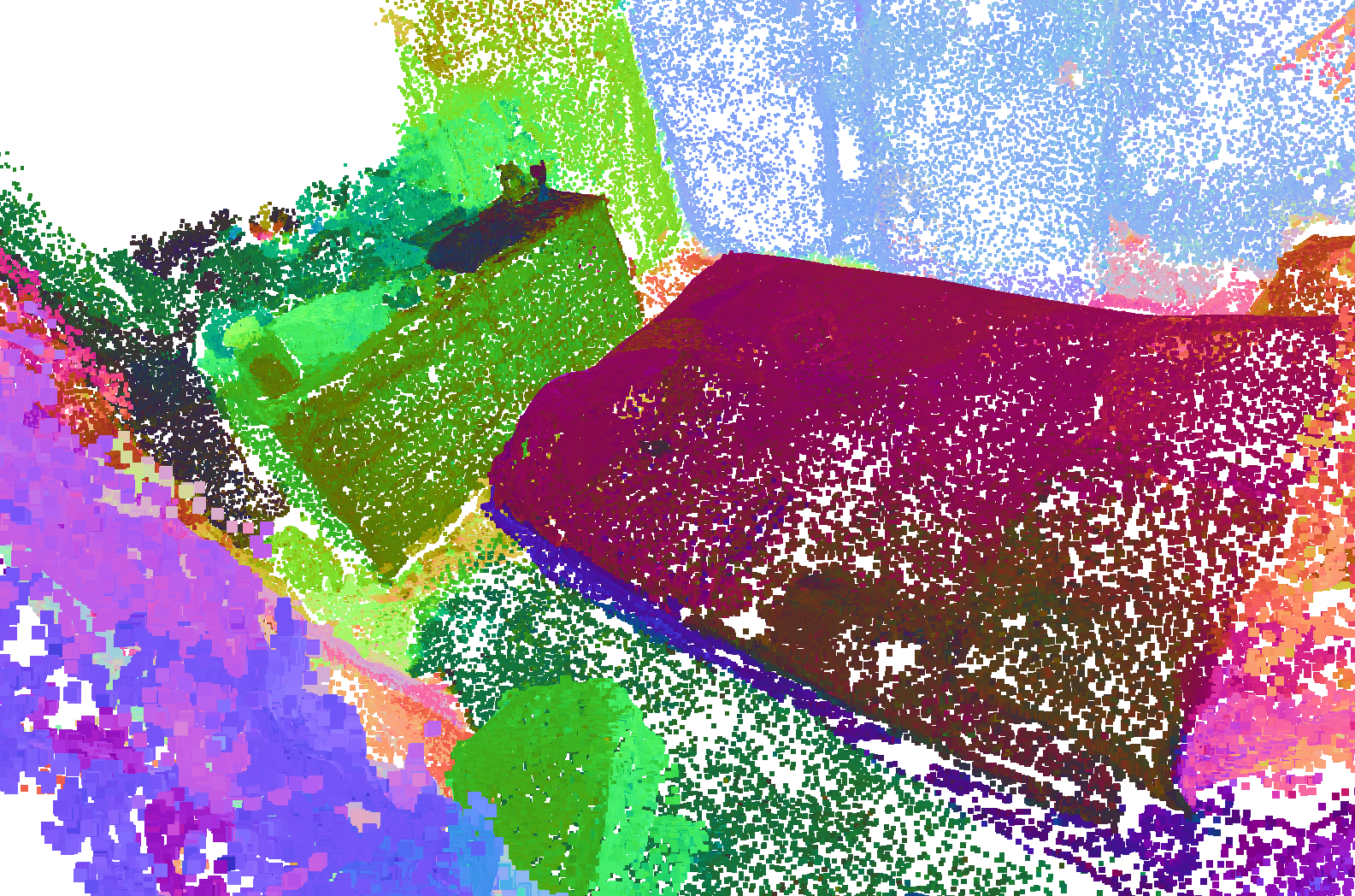}\\
\includegraphics[width=.45\linewidth]{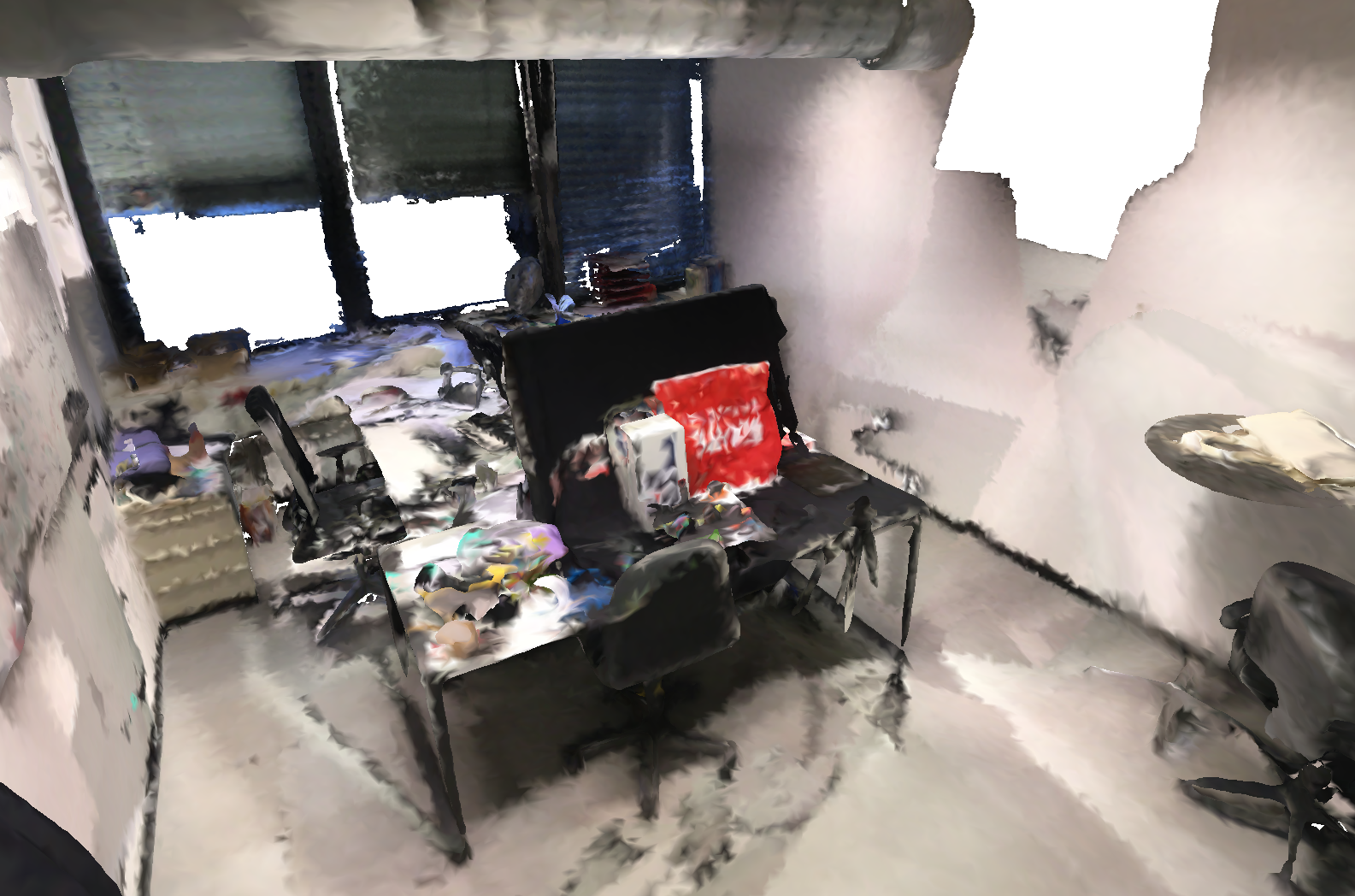}&
\includegraphics[width=.45\linewidth]{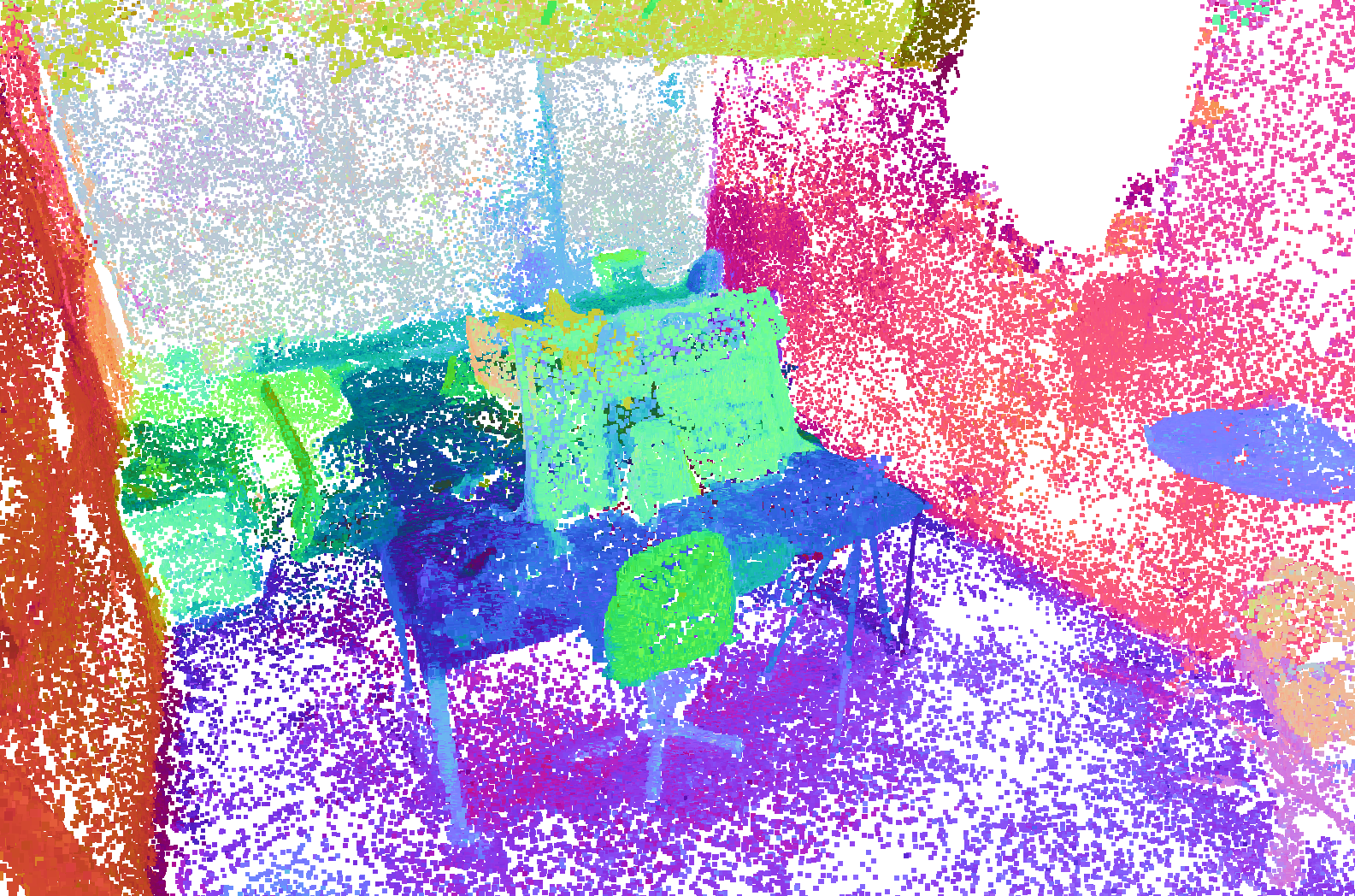}\\
\includegraphics[width=.45\linewidth]{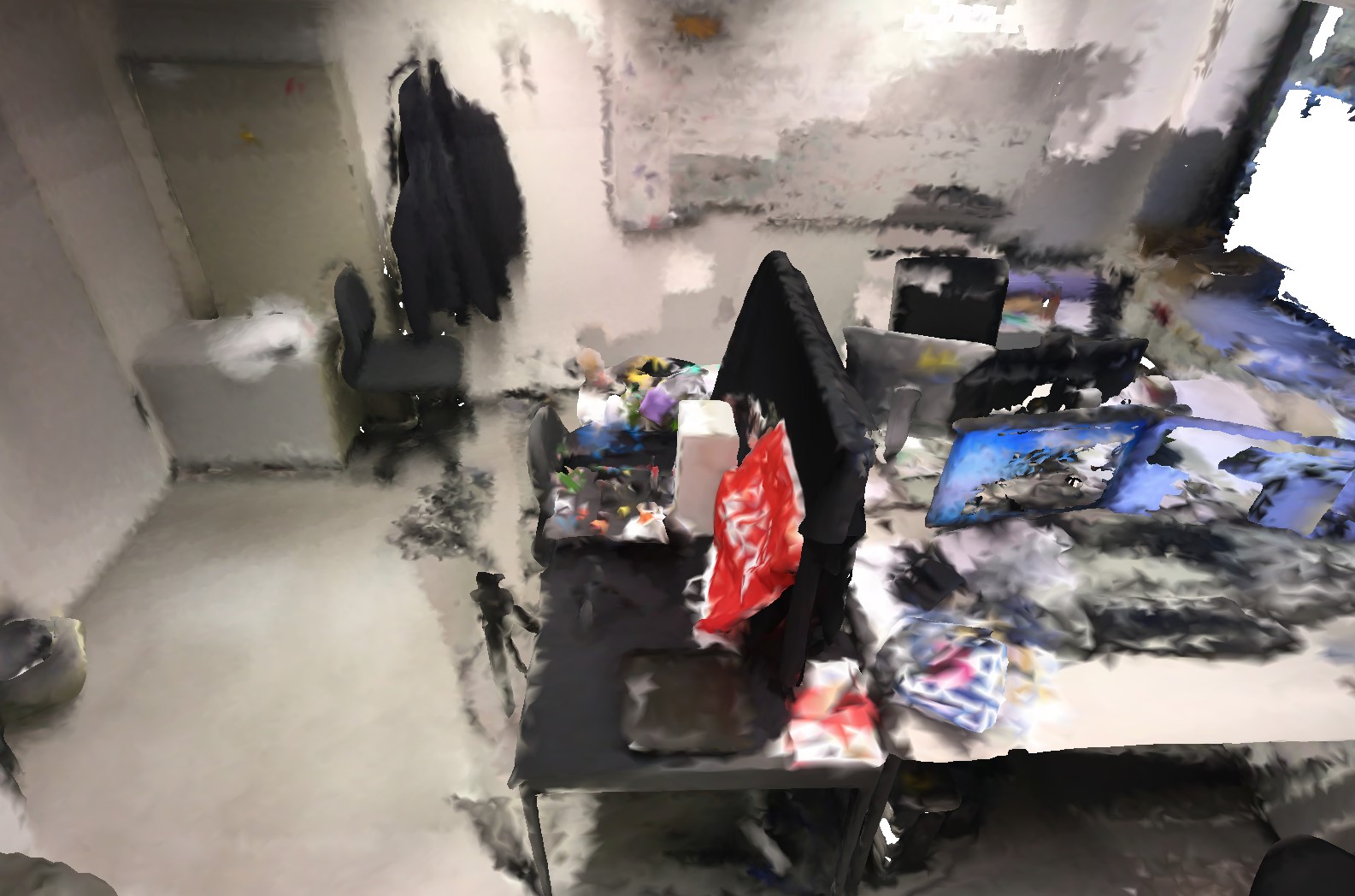}&
\includegraphics[width=.45\linewidth]{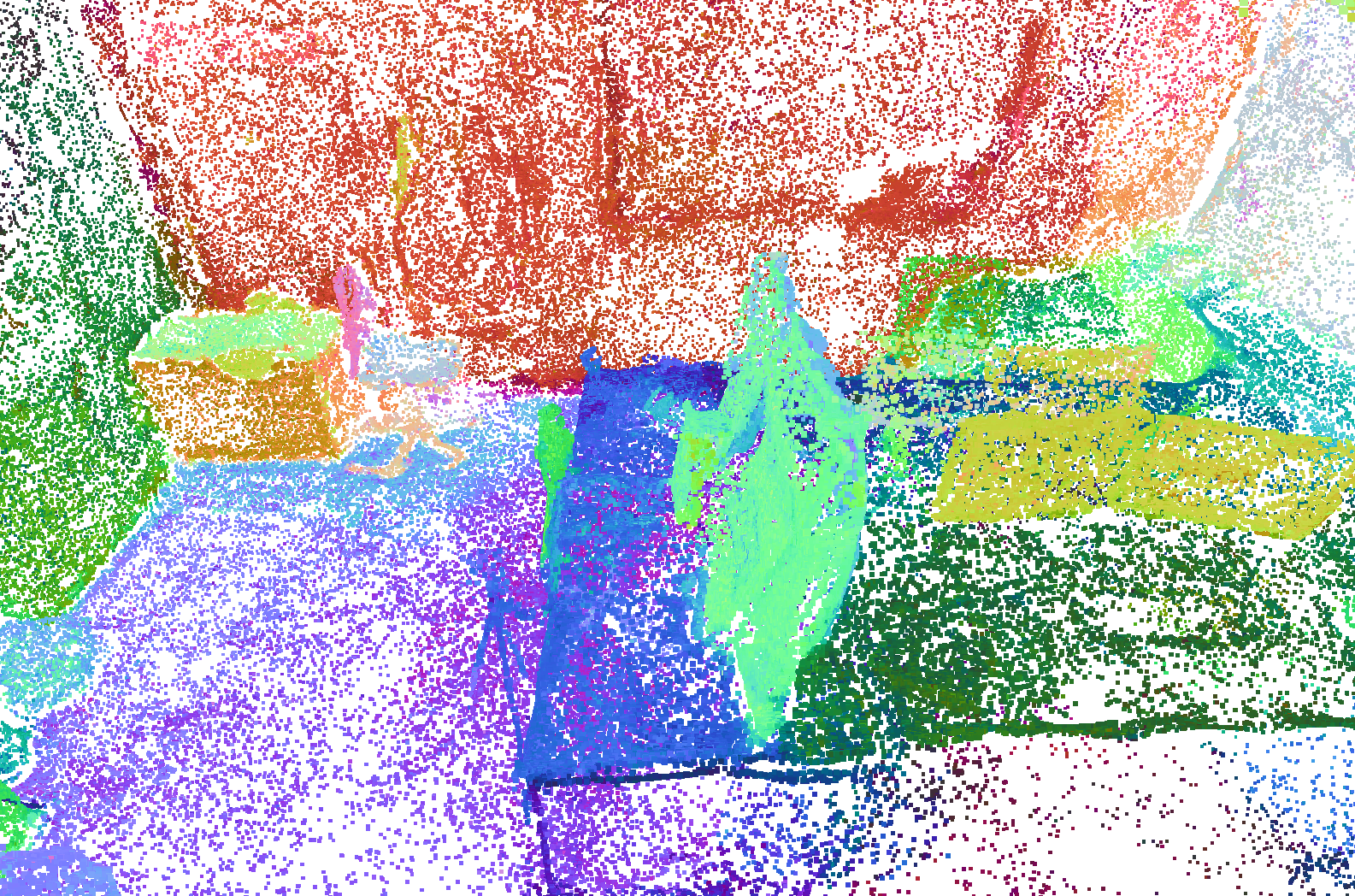}\\
\includegraphics[width=.45\linewidth]{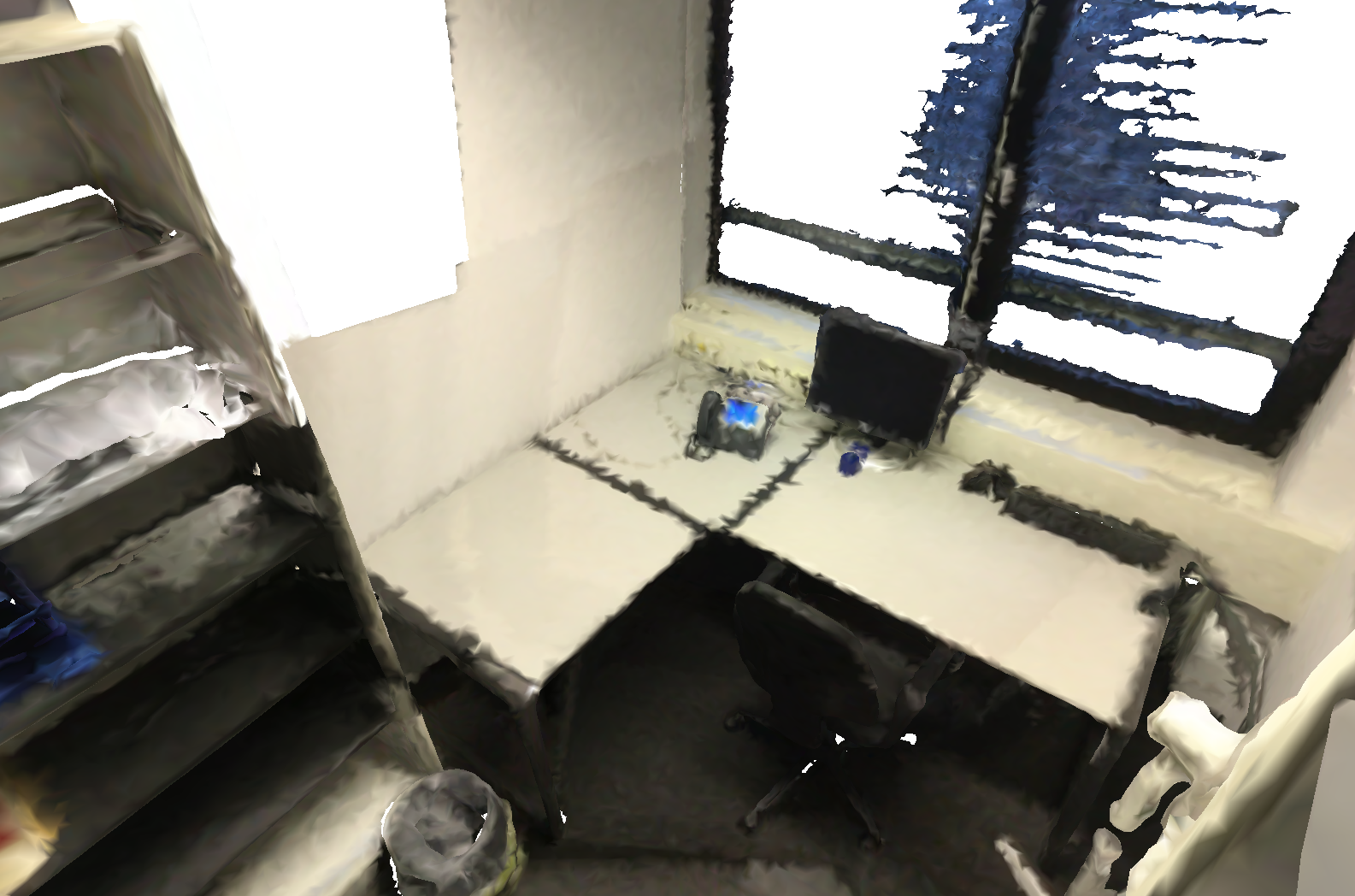}&
\includegraphics[width=.45\linewidth]{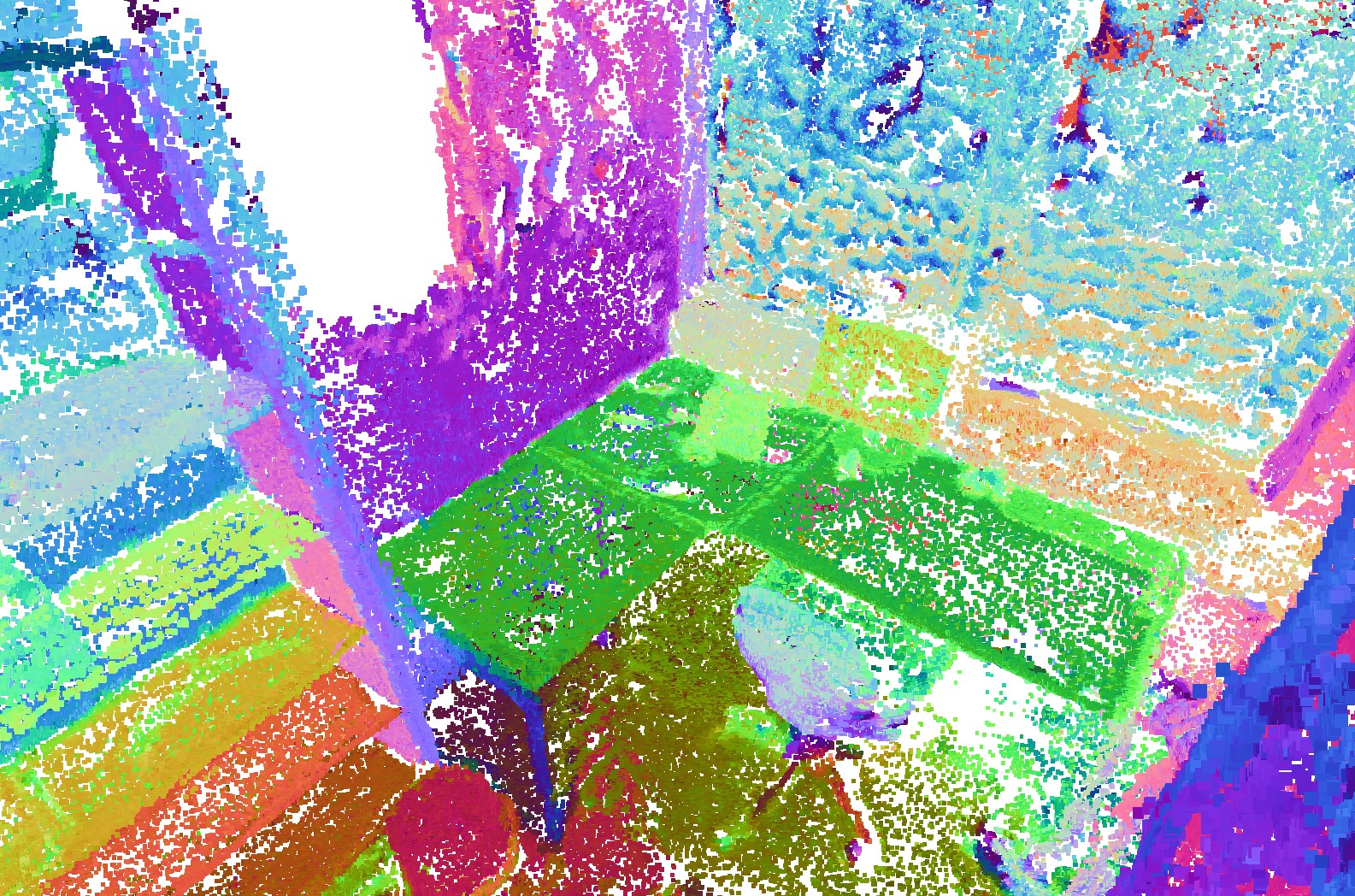}\\
\includegraphics[width=.45\linewidth]{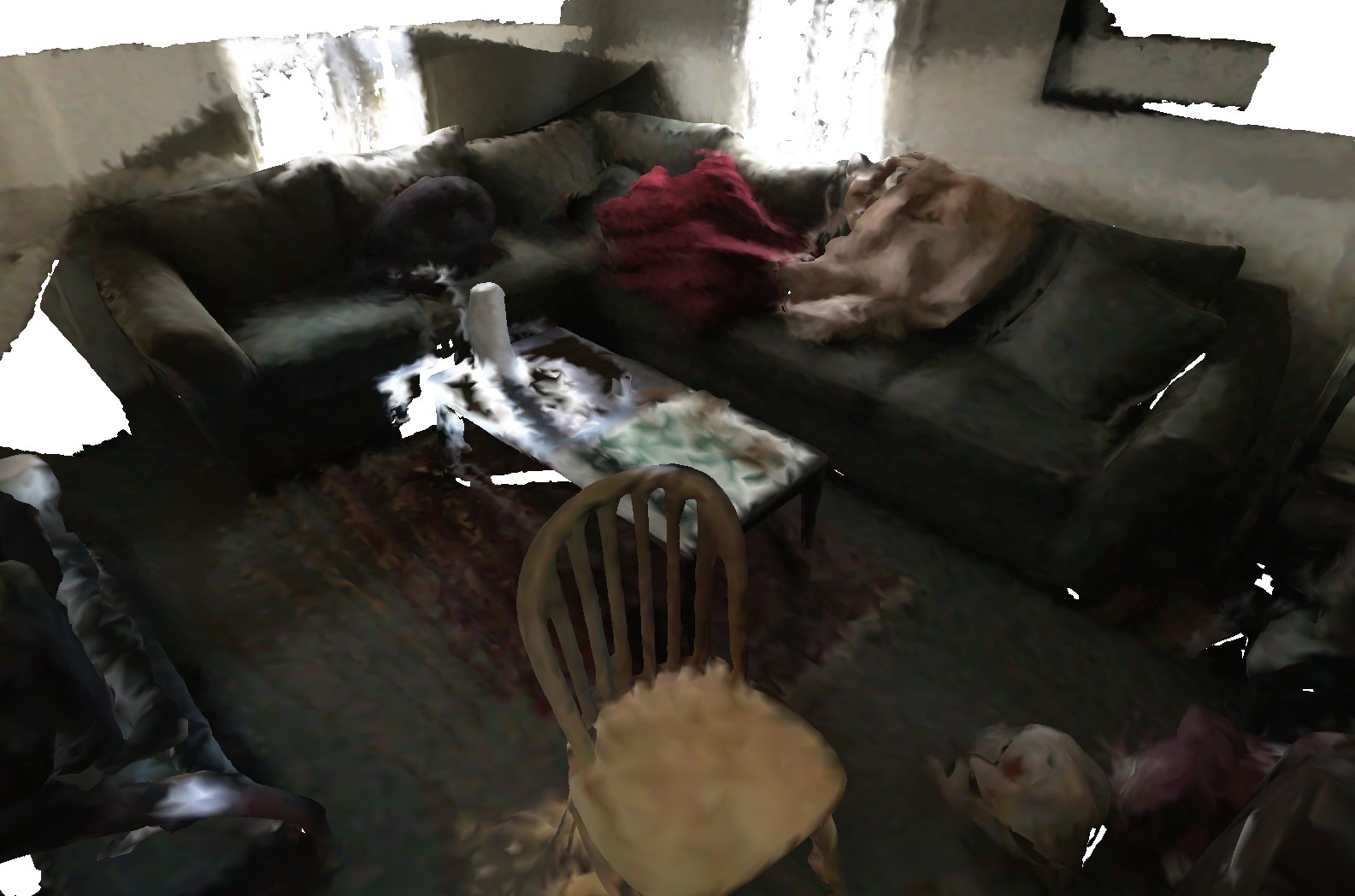}&
\includegraphics[width=.45\linewidth]{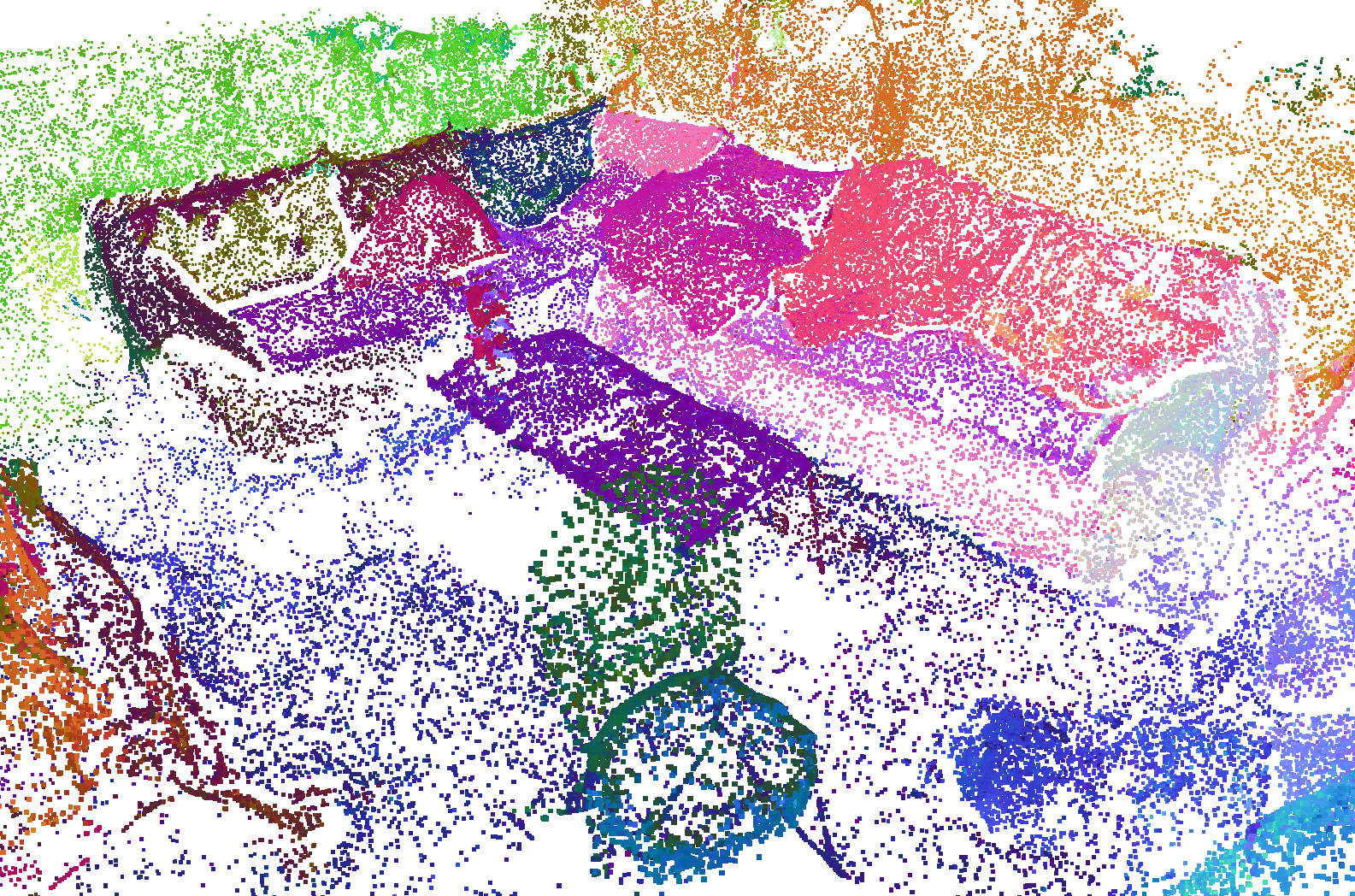}\\
\includegraphics[width=.45\linewidth]{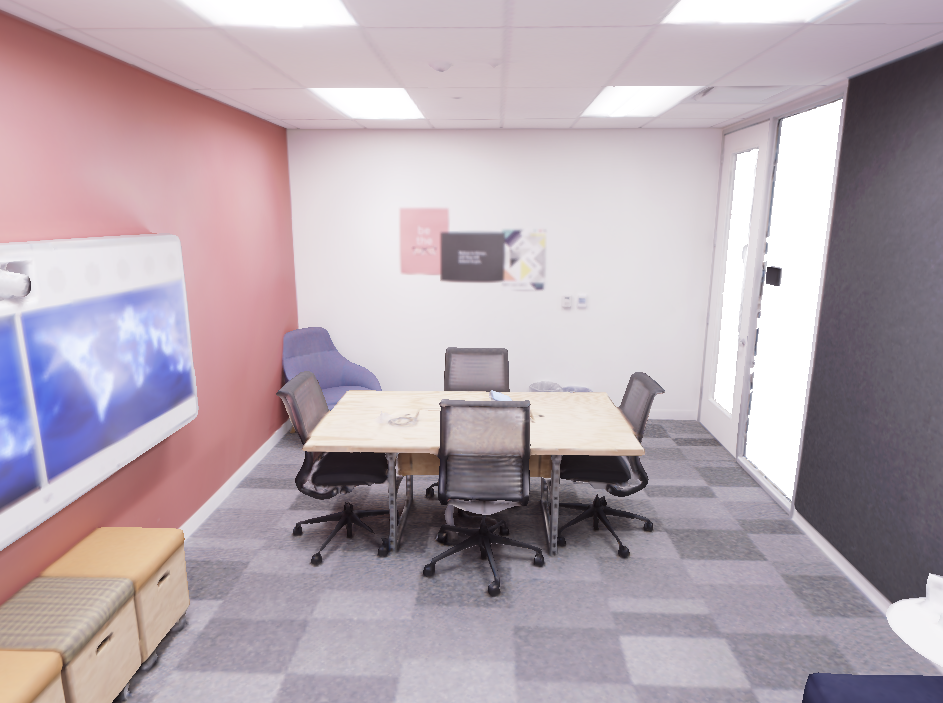}&
\includegraphics[width=.45\linewidth]{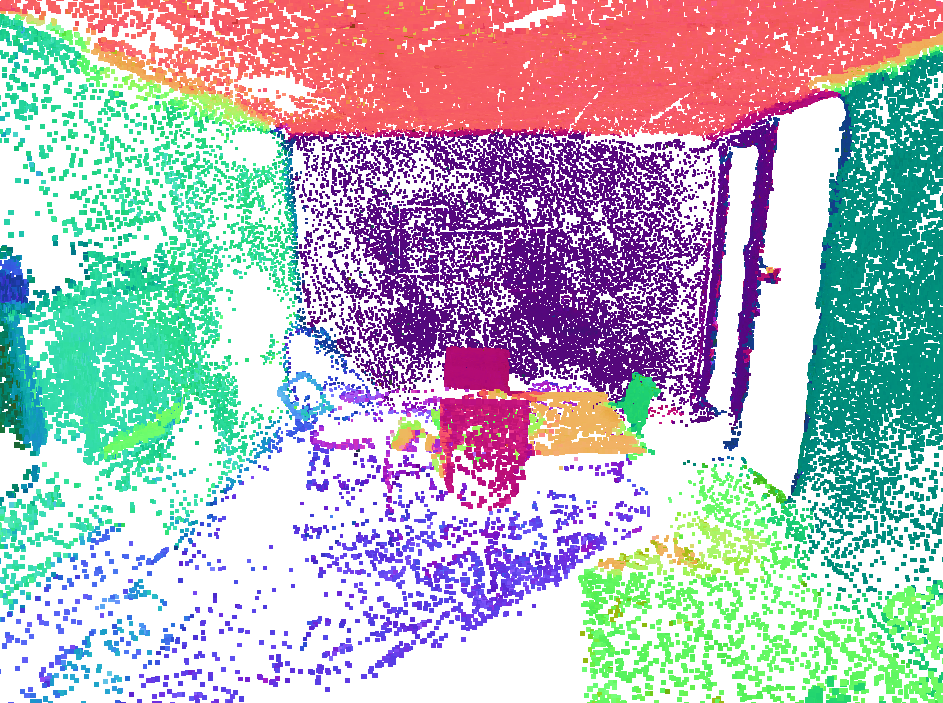}\\
\includegraphics[width=.45\linewidth]{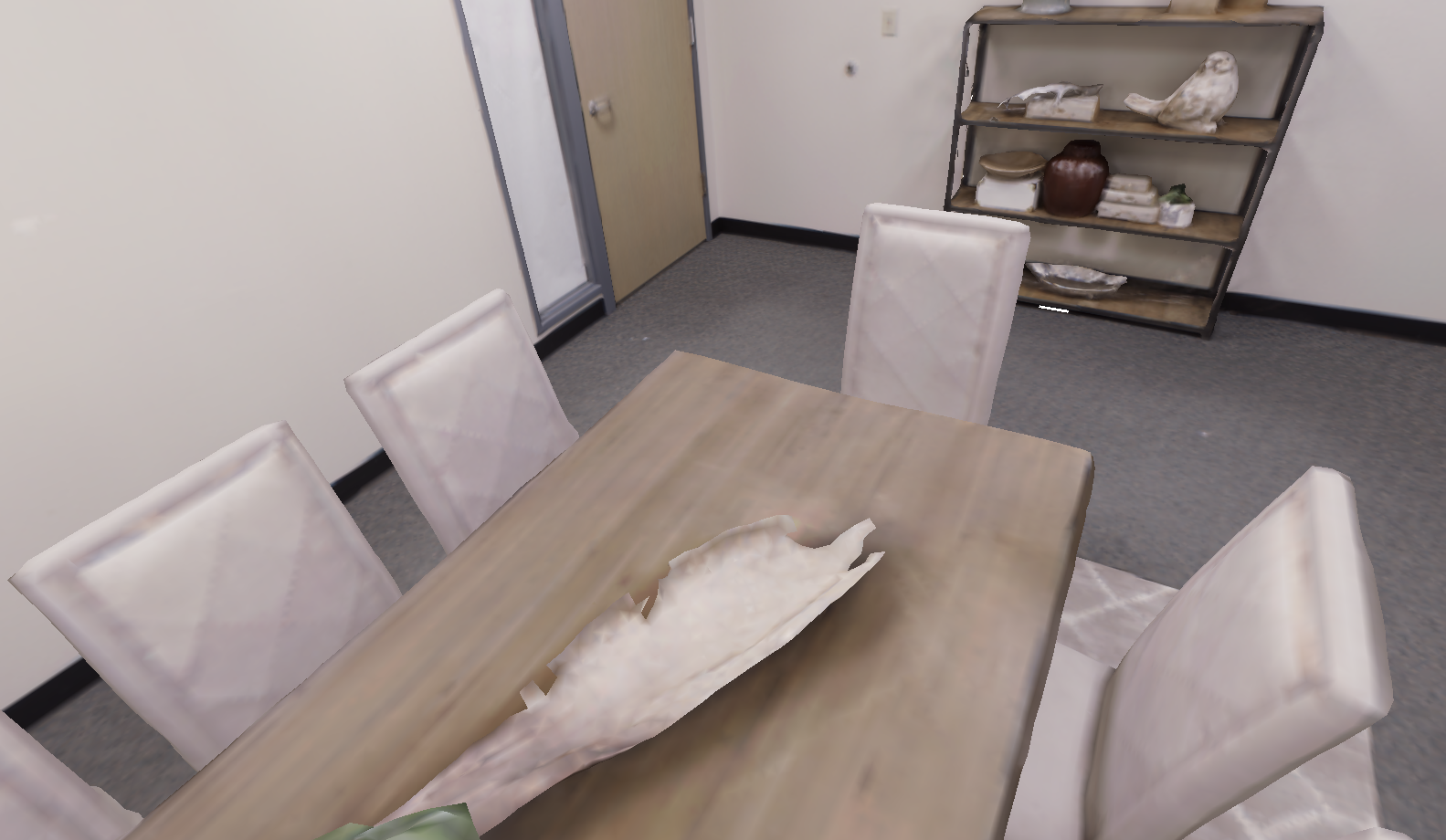}&
\includegraphics[width=.45\linewidth]{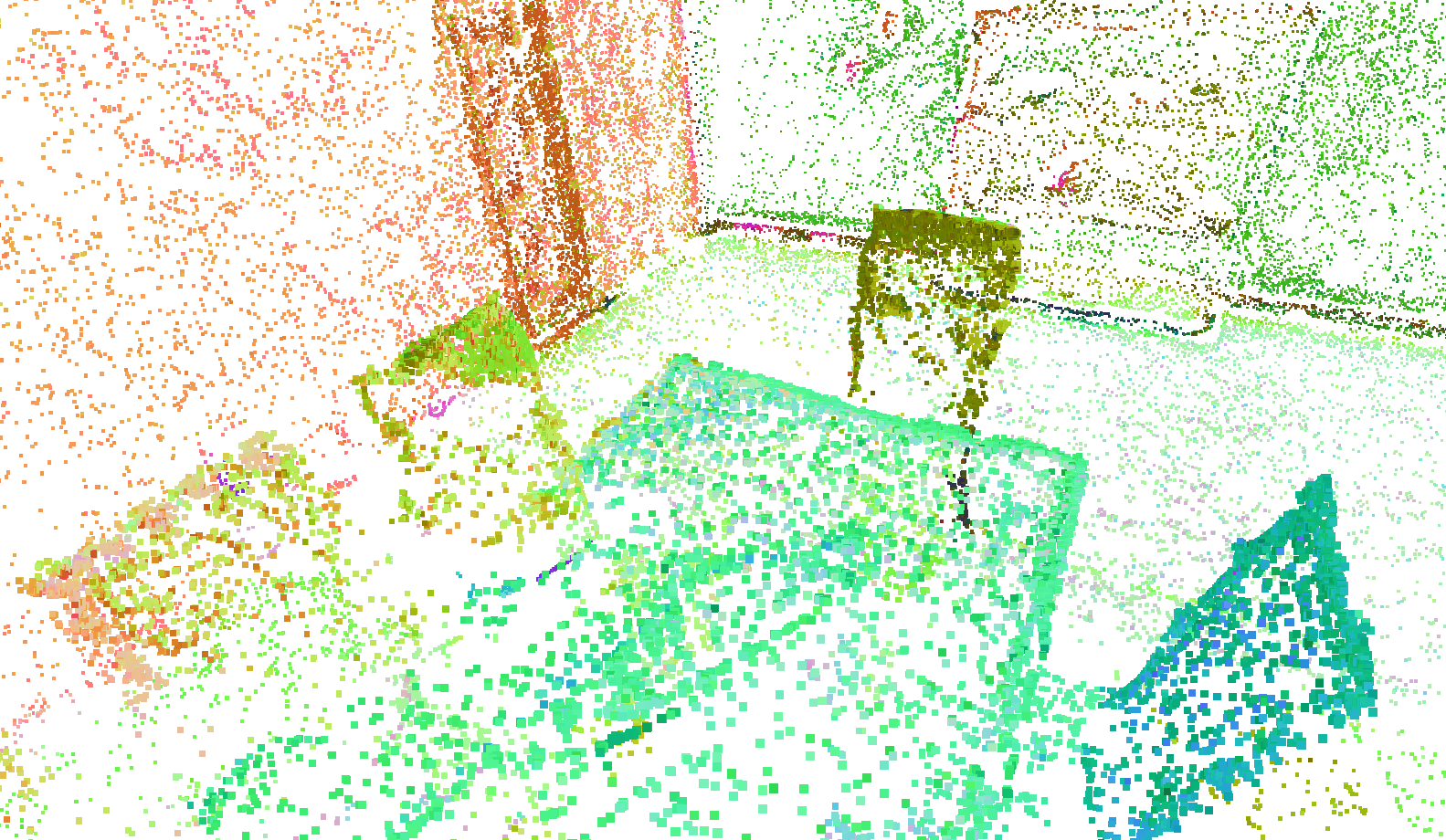}\\
(a) & (b)\\
\caption{\small Visualization of learnt plane descriptors by planar Gaussian splatting for sample ScanNet and Replica scenes. (a) Ground-truth textured meshes, (b) 3D Gaussian point cloud, colorized by the corresponding plane descriptor. }
\label{fig:vis_scannet_desc}
\end{longtable}

\end{document}